\documentclass{elsarticle}
\usepackage[utf8]{inputenc}
\usepackage{amscd}      
\usepackage{amssymb}    
\usepackage{amsbsy}     
\usepackage{amsmath}    
\usepackage{amsfonts}   
\usepackage{mathrsfs}   
\usepackage{enumerate}  
\usepackage{fancybox}   
\usepackage{geometry}   
\usepackage{xcolor}     
\usepackage{academicons} 

\usepackage[linesnumbered,boxed,longend]{algorithm2e} 

\usepackage{tikz}       
\usepackage{pgfplots}   
\usetikzlibrary{arrows,calc,decorations,shapes}

\usepackage{cancel}     

\usepackage{caption}    
\usepackage{fancyhdr}   
\usepackage{float}      
\usepackage{graphics}   
\usepackage{graphicx}   
\usepackage{subcaption} 
\usepackage[export]{adjustbox} 

\usepackage{bbm}        
\usepackage{bm}
\usepackage[OT1]{fontenc}

\usepackage{bookmark}   
\usepackage{hyperref}   

\usepackage{booktabs}   

\usepackage{enumitem}   

\hypersetup{urlcolor=blue,colorlinks=true,linkcolor=black,citecolor=black}

\definecolor{cornflowerblue}{HTML}{6495ed}

\def\Ito{It$\hat{\text{o}}$}

\newcommand{\bfquad}[1]{\textbf{#1}.\quad}

\definecolor{orcidlogocol}{HTML}{A6CE39}

\newcommand{\orcid}[1]{\href{https://orcid.org/#1}{\textcolor{orcidlogocol}{\aiOrcid}}}

\newtheorem{theorem}{Theorem}[section]

\newtheorem{example}[theorem]{Example}

\begin{document}


\begin{frontmatter}

\author[cwi]{Jorino van Rhijn\corref{cor1}}
\ead{J.van.Rhijn@cwi.nl}
\author[uu]{Cornelis W.~Oosterlee}
\author[tud,rabo]{Lech A.~Grzelak}
\author[cwi]{Shuaiqiang~Liu}

\cortext[cor1]{Corresponding author}
\address[cwi]{Centrum Wiskunde en Informatica (CWI), Amsterdam, Netherlands}
\address[uu]{Mathematical Institute, Utrecht University, Utrecht, Netherlands }
\address[tud]{Delft University of Technology, Delft, Netherlands}
\address[rabo]{Rabobank, Utrecht, Netherlands}


\title{Monte Carlo Simulation of SDEs using GANs}







\begin{abstract}
    Generative adversarial networks (GANs) have shown promising results when applied on partial differential equations and financial time series generation. We investigate if GANs can also be used to approximate one-dimensional \Ito\ stochastic differential equations (SDEs). We propose a scheme that approximates the path-wise conditional distribution of SDEs for large time steps. Standard GANs are only able to approximate processes in distribution, yielding a weak approximation to the SDE. A conditional GAN architecture is proposed that enables strong approximation. We inform the discriminator of this GAN with the map between the prior input to the generator and the corresponding output samples, i.e. we introduce a `supervised GAN'. We compare the input-output map obtained with the standard GAN and supervised GAN and show experimentally that the standard GAN may fail to provide a path-wise approximation. The GAN is trained on a dataset obtained with exact simulation. The architecture was tested on geometric Brownian motion (GBM) and the Cox-Ingersoll-Ross (CIR) process. The supervised GAN outperformed the Euler and Milstein schemes in strong error on a discretisation with large time steps. It also outperformed the standard conditional GAN when approximating the conditional distribution. We also demonstrate how standard GANs may give rise to non-parsimonious input-output maps that are sensitive to perturbations, which motivates the need for constraints and regularisation on GAN generators. 
\end{abstract}



\end{frontmatter}

\flushleft

\section{Introduction}
\label{sec:Introduction}

A significant amount of research has been conducted on generative adversarial networks (GANs), with particularly successful application on image generation problems \cite{goodfellow2014generative,radford2016DCGAN,reed2016text_to_image,Zhu_2017_CycleGAN}. However, GANs are also found to be notoriously unstable during training \cite{arjovsky2017towards,arjovsky2017wasserstein}, while their output is difficult to analyse, although various heuristics have been proposed \cite{salimans2016improved,heusel2017gans_incl_rolling_ball}. Interpreting key properties of GANs explicitly, such as the map learned by the generator, or its output distribution, is typically not possible for image problems. In this work, we propose a sampling scheme for \Ito\ stochastic differential equations (SDEs), where we approximate the path-wise conditional distribution of SDEs with a conditional GAN. The SDE framework allows us to interpret qualities such as the map learned by the generator and the output distribution explicitly, since the flow map between two time steps is available explicitly for some SDEs. We investigate whether our GAN-based scheme can provide a \textit{path-wise} approximation \cite{oosterlee2019_book} to one-dimensional \Ito\ SDEs. Compared to traditional methods for solving SDEs, the introduction of deep learning-based schemes offers large potential benefits when scaling to higher dimensional problems and overcoming the curse of dimensionality \cite{berner2020analysis,grohs2018proof}. Our main contributions are as follows: 

\begin{itemize}
    \item We propose a deep learning-based scheme to construct SDE paths for large time steps. A path for any 1D \Ito\ SDE can be sampled by approximating the path-wise conditional distribution with a GAN. 
    \item We propose a `supervised GAN' to study the input-output map learned by the generator and relate this map to the ability to approximate the SDE path-wise. We show that vanilla GANs may produce non-parsimonious input-output maps that are sensitive to perturbations, motivating the use of constraints on the generator map during training. 
\end{itemize}

\subsection{Earlier work}

SDEs are prevalent in models of stochastic dynamical systems in engineering, physics, healthcare, and myriad other domains \cite{oksendal2013stochastic}. In finance, they are cornerstone to the modelling of asset prices and interest rates, with applications in portfolio management or the pricing of financial derivatives and related products \cite{kloeden2012numerical}. In general, the analytical solution to SDEs is not available, which is why practitioners make extensive use of numerical approximations to simulate paths in a Monte Carlo setting \cite{kloeden2012numerical}. However, a high-quality numerical approximation may be too costly in an \textit{online} setting for practical purposes. At the same time, a continuous representation of the path is not of interest in many applications, but rather the solution at specific times along the path. Through \textit{exact simulation} of an SDE, the exact values of the process underlying the SDE are sampled at a pre-determined set of times, cf. \cite{broadie2006exact}. However, for general SDEs, exact simulation may not be available. One alternative technique is the \textit{stochastic collocation Monte Carlo} (SCMC) method \cite{grzelak2019collocating}, in which the conditional inverse distribution of an SDE is approximated with a polynomial expansion, e.g. in a Gaussian random variable. Our goal is not to compete with the SCMC algorithm in financial applications, but rather to initiate a new direction for Monte Carlo estimation of SDEs, where the wide applicability of GANs can be demonstrated. The SCMC method only provides an approximation of the conditional distribution given a fixed choice of the time step, the previous value of the process, and the SDE parameters. In \cite{liu2020sevenleague}, this is addressed by combining the SCMC method with a neural network (NN), the `Seven League Scheme', that predicts the collocation points for the SCMC method, conditional on all model parameters. In our work, the scope is similar, but the conditional distribution will be approximated directly by a conditional GAN, instead of using the SCMC method. This retains the advantage of the Seven League scheme being able to incorporate the dependence on the model parameters and the time step. In addition, if the method can be scaled up to higher dimensions, it exploits the ability of deep learning to combat the curse of dimensionality, where the SCMC method requires the definition of a grid of collocation points \cite{grzelak2019collocating,liu2020sevenleague}, which could be expensive in high dimensions. \par 




GANs have been succesfully applied on solving (stochastic) PDEs \cite{yang2020physics,xie2018tempogan,joshi2019generative}, however these works rely on application of the PDE operator on outcomes generated with a NN. In the case of \Ito\ SDEs, however, the Brownian motion term precludes differentiability of the dynamics. If the diffusion parameter is constant, Abbati et al. show \cite{abbati2019ares} it is possible to define a measure change that allows one to compute the time derivatives of the transformed random process. This would allow one to `match' the moments of the time derivative and solve the SDE, but the requirement for constant diffusion processes is too restrictive for the purposes of this work. \par 

Another approach is to apply the `neural ODEs' by Chen et al. \cite{Chen2018} on SDEs. Kidger et al. \cite{Kidger2021_SDE_GANs} `fit' SDEs to time series data, where the SDE coefficients are given by NNs. A GAN architecture is used here as well, where the solution to the SDE defines the output of the `neural SDE'. The discriminator takes the generated process as input and is itself defined as an SDE, allowing the model to be defined in continuous-time. The model allows the generation of time series data that is equal in distribution to the target, although not necessarily path-wise. We focus on the practical simulation of SDEs, where large time steps are essential and the continuous representation of the process is not of interest. \par 

Instead of focusing directly on solving the SDE, a NN could be used to construct samples that share the same conditional distribution as the target data, which is modelled as a time series, as shown for example in \cite{wiese2019quant,sig_wass_ni2020conditional,fu2019time}. These authors show that the output of their NNs is adapted to the input sequence $\{Z_k\}$ of i.i.d. $N(0,1)$ random variables, which means that it could find a weak solution to the SDE. However, their approaches would provide no guarantees of finding a strong solution to the SDE, i.e.\ path-wise approximation given the same Brownian motion on which the SDE is defined. Details about the difference between weak and strong solutions will be further explored in Section \ref{sec:SDEs_GANs}. \par 

In a similar fashion to neural SDEs, in \cite{cuchiero2020_GANs_LSV} a GAN architecture is proposed that calibrates stochastic local volatility models to market data. This is an example of a data-driven inverse problem using GANs. We, however, will not assume any knowledge about the structure of the SDE. \par 


In this work, we introduce a modified GAN, that we will refer to as a `supervised GAN', which approximates a strong solution to the SDE. We compare this GAN to the `standard' conditional GAN, which only yields a weak approximation. Our setting allows us to interpret the conditional output distribution of the generator using non-parametric statistics, as well as the map learned by the generator explicitly. We show that although `standard' GANs and our modified GAN both approximate the same distribution, their generators may represent very different maps. Meanwhile, the supervised GAN converges faster than the standard GAN during training. Our work motivates the explicit analysis of the map obtained by a model through unsupervised learning, which is relevant in any generative modelling application, from the generation of time series to image generation. 


The paper is structured as follows. First, the necessary background behind SDEs and GANs is introduced. Then, the supervised GAN is introduced to allow strong approximation of SDEs, using a training set obtained from the conditional distribution of the SDE. Section \ref{sec:Results} shows the key results obtained using our method, followed by a discussion in Section \ref{sec:Discussion}. Section \ref{sec:Conclusion} provides a conclusion and outlook.

\section{Preliminaries about SDEs and GANs}
\label{sec:SDEs_GANs}
In this section, we discuss the preliminaries underlying SDEs and GANs, notably weak and strong solutions, discrete-time approximation and conditional GANs. 

\subsection{SDE Definition} 
\label{subsec:SDEs}

Suppose we are given a probability space $(\Omega,\mathcal{F},P)$ and let $\{W_t\}_{t\geq0}$ be a standard Brownian motion on $\mathbb{R}$, adapted to its natural filtration $\mathcal{F}_t:=\sigma\left(\{W_s: s\leq t\}\right)$. A one-dimensional SDE of the \Ito\ type is then defined as follows \cite{oksendal2013stochastic}: 
\begin{align}
    \begin{split}
        &dS_t = A(t,S_t)dt + B(t,S_t)dW_t,\quad  S_0 \in \mathbb{R},
        \label{eq:SDE_diff}
    \end{split}
\end{align}
where $\{S_t\}_{t\geq 0}$ is a continuous-time random process on $\mathbb{R}$ adapted to $\mathcal{F}_t$. $A(t,S_t)$ and $B(t,S_t)$ are themselves $\mathcal{F}_t$-measurable random processes on $\mathbb{R}$. One could write the SDE equivalently in its \Ito\ integral form, as follows \cite{oksendal2013stochastic}: 
\begin{equation}
    S_t = S_0 + \int_{0}^t A(\tau,S_{\tau})d\tau + \int_0^t B(\tau,S_{\tau})dW_{\tau},\ \ \  \forall t \geq 0\ \ \  P\text{-a.s.}
    \label{eq:SDE_int}
\end{equation}
From now on, we write a random process $\{\cdot\}_{t\geq 0}$ succinctly as $\{\cdot\}$. We will refer to a realisation of the process $\{S_t\}$ over a finite time period as a \textit{path}. Note that a path is completely defined once the Brownian motion $\{W_t\}$ has taken a realisation on the respective time interval. The nature of $\{S_t\}$ as a random process complicates the notion of the existence and uniqueness of the solution of an SDE. Suppose that $\{S_t\}$ is a solution to Equation \eqref{eq:SDE_diff}. A solution is called \textit{path-wise unique} if the following holds for any other $\mathcal{F}_t$-adapted solution $\{S_t^{\prime}\}$ \cite{klenke2008probability}:
\begin{equation}
    P \left(  S_t = {S}^{\prime}_t \right) = 1, 
\end{equation}
i.e.\ the paths corresponding to the solution are equal $P\text{-a.s}$. We distinguish between a \textit{strong solution} and a \textit{weak solution}. If we are given a Brownian motion $\{W_t\}$, a strong solution is the path-wise unique solution of Equation \eqref{eq:SDE_diff} corresponding to that Brownian motion. A weak solution also satisfies Equation \eqref{eq:SDE_diff}, but may be defined with respect to a different Brownian motion than $\{W_t\}$ or even a different probability space. A solution is called \textit{weakly unique} if it is equal in law to any other solution $\{S_t^{\prime}\}$: $S_t \overset{\mathscr{L}}{=} S_t^{\prime}$ \cite{klenke2008probability}. Both weak and strong solutions are weakly unique, but only a strong solution is path-wise unique \cite{oksendal2013stochastic}. A unique strong solution exists if $A(t,S_t)$ and $B(t,S_t)$ satisfy Lipschitz conditions, if $S_0$ is independent of $\mathcal{F}_t$ and if the process $\{S_t\}$ is square-integrable for all $t$, see \cite{oksendal2013stochastic} for details. A sufficient condition for a weak solution is that $A(t,S_t)$ and $B(t,S_t)$ must be bounded and continuous and $\lvert B(t,S_t)\rvert \geq \varepsilon > 0$ for some positive real $\varepsilon$ \cite{kloeden2012numerical}. The often cited conditions for weak and strong solutions are sufficient, but not necessary, as is clear from multiple examples of SDEs that do not satisfy the conditions, but still have a strong solution \cite{oksendal2013stochastic,kloeden2012numerical}. In the following, we will restrict ourselves to SDEs for which a strong solution exists. 


\subsection{Discrete-time schemes}
\label{subsec:weak_strong_convergence}
It is possible to approximate Equation \eqref{eq:SDE_int} by a discrete-time scheme, based on a stochastic Taylor expansion, such as the Euler or Milstein schemes \cite{kloeden2012numerical}. Recall that in the 1D case, the Euler and Milstein schemes are given by \cite{kloeden2012numerical}:
\begin{align}
    \tilde{S}_{t+\Delta t} &= \tilde{S}_t + A(t,\tilde{S}_t)\Delta t + B(t,\tilde{S}_t)\Delta W_t + \zeta\left[ \frac{1}{2}B(t,\tilde{S}_t)B^{\prime}(t,\tilde{S}_t)\left( \Delta W_t^2 - \Delta t \right) \right], 
\end{align}
where $\zeta = 0$ for the Euler and $\zeta = 1$ Milstein scheme. $\Delta t$ is the time step of the discretisation, $\Delta W_t := W_{t+\Delta t} - W_t$ and $B^{\prime}:= \frac{\partial B}{\partial S_t}$. We denote the discrete-time approximation of $S_{t}$ by $\tilde{S}_t$. A key property of these schemes is that they approximate the strong solution $\{S_t\}$ of an SDE, if it exists \cite{kloeden2012numerical}. In order to quantify their accuracy, we define the \textit{weak error} $e_w$ and the \textit{strong error} $e_s$ as follows, for $t\geq 0$: 
\begin{align}
    e_w &:= \left\lvert \mathbb{E} f\big( S_{t} \big) - \mathbb{E} f\big( \tilde{S}_{t}\big) \right\rvert, \\ 
    e_s &:= \mathbb{E} \big| S_{t} - \tilde{S}_{t} \big|, 
    \label{eq:weak_strong_err}
\end{align}
where $f$ is some real-valued polynomial function. Note how the weak error describes how much the approximation differs in distribution, i.e.\ how it differs from a weakly unique solution, while the strong error indicates how much the approximation differs path-wise from the strong solution. The convergence rate of a discrete-time scheme can be expressed in terms of $\Delta t$: the weak error of both the Euler and Milstein schemes can be shown to be of $O(\Delta t)$, while the strong error is of $O(\sqrt{\Delta t})$ for the Euler scheme and of $O(\Delta t)$ for the Milstein scheme \cite{kloeden2012numerical}. 

\subsection{Generative Adversarial Networks}
\label{subsec:GANs}

A GAN is a combination of two NNs that are trained adversarially, cf. \cite{goodfellow2014generative}. During training, the \textit{generator} network iteratively maps a prior input to a new random sample, while the \textit{discriminator} network alternatingly receives either a sample from the generator or the training set of reference samples. The discriminator assigns a score on $[0,1]$ to the input it receives. The input samples to the discriminator are labeled either 0 (`fake', coming from the generator) or 1 (`real', coming from the training set)\footnote{Note that it is possible to define alternative labels, e.g. 0.1 and 0.9, or smooth variants, which sometimes improves results in practice \cite{salimans2016improved}.}. The output of the discriminator can be interpreted as the confidence it assigns to the input being `real'. Suppose that the generator $G_{\theta}$ is parameterised by $\theta \in \mathbb{R}^p$ and the discriminator $D_{\alpha}$ is parameterised by $\alpha \in \mathbb{R}^q$, for some $p,q \in \mathbb{N}$. The GAN objective function can then be defined in terms of both $G_{\theta}$ and $D_{\alpha}$ as follows: 
\begin{equation}
\label{eq:GAN_value_function}
    V(G_{\theta},D_{\alpha}) =  \mathbb{E}_{X\sim P^*} \left[  \log D_{\alpha}(X) \right] + \mathbb{E}_{Z\sim P_Z} \left[ \log \left(1-D_{\alpha} \circ G_{\theta}(Z)\right) \right],
\end{equation}
where $P^*$ is the target distribution associated with the training data and $P_Z$ is the prior distribution from which input samples to the generator are drawn. `$\circ$' denotes the composition of functions. The value function captures the degree to which the discriminator succeeds in recognising real samples (first term) and recognising `fake' samples (second term). From the generator's point of view, this is the other way around and the second term is inversely related to its performance. The roles of the generator and discriminator give rise to the following adversarial objective:
\begin{equation}
\label{eq:GAN_minimax_formulation}
    \underset{\theta}{\inf} \ \underset{\alpha}{\sup}\ V(G_{\theta},D_{\alpha}),
\end{equation}

Note the resemblance with a two-player zero-sum game and minimax theory \cite{goodfellow2014generative,peters2015GameTheory}. It can be shown that a solution to Equation \eqref{eq:GAN_minimax_formulation} coincides with equality in distribution between the target $P^*$ and generator output distribution $P_{\theta}$ \cite{goodfellow2014generative,biau2020some}. 

The generator and discriminator are each given their own loss function, based on Equations \eqref{eq:GAN_value_function} and \eqref{eq:GAN_minimax_formulation}: 
\begin{align}
    L_D &= -\mathbb{E}_{X \sim P^*} \left[ \log (D_{\alpha}(X)) \right] - \mathbb{E}_{Z \sim P_Z} \log \left(1-D_{\alpha} \circ G_{\theta}(Z)\right) \label{eq:L_D}, \\ 
    L_G &= \mathbb{E}_{Z \sim P_Z} \log \left(1-D_{\alpha} \circ G_{\theta}(Z)\right), \label{eq:L_G}
\end{align}
for which the minima are found with a suitable gradient descent algorithm. Equation \eqref{eq:L_G} tends to give vanishing gradients during training, which is why it is often replaced by $L_G = - \mathbb{E}_{Z \sim P_Z} \log \left(D_{\alpha} \circ G_{\theta}(Z)\right)$ \cite{arjovsky2017towards}. We adopt this modification as well. \par 

We will refer to the GAN described so far as the `vanilla GAN', as it forms the basis for further extensions. One key extension is the \textit{conditional GAN}, introduced in \cite{mirza2014conditional}. In this architecture, the generator and discriminator receive a vector with conditional information as an additional input, which allows the GAN to learn how the output should vary based on a condition label, e.g.\ generating images of apples or oranges based on the given input. Aside from the appearance of the conditional label, the loss functions remain unchanged. If we let $y \in \mathbb{R}^{n_c}$ be a vector with $n_c$ conditional inputs, the joint objective function becomes \cite{mirza2014conditional}: 
\begin{equation}
\label{eq:conditional_GAN_value_function}
    \underset{\theta}{\inf} \ \underset{\alpha}{\sup} \Big[ \mathbb{E}_{X\mid y \sim P^*} \log D_{\alpha}(X \mid y) + \mathbb{E}_{Z\sim P_Z} \log \left(1-D_{\alpha} \circ G_{\theta}(Z \mid y)\right) \Big],  
\end{equation}
with similar expressions for the loss functions as in equations \eqref{eq:L_D} and \eqref{eq:L_G}.

\subsection{The generator as a parametric map}
\label{subsec:GAN_as_parametric_map}

The GAN is part of a class of methods with for approximating the distribution of a target random variable $X \sim P_X$, starting from a prior $Z\sim P_Z$. Let us assume that both $X,Z \in \mathbb{R}^n$. The model that approximates the target is defined by a map $\varphi_{\theta}: \mathbb{R}^n \rightarrow \mathbb{R}^n$, $z \mapsto \varphi_{\theta}(z)$, with parameter set $\theta \in \mathbb{R}^p$, for some integer dimensions $n,p$. Let us assume the distribution of $\varphi_{\theta}(Z)$ is given by $P_{\theta}$, i.e. the distribution of the output samples is $P_{\theta}$. The goal is then to change the parameters $\theta$ such that $P_{\theta}\overset{d}{\approx} P_X$. In our case, the role of $\varphi_{\theta}$ is taken by the GAN generator. We can use common methods to quantify the `difference' between the distributions $P_{\theta}$ and $P_X$, such as the Jensen-Shannon (JS) divergence \cite{lin1991_JSdivergence}. This quantity is defined for any two absolutely continuous distribution measures $P$ and $Q$ through the Kullback-Leibler (KL) \cite{KLdiv} divergence as: 
\begin{align}
    JS(P\rVert Q) &= \frac{1}{2} \left( KL(P \rVert M) + KL(Q \rVert M) \right), \\
    KL(P\rVert Q) &= \int_{\mathcal{X}} \log \left(\frac{p(x)}{q(x)}\right) q(x) dx, 
\end{align}
where $p$ and $q$ are the densities associated with respectively $P$ and $Q$, $M = \frac{P+Q}{2}$ and $\mathcal{X} \subseteq \mathbb{R}^n$ is the support of both distributions. It can be shown that $JS(P\rVert Q) \geq 0$, with equality iff $P=Q$ \cite{lin1991_JSdivergence}. The goal is to choose $\theta$ such that it minimises $JS(P_{\theta}\rVert P_X)$. This could be done with standard techniques such as stochastic gradient descent (SGD) \cite{bottou2010_SGD}. However, we now focus on how the map $\varphi_{\theta}$ relates to the induced distribution $P_{\theta}$. In typical problems, this map is not of interest, such as in image generation problems, where no reasonable model exists for $\varphi_{\theta}$, making it very difficult to draw conclusions based on the learned map $\varphi_{\theta}$. However, in this work, we study the map $\varphi_{\theta}$ explicitly, which is required for our strong approximation criterion. Meanwhile, it enables us to make qualitative statements about the map learned by the GAN generator. \\ 

Let us turn to a simple example where we can write the map from $Z$ to $X$ explicitly: the lognormal distribution. 
\begin{example}[\textbf{Lognormal distribution}]
\label{ex:lognormal}
    Let $X,Z \in \mathbb{R}$ and let $X = e^Z$ with $Z\sim N(0,1)$. One function that minimises $JS(P_{\theta} \rVert P_X)$ is given by $\varphi^{+} := e^Z$. However, it is not unique, as we could have equally chosen $\varphi^- := e^{-Z}$ by symmetry of the normal distribution. Both choices yield a JS divergence of exactly $0$ and we may expect an SGD-based algorithm to find either of the solutions in some proportion. 
\end{example}



If the lognormal distribution in Example \ref{ex:lognormal} was approximated by a NN with infinite capacity, i.e.\ one which can represent any continuous map $\varphi: \mathbb{R}^n \rightarrow \mathbb{R}^n$, including $\varphi^{+}$ or $\varphi^{-}$, the set of maps minimising the JS divergence would be $\{\varphi^+,\varphi^-\}$. Note that in $n$ dimensions, i.e. $\bm{X}=[e^{Z_1},e^{Z_2},\hdots,e^{Z_n}]^T$, the set of candidate functions with JS divergence exactly $0$ grows as $2^n$, as each $Z_i\sim N(0,1)$ is individually symmetric about the origin. \par 

In reality, NNs do not have infinite capacity, but the set of maps they can represent is restricted by the parameter set $\Theta \subseteq \mathbb{R}^p$. This means that in general, $JS(P_{\theta} \rVert P_X) > 0$. The parametric map $\varphi_{\theta}$ may only be able to bring the JS divergence down to some $\varepsilon > 0$. The key question we are interested in is how many maps lie within an $\varepsilon$ from optimality, and how different these maps are from the `true' optimum, e.g.\ $\varphi^+$ or $\varphi^-$ in Example \ref{ex:lognormal}. Let us define the collection of maps that lie within an $\varepsilon$ from optimality in terms of the JS divergence as: 
\begin{equation}
\label{eq:JS_eps}
    {\mathcal{V}}_{\Theta}^{\varepsilon}:= \left\{\ \varphi_{\theta}: \Big( 0 <  JS(P_{\theta} \rVert P_X)  < \varepsilon \Big), \theta \in \Theta \right\},
\end{equation}
where we stress the dependence on $\varepsilon$ and $\Theta$. Since we did not make any assumptions on $\varphi_{\theta}$, the class of functions ${\mathcal{V}}_{\Theta}^{\varepsilon}$ found after applying SGD may be very large. The number of elements of ${\mathcal{V}}_{\Theta}^{\varepsilon}$ should increase with $\varepsilon$, as more maps give rise to distributions that lie within an $\varepsilon$ of $P_X$. 

In addition to the finite capacity of the parameter set $\theta \in \Theta$, NNs are trained on finite datasets, not perfectly representing $P_X$. Thus, even if we know the `true' underlying map $\varphi^*(Z)$ from which a dataset $X$ was constructed, we may find a map $\varphi_{\theta} \in {\mathcal{V}}_{\Theta}^{\varepsilon}$ with a gradient descent algorithm that is very different from $\varphi^{*}$. In other words, maps that are close in JS divergence may not be close in function space. Note that although we chose the JS divergence to illustrate the point, we could define similar classes of functions for other divergence measures, such as the KL divergence, total variation distance, etc. The JS divergence is particularly relevant in the case of GANs, as one can show that the optimal generator in Equation \eqref{eq:GAN_minimax_formulation} - given the optimal discriminator - minimises the quantity $JS(P_{\theta}\lVert P^*)$ \cite{goodfellow2014generative,biau2020some}. 


The key observation is that algorithms minimising a distributional quantity or metric do not impose any restrictions on the map $\varphi_{\theta}$. It may for example be highly non-smooth in regions along its support, even though the approximation in distribution is close. Therefore, although it is typically not tractable to study $\varphi_{\theta}$, we argue that qualitative properties of $\varphi_{\theta}$ should be of interest in generative modelling, given its implications for the robustness of the resulting sampling scheme. 




\section{Methodology}
\label{sec:Methodology}

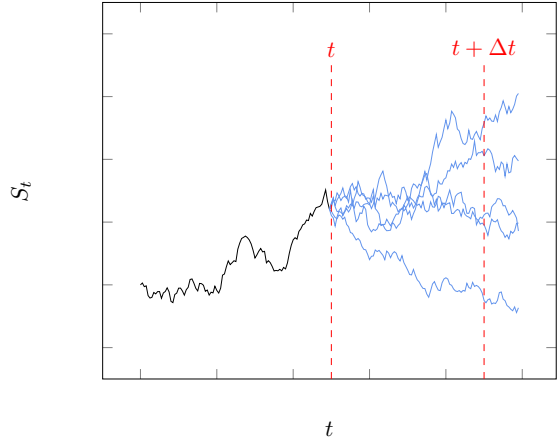
\begin{figure}[h]
    \centering
    \resizebox{0.5\linewidth}{!}{%
    \begin{tikzpicture}
    \begin{axis}[
        ymin=0.7,
        ymax = 1.9,
          xlabel=$t$, 
        ylabel=$S_t$,
        xticklabels={,,},
        yticklabels={,,},
          legend style={at={(0.5,-0.2)},anchor=north},
          x tick label style={rotate=0,anchor=north,yshift=20pt}
        ]
        \addplot[color=black] table [x=t_single, y=single, col sep=comma,mark=none] {./tikz/GBM_tikzpicture_1.csv};
        \foreach \i in {0,...,4}{
        \addplot[color=cornflowerblue] table [x=t_next, y=X_next\i, col sep=comma,mark=none] {./tikz/GBM_tikzpicture_1.csv};
        }
        \addplot[color=red,dashed] coordinates { (0.5,0.7) (0.5,1.7) } node[above] {$t$};
        \addplot[color=red,dashed] coordinates { (0.9,0.7) (0.9,1.7) } node[above] {$t+\Delta t$};
    \end{axis}
    \end{tikzpicture}
    
    }
    \caption{Illustration of the problem setting: given the process $\{S_t\}$ up to time $t$, obtain samples from the process at time $t+\Delta t$ using a GAN.}
    \label{fig:constructing_paths}
\end{figure}

Let $\{S_t\}$ be the strong solution to an SDE as defined in Equation \eqref{eq:SDE_diff}. Suppose we are interested in the solution at $N$ times, i.e.\ $0=t_0<t_1<\hdots<t_N=T$. Let us denote the conditional distribution function of the solution at time $t_k$ given the previous sample by $F_{S_{t_k}\mid S_{t_{k-1}}}$ for $k\in \{1,\hdots,N\}$. When using exact simulation, one samples iteratively from the distribution of $S_{t_k}\mid S_{t_{k-1}}$, cf. \cite{broadie2006exact}. This is possible due to the \textit{Markov property} of \Ito\ SDEs \cite{klenke2008probability}, i.e.\ each $S_{t_k}\mid S_{t_{k-1}}$ is independent of $\mathcal{F}_{t_{k-1}}$ for $k\in \{1,\hdots,N\}$. This allows one to construct a path $\{S_{t_0},S_{t_1},\hdots,S_{t_N}\}$ along the time discretisation by iterated sampling from the distribution of ${S_{t_k}\mid S_{t_{k-1}}}$. Let us from now on assume, without loss of generality, that our discretisation consists of time steps of equal size $\Delta t$. Paths can then be constructed by iteratively sampling from the conditional distribution of $S_{t+\Delta t}\mid S_t$, having initialised the process at some $S_0\in\mathbb{R}$ at $t=0$, which is illustrated in Figure \ref{fig:constructing_paths}. 


If the conditional distribution of $S_{t+\Delta t}\mid S_t$ is approximated with a conditional GAN, i.e.\ conditional on $\Delta t$ and $S_t$, new points can be sampled iteratively along the path with the conditional GAN as follows: 
\begin{equation}
    \hat{S}_{t+\Delta t}\mid \hat{S}_t = G_{\theta}\big( Z,\Delta t,\hat{S}_t \big), 
    \label{eq:iterative_network_definition}
\end{equation}
where $Z\sim N(0,1)$. We denote the approximation of $S_t$ by the GAN by $\hat{S}_t$. This could be further generalised by conditioning on the SDE parameters contained in $A(t,S_t)$ and $B(t,S_t)$ as well. In this work, we will hold them fixed and train the conditional GAN on a dataset of tuples $((S_{t+\Delta t}\mid S_t),S_t,\Delta t)$ with varying $\Delta t$ and $S_t$. 

\subsection{Supervised GAN}
\label{subsec:constrained_GAN}
Since $S_{t+\Delta t}\mid S_t$ is a continuous random variable and $F_{S_{t+\Delta t}\mid S_t}\sim U(0,1)$, for each realisation of $S_{t+\Delta t}\mid S_t$, there is a unique realisation of $U\sim U(0,1)$. Let $F_Z$ denote the cumulative distribution function of the random variable $Z\sim N(0,1)$. Both $F_{S_{t+\Delta t}\mid S_t}$ and $F_Z$ are strictly increasing, as they are based on continuous random variables. Therefore, their distribution functions are bijections and their inverses exist. Thus, for each realisation of $\omega \in \mathcal{F}_t$ and corresponding realisation of the process $\left(S_{t+\Delta t}\mid S_t \right) (\omega)$, there is a unique realisation $U(\omega)$. In turn, for each $U(\omega)$ there is a unique realisation $Z(\omega)$. These realisations are related as follows:

\begin{equation}
    \left( S_{t+\Delta t}\mid S_t \right) (\omega) = F^{-1}_{S_{t+\Delta t}\mid S_t}\left( F_Z(Z(\omega)) \right). 
    \label{eq:sample_S_from_Z}
\end{equation}
In the SCMC scheme, Equation \eqref{eq:sample_S_from_Z} is approximated by a polynomial expansion in $Z$ \cite{grzelak2019collocating,liu2020sevenleague}. This allows path-wise comparison of the SCMC scheme to e.g. a Milstein scheme in \cite{liu2020sevenleague}, where $Z(\omega)$ is used to define the Brownian motion increment between time $t$ and $t+\Delta t$. In this work, we approximate the conditional inverse function directly using the generator of a GAN. However, as we saw in Section \ref{subsec:GAN_as_parametric_map}, an approximation to the target distribution does not imply the underlying map is unique. For a strong approximation to the process, we must approximate Equation \eqref{eq:sample_S_from_Z}. That is, we must ensure that we preserve the relation between the prior $Z(\omega)$ and $(S_{t+\Delta t}\mid S_t)(\omega)$. To this end, we can build a training set of samples $(Z,(S_{t+\Delta t}\mid S_t),S_t,\Delta t)$ as input to both the generator \textit{and} discriminator, where $Z$ is found using the `inverse' of Equation \eqref{eq:sample_S_from_Z}:
\begin{equation}
    Z(\omega) = F_Z^{-1} \left( F_{S_{t+\Delta t}\mid S_t}\left(  S_{t+\Delta t}\mid S_t \right)(\omega) \right). 
    \label{eq:sample_Z_from_S}
\end{equation}
This way, we do not only let the GAN learn the distribution $F_{S_{t+\Delta t}\mid S_t}$, but the map from $Z(\omega)$ to $\left(S_{t+\Delta t}\mid S_t \right) (\omega)$, which carries the information about the event $\omega \in \mathcal{F}_t$ that corresponds to the realisation of the specific value $\left(S_{t+\Delta t}\mid S_t \right) (\omega)$. We will call this architecture the `supervised GAN', as it is a GAN-based equivalent to training a standard feed-forward network on the mean squared error between $F^{-1}_{S_{t+\Delta t}\mid S_t}\left( F_Z(Z(\omega))\right)$ and $G_{\theta}(Z(\omega),S_t,\Delta t)$. The supervised GAN discriminator receives as input $\left(Z(\omega),\left(S_{t+\Delta t}\mid S_t \right) (\omega),S_t,\Delta t\right)$, while the vanilla GAN discriminator only receives $\left(\left(S_{t+\Delta t}\mid S_t \right) (\omega),S_t,\Delta t\right)$ but not $Z(\omega)$. This constrains which input-output map from the generator is allowed. It allows the supervised GAN to perform a path-wise approximation, while the vanilla GAN is only guaranteed to approximate the process in conditional distribution (e.g. may fail to distinguish the output given $-Z$ from $+Z$), yielding a weak approximation. \par 


\subsection{Analysis of the output distribution}
\label{subsec:analysis}

In order to quantify the difference between the conditional distribution of the generator output and $F_{S_{t+\Delta t}\mid S_t}$, we use two non-parametric statistics: the \textit{Kolmogorov-Smirnov} (KS) statistic and the \textit{Wasserstein distance} in 1D. The 2-sample KS statistic is defined as follows, cf. \cite{KS_dist_simard2011computing}: 
\begin{equation}
    u_n := \underset{x}{\sup}\left\lvert F_X(x) - F_Y(x) \right\rvert, 
    \label{eq:KS_stat}
\end{equation}
where $F_X$ and $F_Y$ are two empirical cumulative distribution functions (ECDFs), one of which corresponding to the generator output and the other to the reference distribution. Note that if the CDF of the target distribution is available analytically, we can use it in Equation \eqref{eq:KS_stat} instead of its ECDF. \par 

In 1D, the $r$-Wasserstein-distance (i.e.\ based on the $r$-norm) between two distribution functions $F_X$ and $F_Y$ can be epxressed as follows, for some $r\in \mathbb{R}^+$ \cite{ramdas2017wasserstein}: 
\begin{equation}
    v_r(F_X,F_Y) = \left( \int_0^1 \lvert F_X^{-1}(x) - F_Y^{-1}(x)\rvert^r dx \right)^{\frac{1}{r}}, 
    \label{eq:Wass_dist_1D}
\end{equation}
where $F_{(\cdot)}^{-1}$ denotes the inverse distribution function, i.e.\ the quantile function of the random variable under consideration. In this work, we will set $r=1$ and use the 1-Wasserstein distance. $F_X$ and $F_Y$ are computed from a dataset of samples $\{\hat{X}_i \}_{i=1}^n$ obtained through inference with the GAN and a dataset $\{{X}_i \}_{i=1}^n$ drawn from the reference distribution. \par 

The KS-statistic computes the largest difference between two (E)CDFs, i.e.\ `vertical differences' in the plane $F_X(x)$ versus $x$. In 1D, the Wasserstein distance can be thought of as the average distance between the quantiles of two distributions, i.e.\ `horizontal differences' \cite{ramdas2017wasserstein}. The combination of these statistics simultaneously thus allows us to capture different features of both distributions. \par 

The challenge is, however, to interpret the value of both statistics given a sample of size $N_{\text{test}}$ of both the reference distribution and the GAN output. We could construct a reference value by drawing two i.i.d. vectors of size $N_{\text{test}}$ containing realisations of the reference distribution, say $X,Y \overset{iid}{\sim} F_{S_{t+\Delta t}\mid S_t}$. If we choose $N_{\text{test}}$ too low (e.g.\ $100$), the approximation of the distribution function will be very coarse and both statistics would exhibit a large variance. If we choose $N_{\text{test}}$ high, e.g.\ $10^5$, the KS statistic and Wasserstein distance of this reference value will tend towards $0$. This dependence on $N_{\text{test}}$ makes comparison between the statistics on the GAN output and reference value difficult. In order to avoid a particular choice of $N_{\text{test}}$, we will compute the statistics for a range of values of $N_{\text{test}}$, e.g.\ $\{100,1000,\hdots,10^5\}$ and plot the result versus $N_{\text{test}}$. We will repeat this experiment on a set of $N_{\text{test}}$ samples obtained with a single-step Euler and Milstein approximation, based on the same time step $\Delta t$ and `starting value' $S_t$ that the GAN is tested on. 

\subsection{Data pre-processing}
\label{subsec:pre-processing}
In our setting, the only knowledge of the process $S_{t+\Delta t}\mid S_t$ that we assume to be available are the SDE parameters and the latest value of the process $S_t$. We can leverage the fact that the sample $S_t$ is available, by training the network on the \textit{relative} increase of $S_{t+\Delta t}\mid S_t$ compared to $S_t$, instead of its absolute value. This way, the NN does not need to learn where to place the distribution for each $S_t$, but automatically outputs a distribution in a neighbourhood of $S_t$. Following \cite{wiese2019quant}, we use logreturns and let the conditional GAN approximate the logreturns-transformed process: 
\begin{equation}
    R_{t+\Delta t}\mid S_t := \log\left(\frac{S_{t+\Delta t}\mid S_t}{S_t}\right). 
    \label{eq:logreturns}
\end{equation}
The approximation of the process $S_{t+\Delta t}\mid S_t$ is then obtained with the inverse transform: 
\begin{equation}
    \hat{S}_{t+\Delta t}\mid S_t = S_t e^{G_{\theta}(Z,S_t,\Delta t)}.
    \label{eq:inv_logreturns}
\end{equation}
Using logreturns comes with the additional benefit of centering the distribution near the origin. NNs typically converge faster if the training set is standardised \cite{lecun1998neural}, i.e. if the inputs to the network are of mean zero and unit variance. The variance will still vary with the model parameters, and as we do not assume that the moments of the target distribution are known, we cannot simply standardise the dataset and invert the standardisation step after training. Moreover, financial SDEs are typically heavy-tailed, which makes standardisation with point estimates ineffective. \par

The logreturns transformation comes with a complication for SDEs that can reach values arbitrarily close to zero, such as the CIR process \cite{oosterlee2019_book,cox2005theory}. This means that the numerator and denominator in Equation \eqref{eq:logreturns} can differ by many orders of magnitude (e.g. a sample starting at $0.1$ and jumping to $10^{-6}$ and vice-versa), which leads to large and potentially unbounded output domains after the logreturns transform, which is undesirable. Therefore, an SDE that can jump to and from values near the origin should be pre-processed in a different way. Since we assume the model parameters are known, one could use this knowledge as an alternative to standardisation. For example, the CIR process reverts to a long-term mean $\bar{S}$, which is assumed to be known. We use this parameter to shift and scale the distribution as follows: 
\begin{equation}
    R_{t+\Delta t}\mid S_t := (S_{t+\Delta t}\mid S_t - \bar{S})/\bar{S}, 
    \label{eq:scale_S_bar}
\end{equation}
which is approximated with the conditional GAN. The approximation of $S_{t+\Delta t}\mid S_t$ is then obtained by inverting Equation \eqref{eq:scale_S_bar}. Since values of the process can get arbitrarily close to zero, the generator may output negative values very close to $0$. We `rectify' the output by taking the absolute value of the generator output: $\lvert (R_{t+\Delta t}\mid S_t + 1)\bar{S} \rvert$, making sure the final approximation of the process is in $\mathbb{R}^+$.

\subsection{Network architecture}
\label{subsec:network_architecture}
The generator and discriminator are both implemented as feed-forward NNs, using 4 hidden layers and 200 neurons per layer. A LeakyReLU activation (i.e. $x\mapsto \max(x,0)+a \min(x,0)$, for some $a\in \mathbb{R}$) \cite{maas2013leakyReLU} is chosen as the non-linearity after each layer, except the output layers of the generator and discriminator. This activation is chosen, since the distribution of the inputs and hidden state of the network is heavy-tailed. Saturating activations, such as the tanh function, were therefore found to be less effective. The discriminator is given a logistic function at the output, to force the output to be on $[0,1]$. The generator has no output activation. All implementations are made using PyTorch \cite{paszke2019pytorch} and run on an NVIDIA RTX 2070 Super GPU with 8 GB of memory. See Appendix \ref{appendix:NN_architecture_training} for more details on the architecture and training process. 

\section{Results}
\label{sec:Results}

To assess the GAN, we study three different properties that allow us to compare the vanilla GAN to the supervised GAN. Firstly, we study the ability of both GANs to approximate the conditional distribution $F_{S_{t+\Delta t}\mid S_t}$, for several test values of $\Delta t$ and $S_t$. Secondly, we compute the weak and strong error of artificial paths constructed with the vanilla GAN and supervised GAN. Thirdly, we explicitly study the map from prior sample $Z$ to the sample $S_{t+\Delta t}\mid S_t$ learned by the generator for both GANs. 

\subsection{SDEs under consideration}
\label{subsec:GBM_CIR}

To test the supervised GAN, we choose two common SDEs that have a strong solution: \textit{geometric Brownian motion} (GBM) \cite{leGall2016brownian} and the \textit{Cox-Ingersoll-Ross} (CIR) process \cite{cox2005theory}. The dynamics are given by:
\begin{align}
    \mathrm{GBM}&:\ \ \ dS_t = \mu S_t dt + \sigma S_t dW_t, \label{eq:SDE_GBM} \\
    \mathrm{CIR}&:\ \ \ dS_t = \kappa (\bar{S} - S_t)dt + \gamma \sqrt{S_t} dW_t, \label{eq:SDE_CIR} 
\end{align}
where $\mu$ and $\sigma$ of GBM denote respectively the drift and volatility of the underlying asset. $\kappa$ controls the rate at which the CIR process reverts to its long-term mean $\bar{S}$, while $\gamma$ represents the volatility of the CIR process. 

The conditional distribution of ${S_{t+\Delta t}\mid S_t}$ is available explicitly for both SDEs, which allows the construction of an exact training set and simplifies the interpretation of the results. In the GBM case, application of \Ito's lemma on the process $\log S_t$ allows one to immediately derive that the solution is lognormally distributed \cite[p. 226]{leGall2016brownian}. For the CIR process, one can show that $S_{t+\Delta t} \mid S_t$ follows a scaled non-central $\chi^2$-distribution with some non-centrality parameter $\xi$, degrees of freedom $\delta$ and scaling factor $\bar{c}$ \cite{cox2005theory}: 
\begin{equation}
    S_{t+\Delta t} \mid S_t\  \sim\  \bar{c}\ \chi^2(\xi,\delta),
    \label{eq:sample_CIR}
\end{equation}
where $\bar{c}, \xi$ and $\delta$ are expressed in terms of the SDE parameters \cite{oosterlee2019_book}, see Appendix \ref{appendix:model_parameters} for details. 
The presence of the square root in Equation \eqref{eq:SDE_CIR} introduces a complication when approximating the SDE with a discrete-time scheme, which could take negative values. Therefore, we apply a modified, truncated version of the Euler \cite{labbe2010discr_CIR} and Milstein \cite{hefter2018Milstein_CIR} schemes on the CIR process, see Appendix \ref{appendix:model_parameters} for details. \par 

If $\delta < 2$, the non-central $\chi^2$-distribution exhibits near-singular behaviour in a region near zero, i.e. $(0,q]$ for arbitrarily small $q>0$, allowing the process to `hit' zero \cite{cox2005theory}. If $\delta \geq 2$, the process does not exhibit this property and remains strictly positive. This regime for $\delta$ is known as the \textit{Feller condition} \cite{albrecher2007little_Feller_cond}. For our numerical experiments, we chose two regimes of parameters, one in which the Feller condition is satisfied and one in which it is violated. The near-singular behaviour of the distribution makes the latter case the most challenging. 


\subsection{Approximating the conditional distribution}
\begin{figure}[h!]
\centering
    \begin{subfigure}{0.49\linewidth}
        \centering
        \includegraphics[width=\linewidth]{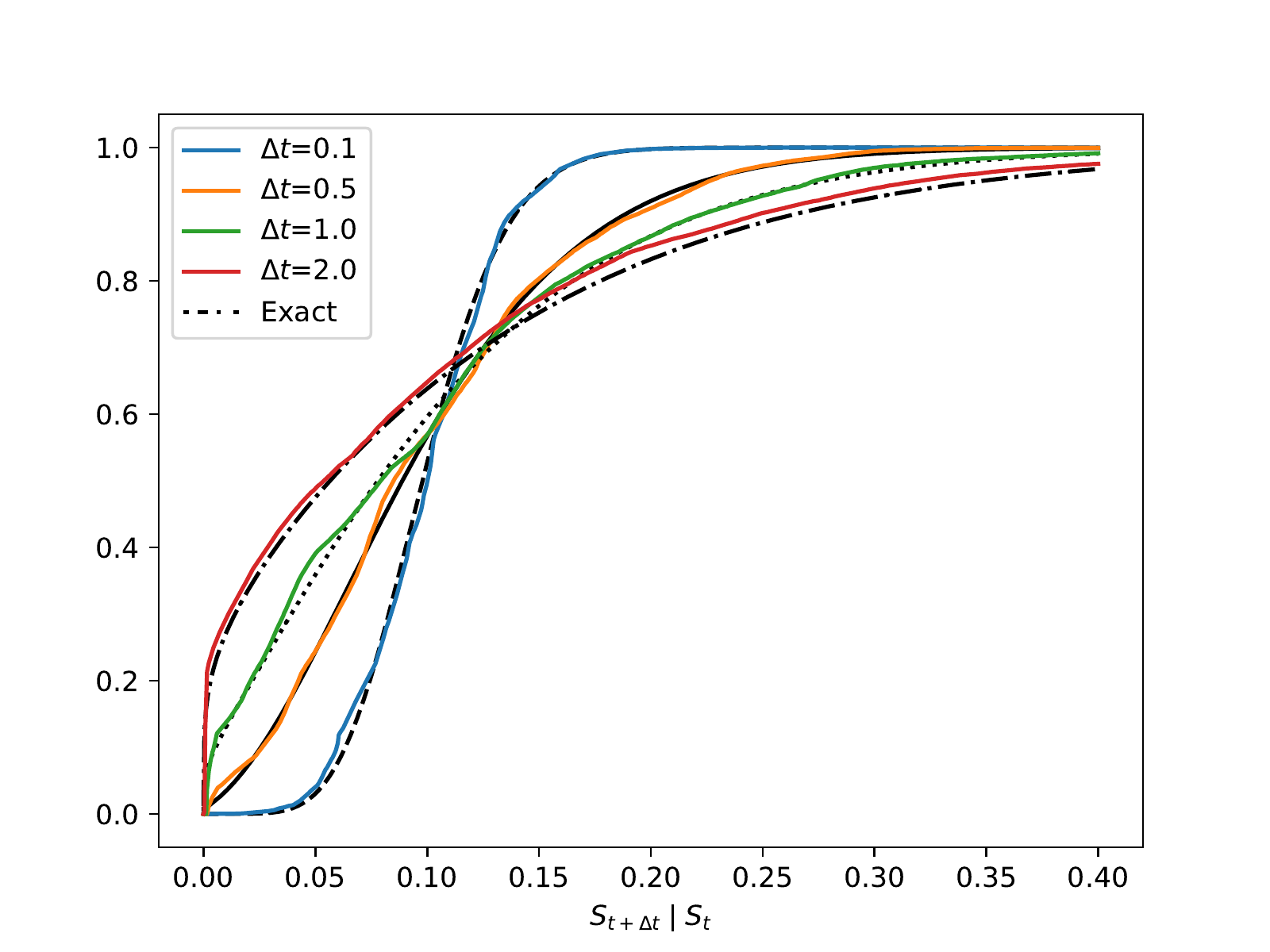}
        \caption{Vanilla GAN}
        \label{fig:ECDF_Feller_not_vanilla_Delta_t}
    \end{subfigure}
    \begin{subfigure}{0.49\linewidth}
        \centering
        \includegraphics[width=\linewidth]{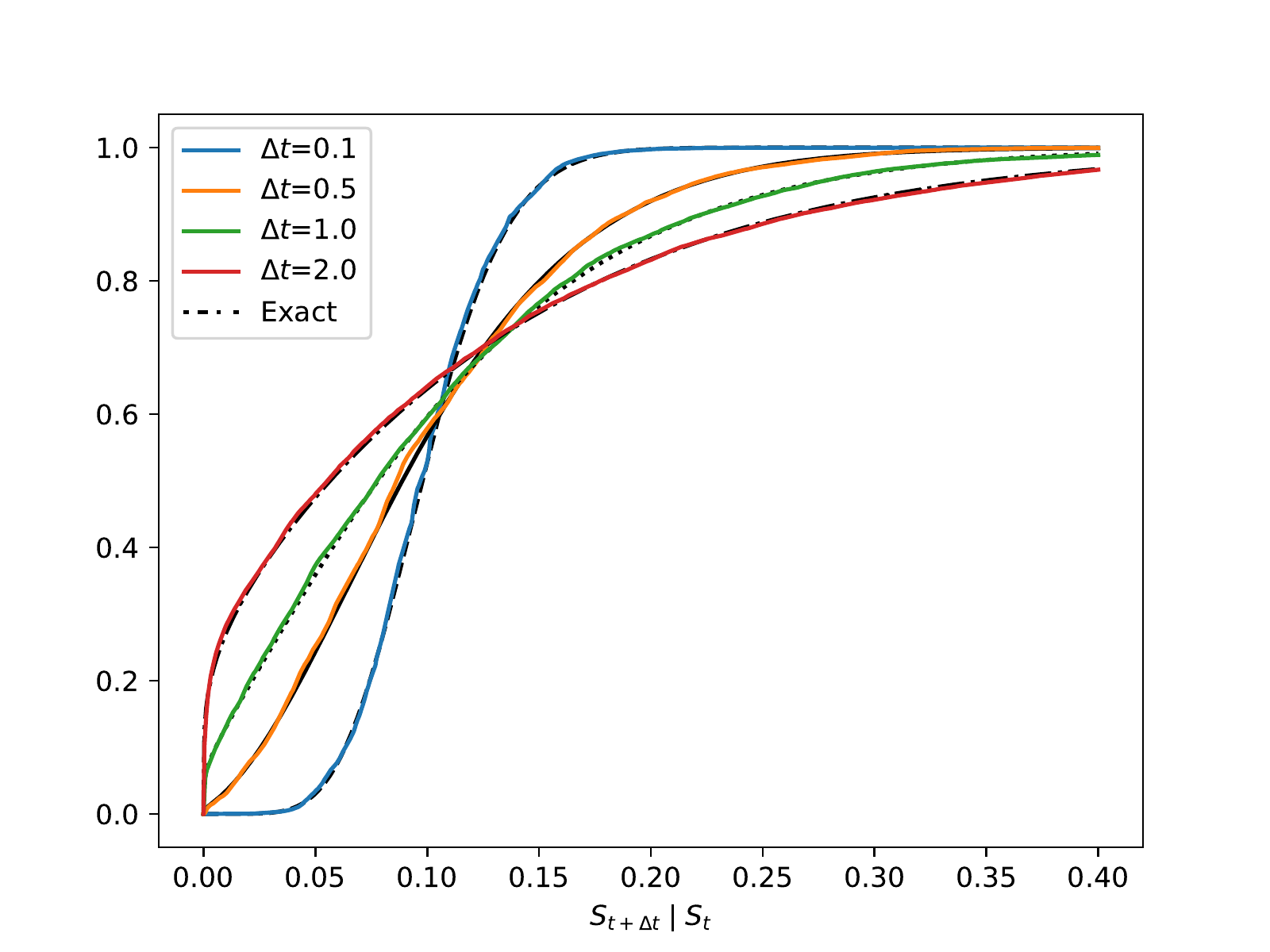}
        \caption{Supervised GAN}
        \label{fig:ECDF_Feller_not_constrained_Delta_t}
    \end{subfigure}
    \caption{ECDF plots of the vanilla and supervised GAN output with $S_t=0.1$ and $\Delta t \in \{0.1,0.5,1,2\}$.}
    \label{fig:ECDF_Feller_not_Delta_t}
\end{figure}

We focus on the CIR dynamics for which the Feller condition is violated. The results for GBM and the case where the Feller condition is satisfied are provided in Appendix \ref{appendix:additional_results}. First, we present the distribution of the output of the vanilla and supervised GAN in Figure \ref{fig:ECDF_Feller_not_Delta_t}, which shows the ECDF of $\hat{S}_{t+\Delta t}\mid S_t$ for fixed $S_t=0.1$ and four choices of $\Delta t$. We compare this with the exact distribution given in black. We see that both GANs adapt the shape of the output distribution to match the exact distribution, while the supervised GAN appears to provide a more accurate approximation. \par

\begin{figure}[h!]
\centering
    \begin{subfigure}{0.49\linewidth}
        \centering
        \includegraphics[width=\linewidth]{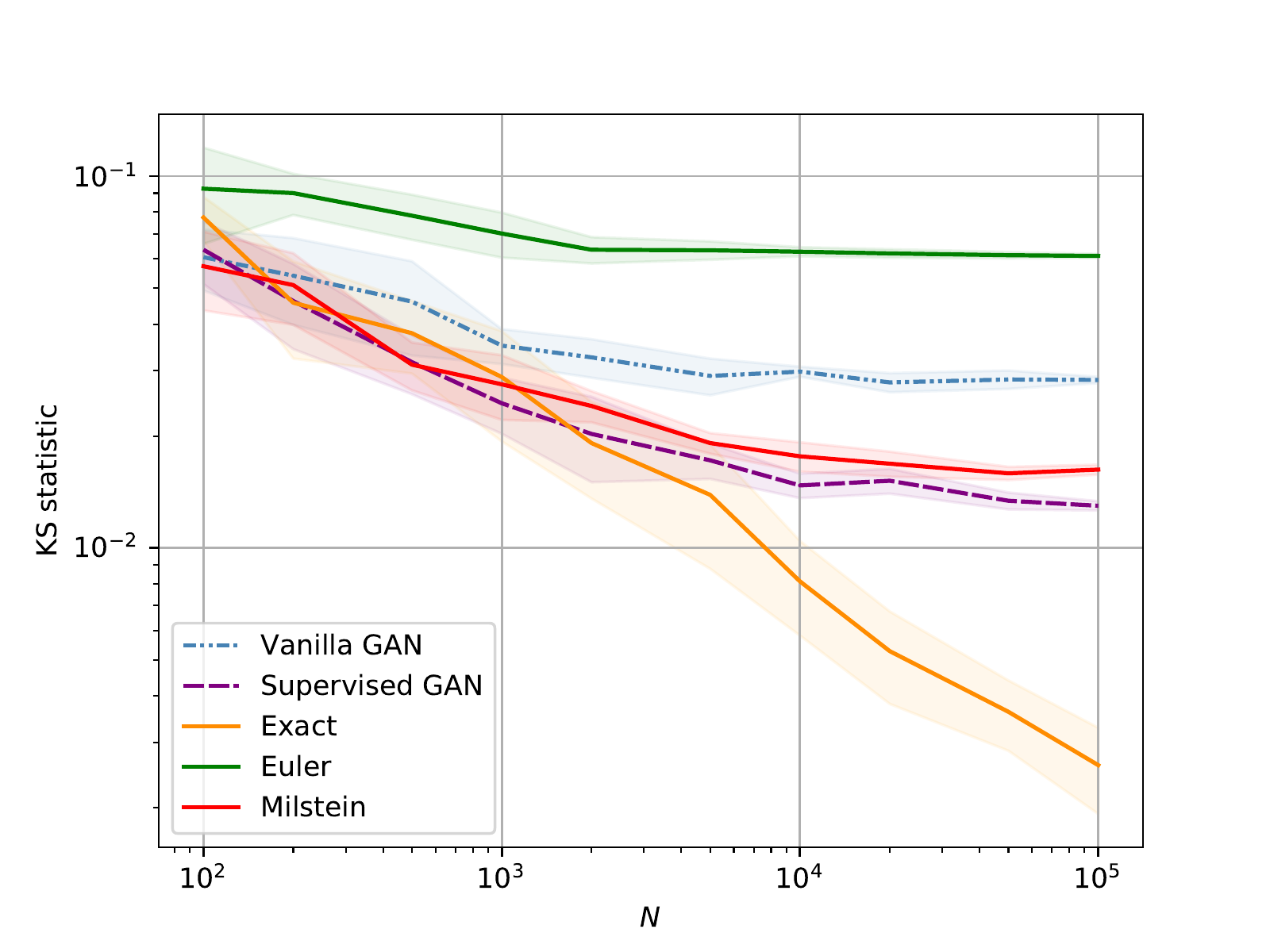}
        \caption{KS statistic}
        \label{fig:benchmark_N_Feller_not_t_0_4_KS}
    \end{subfigure}
    \begin{subfigure}{0.49\linewidth}
        \centering
        \includegraphics[width=\linewidth]{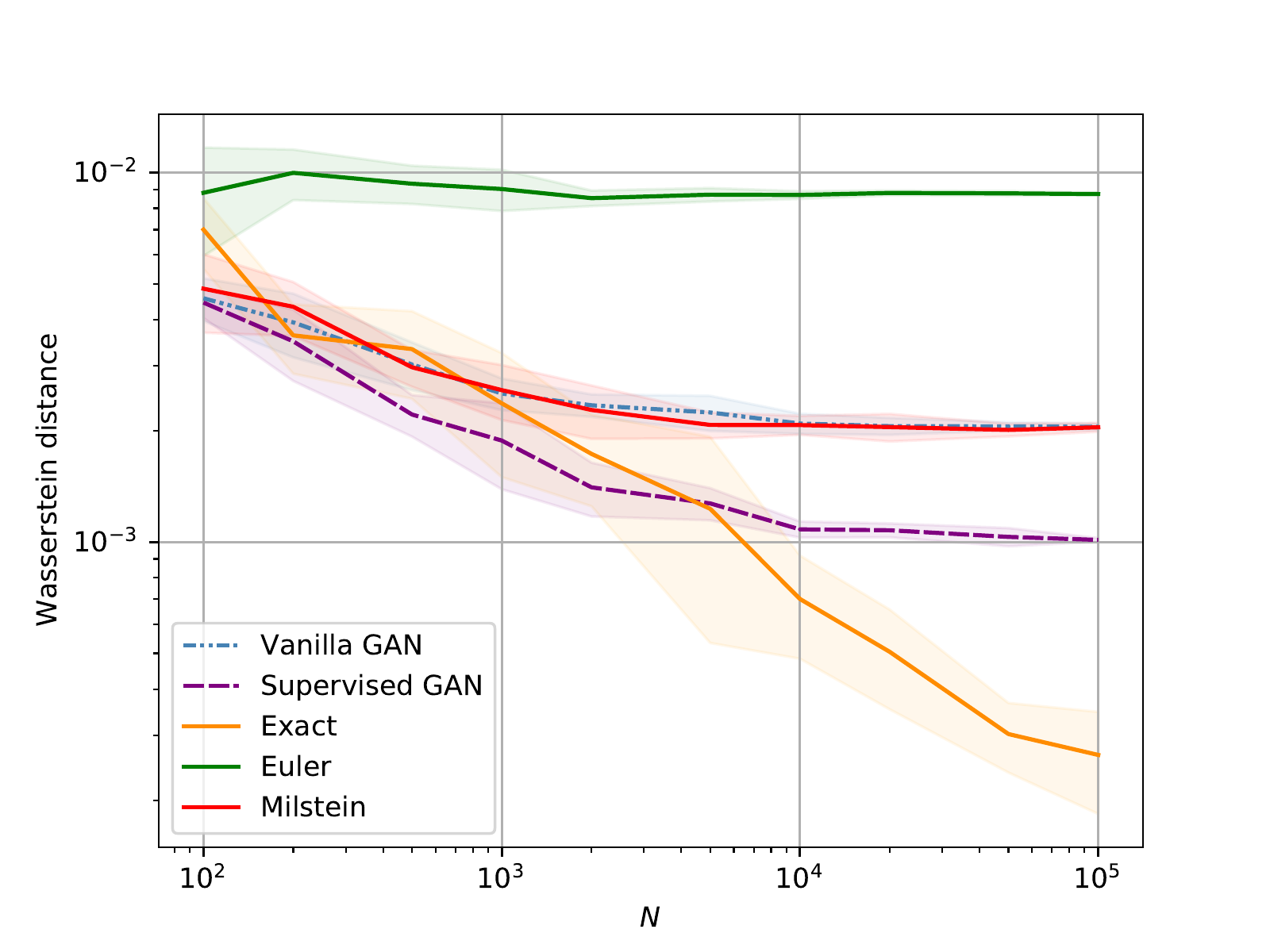}
        \caption{Wasserstein distance}
        \label{fig:benchmark_N_Feller_not_t_0_4_Wass}
    \end{subfigure}
    \caption{KS statistic and Wasserstein distance at $\Delta t=0.4$, versus the size of the test set. The confidence bands show the standard deviation based on 10 repetitions of the experiment, i.e.\ 10 i.i.d. samples of $N$ random inputs to both GANs. The mean of both statistics is reported in the solid and dashed lines.}
    \label{fig:benchmark_N_Feller_not_t_0_4}
\end{figure}

In Figure \ref{fig:benchmark_N_Feller_not_t_0_4}, the KS statistic and Wasserstein distance are reported for a range of test sizes, using the method described in Section \ref{subsec:analysis}. $S_t$ was set to $0.1$. $\Delta t$ was chosen to be $0.4$ in Figure \ref{fig:benchmark_N_Feller_not_t_0_4}, for which the KS statistic was close to the Milstein scheme for the supervised GAN. For $\Delta t > 0.4$, both statistics of the supervised GAN were lower than those of the Milstein scheme, i.e.\ the supervised GAN outperforms both the Euler and Milstein schemes for $\Delta t > 0.4$. The supervised GAN outperforms the vanilla GAN on both statistics. Similar plots can be made for different choices of $\Delta t$ and $S_t$ and similar results were found for GBM and the case where the Feller condition was satisfied. 

\begin{figure}[h!]
\centering
    \begin{subfigure}{0.49\linewidth}
        \centering
        \includegraphics[width=\linewidth]{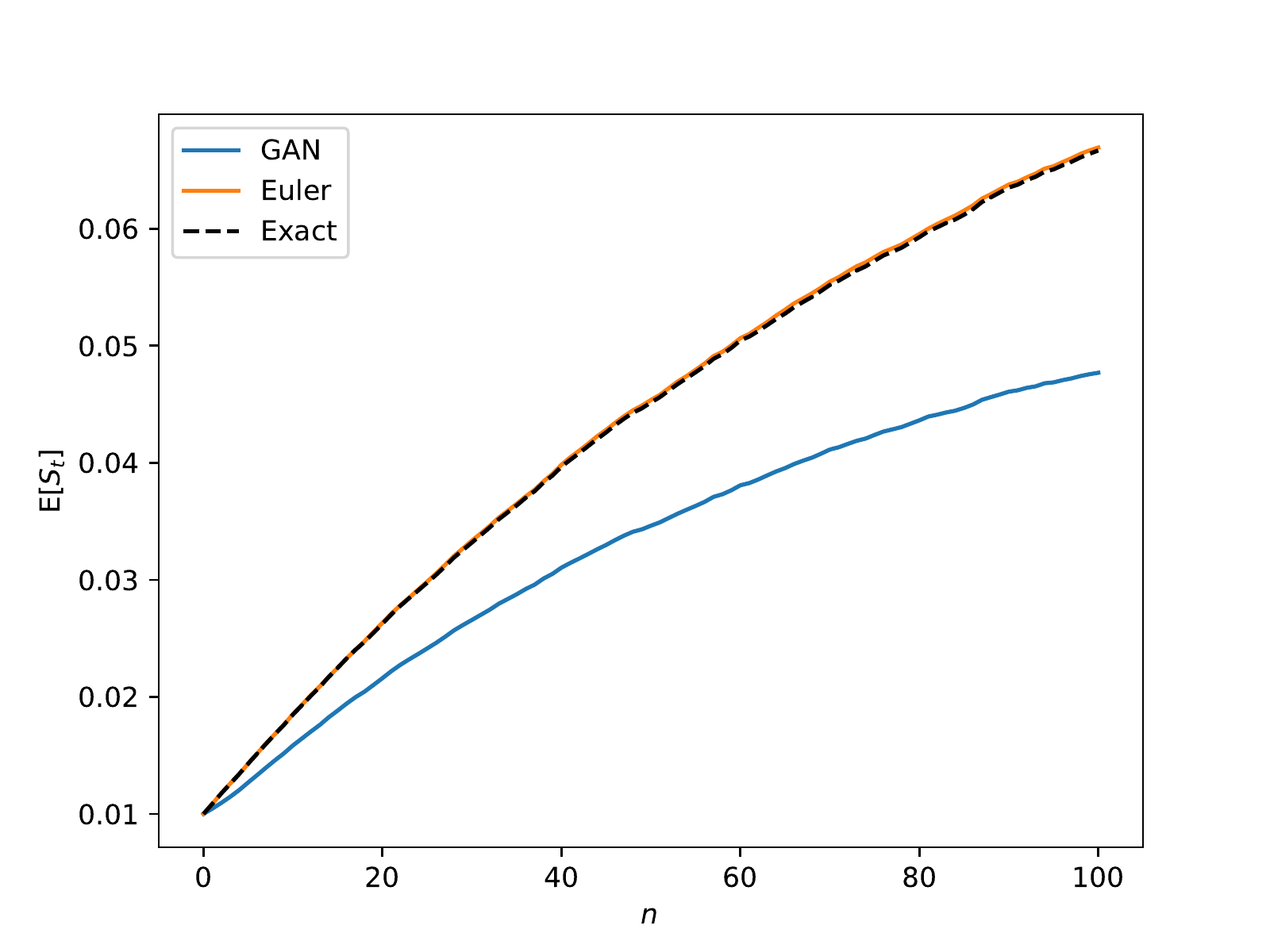}
        \caption{$\Delta t=0.1$}
        \label{fig:Feller_not_mean_of_paths_S0_0_01_t_0_10}
    \end{subfigure}
    \begin{subfigure}{0.49\linewidth}
        \centering
        \includegraphics[width=\linewidth]{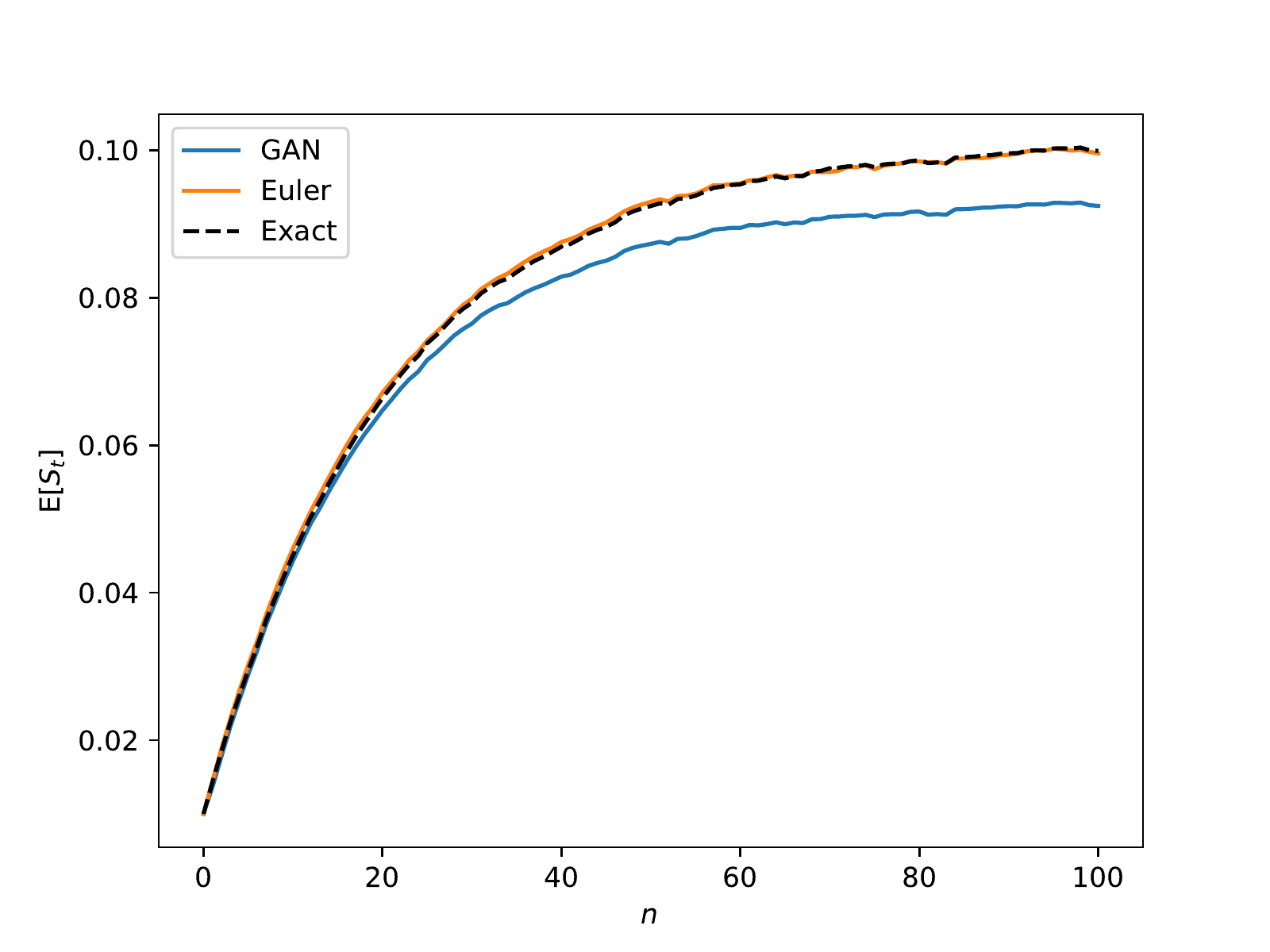}
        \caption{$\Delta t=0.5$}
        \label{fig:Feller_not_mean_of_paths_S0_0_01_t_0_50}
    \end{subfigure}
    \begin{subfigure}{0.49\linewidth}
        \centering
        \includegraphics[width=\linewidth]{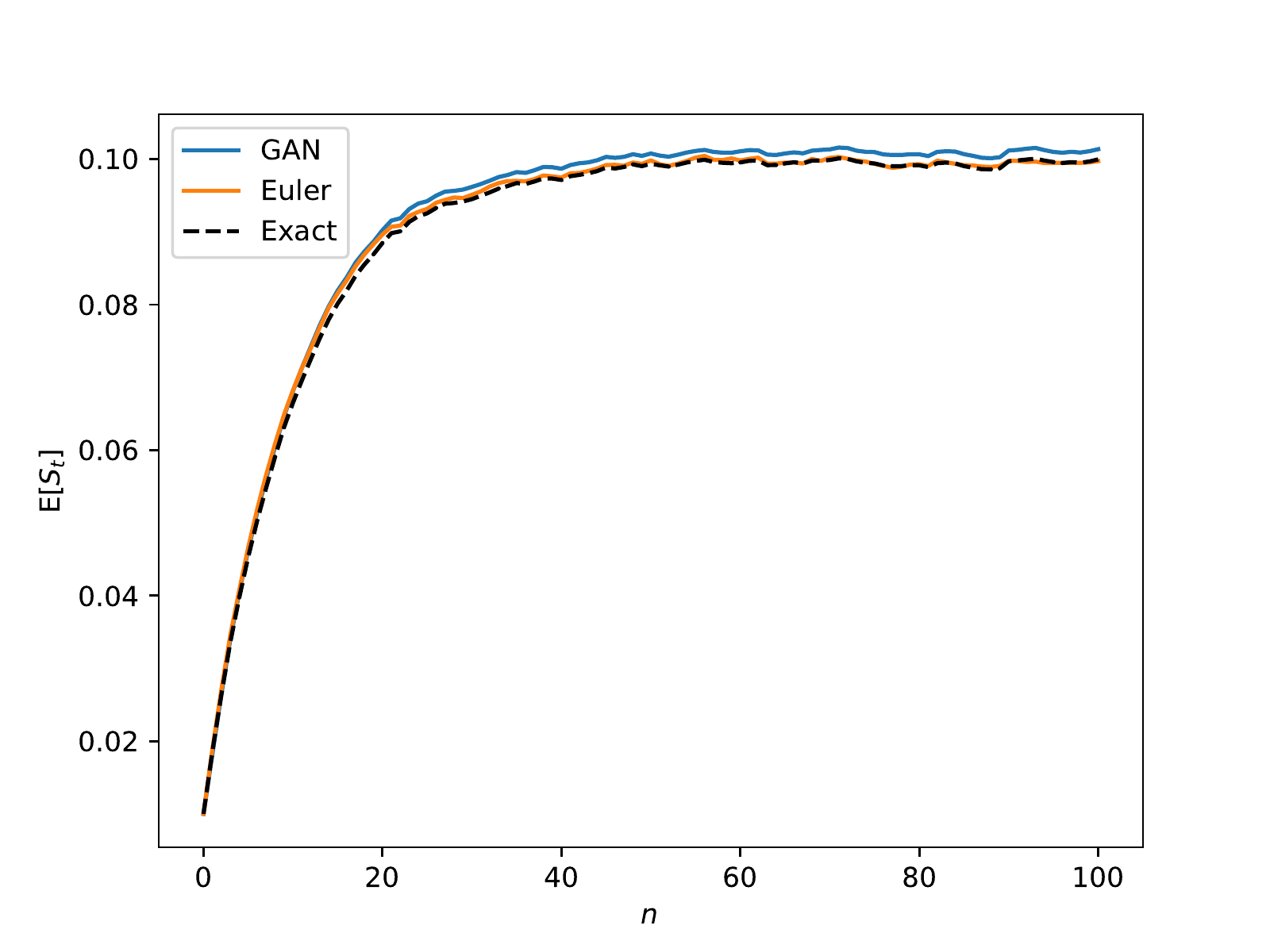}
        \caption{$\Delta t=1$}
        \label{fig:Feller_not_mean_of_paths_S0_0_01_t_1_00}
    \end{subfigure}
    \caption{Mean of $10^5$ paths obtained with the supervised GAN after $n$ repetitions of $G_{\theta}(Z,S_t,\Delta t)$, starting from $S_0=0.01$. The mean reversion parameter was set to $\bar{S}=0.1$. The paths generated by the supervised GAN indeed exhibit mean reversion, although the GAN does not revert to the correct mean for every $\Delta t$. As $\Delta t$ decreases, the error in the mean to which the GAN samples revert increases. This shows that the approximation of the conditional distribution is less accurate for smaller $\Delta t$, in line with our remaining benchmarks.}
    \label{fig:Feller_not_mean_of_paths}
\end{figure}

Figures \ref{fig:ECDF_Feller_not_Delta_t} and \ref{fig:benchmark_N_Feller_not_t_0_4} provide a `snapshot' of the output of both GANs for one or more conditional inputs. For the CIR process, we can test a qualitative property that requires the GAN to accurately capture the conditional dependence on $\Delta t$ and $S_t$. We show this for the supervised GAN. The CIR process reverts in the mean to the parameter $\bar{S}$ with increasing $t$ at a rate defined by $\kappa$. We can test this property by sampling $N$ values of $S_{t+\Delta t}\mid S_t$ repeatedly (e.g.\ 100 times) and taking the mean over all paths. The simulated process should converge in mean to $\bar{S}$. The result is shown in Figure \ref{fig:Feller_not_mean_of_paths}. The GAN indeed appears to revert to a mean, although it does not revert to the correct mean for each $\Delta t$. For lower values of $\Delta t$, the mean to which the GAN reverts is not equal to $\bar{S}$, which indicates that the distribution is captured less accurately than at higher values of $\Delta t$. Note that this experiment `stress-tests' the iterative sampling method, as it is repeated 100 times, allowing errors to accumulate. In practice, one most likely only repeats the GAN output several times on the previous output. However, if many repeated samples are of interest, the architecture should be extended to include the possibilities for online corrections along the path. 

\subsection{Weak and strong error}
\label{subseq:weak_strong_error}
\begin{figure}[h!]
\centering
    \begin{subfigure}{0.49\linewidth}
        \centering
        \includegraphics[width=\linewidth]{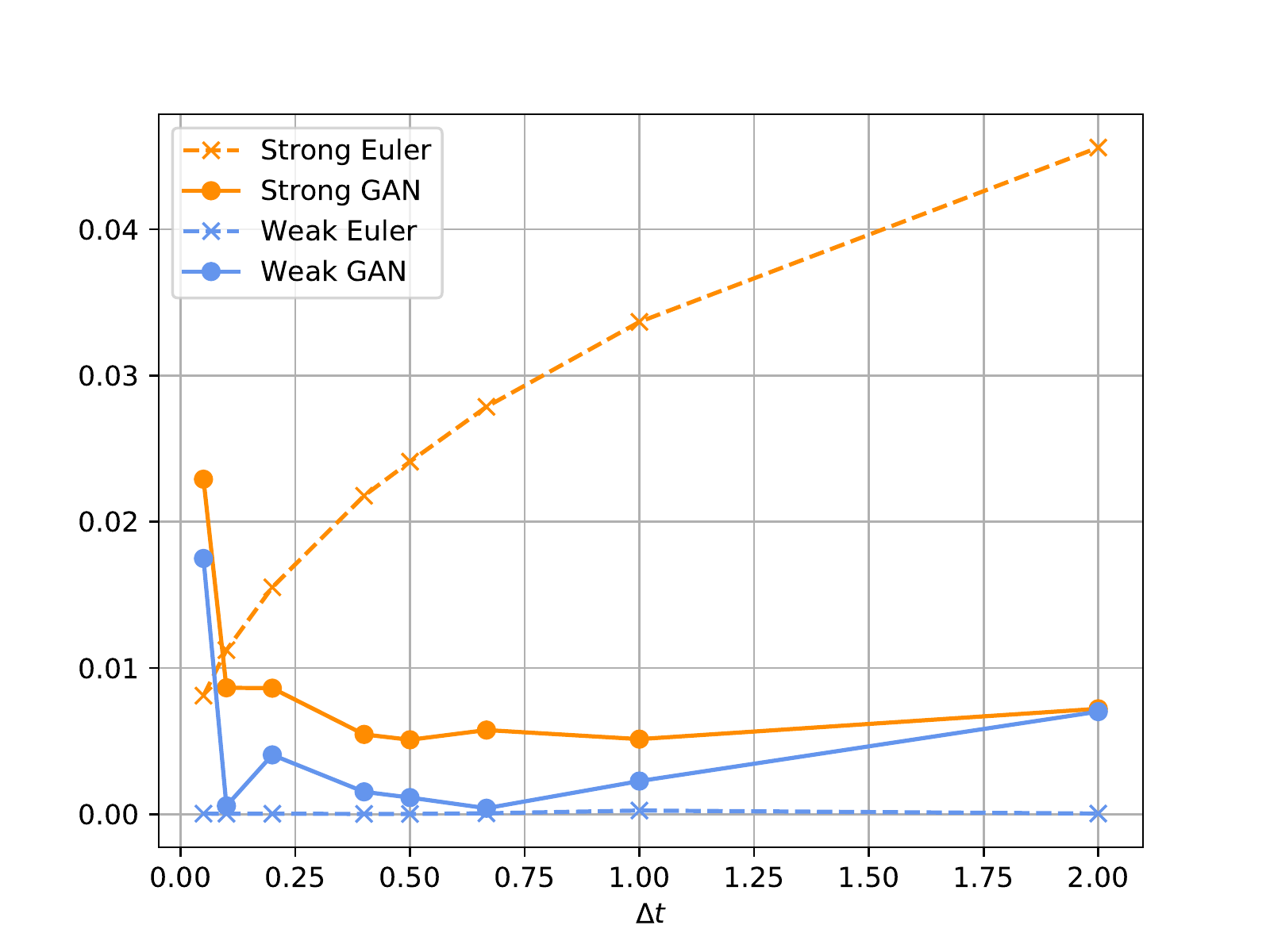}
        \caption{$S_0=0.1$, vanilla}
        \label{fig:Feller_not_weak_strong_S0_0_1_vanilla}
    \end{subfigure}
    \begin{subfigure}{0.49\linewidth}
        \centering
        \includegraphics[width=\linewidth]{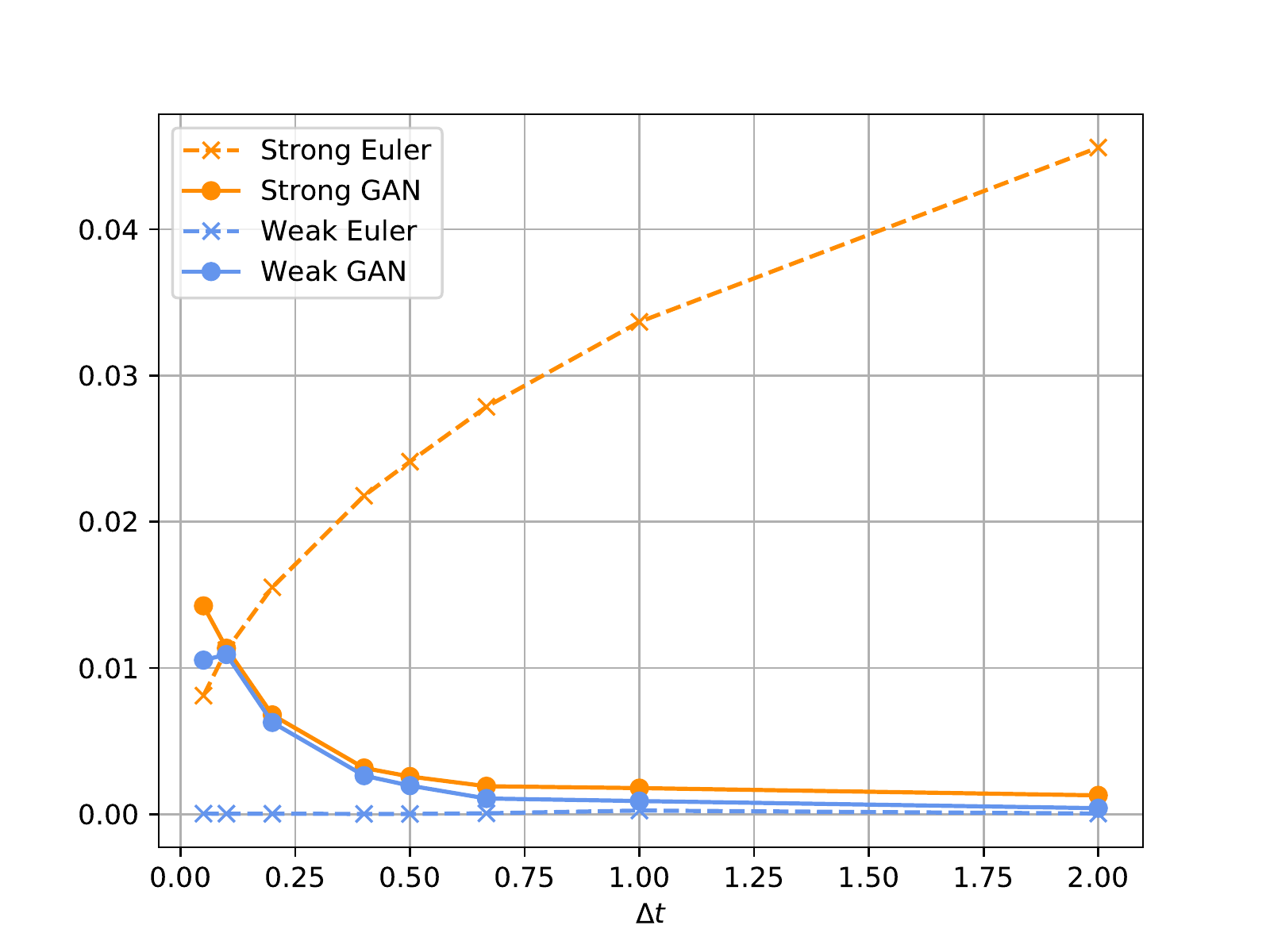}
        \caption{$S_0=0.1$, supervised}
        \label{fig:Feller_not_weak_strong_S0_0_1_constrained}
    \end{subfigure}
    \begin{subfigure}{0.49\linewidth}
        \centering
        \includegraphics[width=\linewidth]{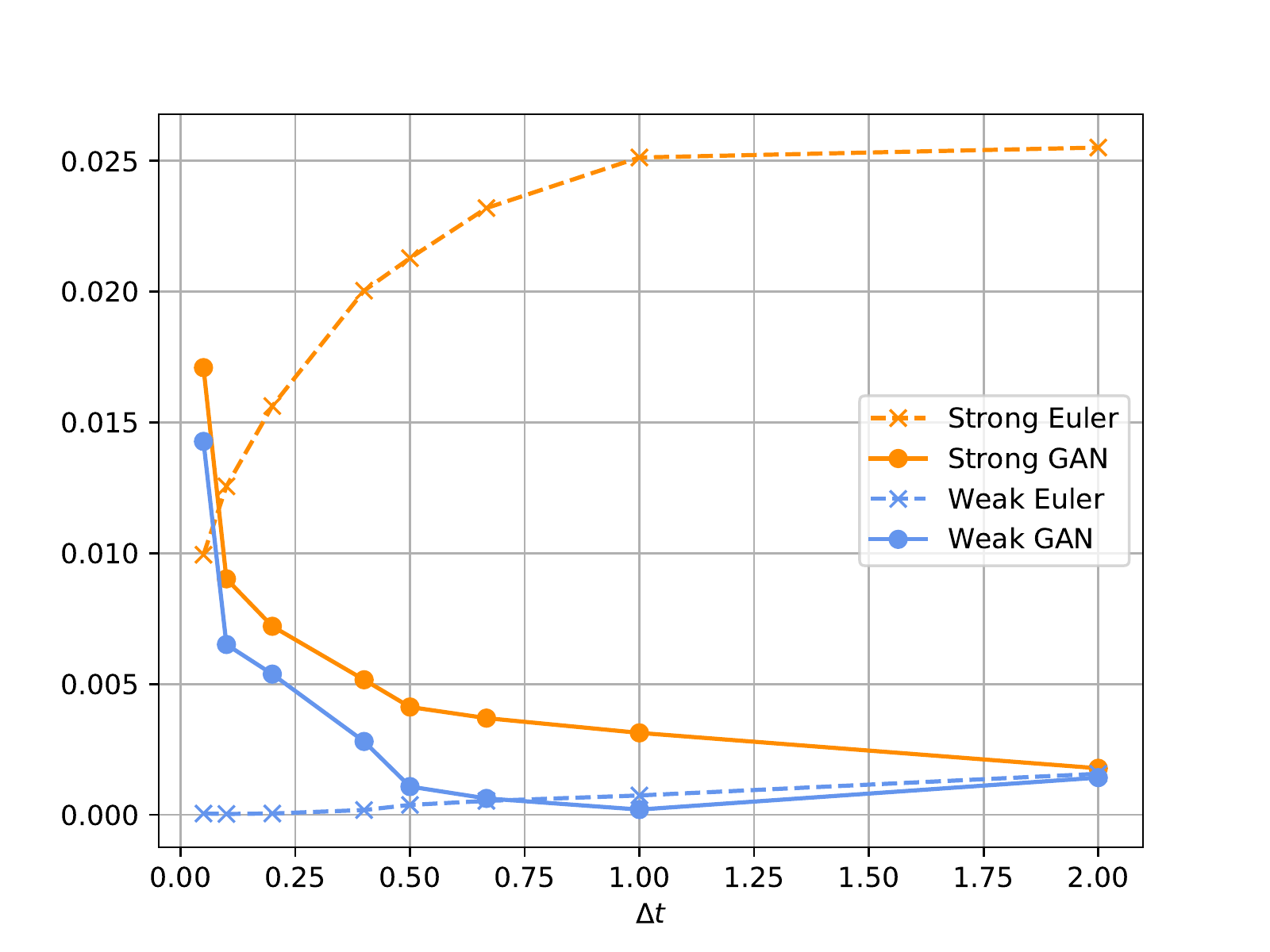}
        \caption{$S_0=0.01$, vanilla}
        \label{fig:Feller_not_weak_strong_S0_0_01_vanilla}
    \end{subfigure}
    \begin{subfigure}{0.49\linewidth}
        \centering
        \includegraphics[width=\linewidth]{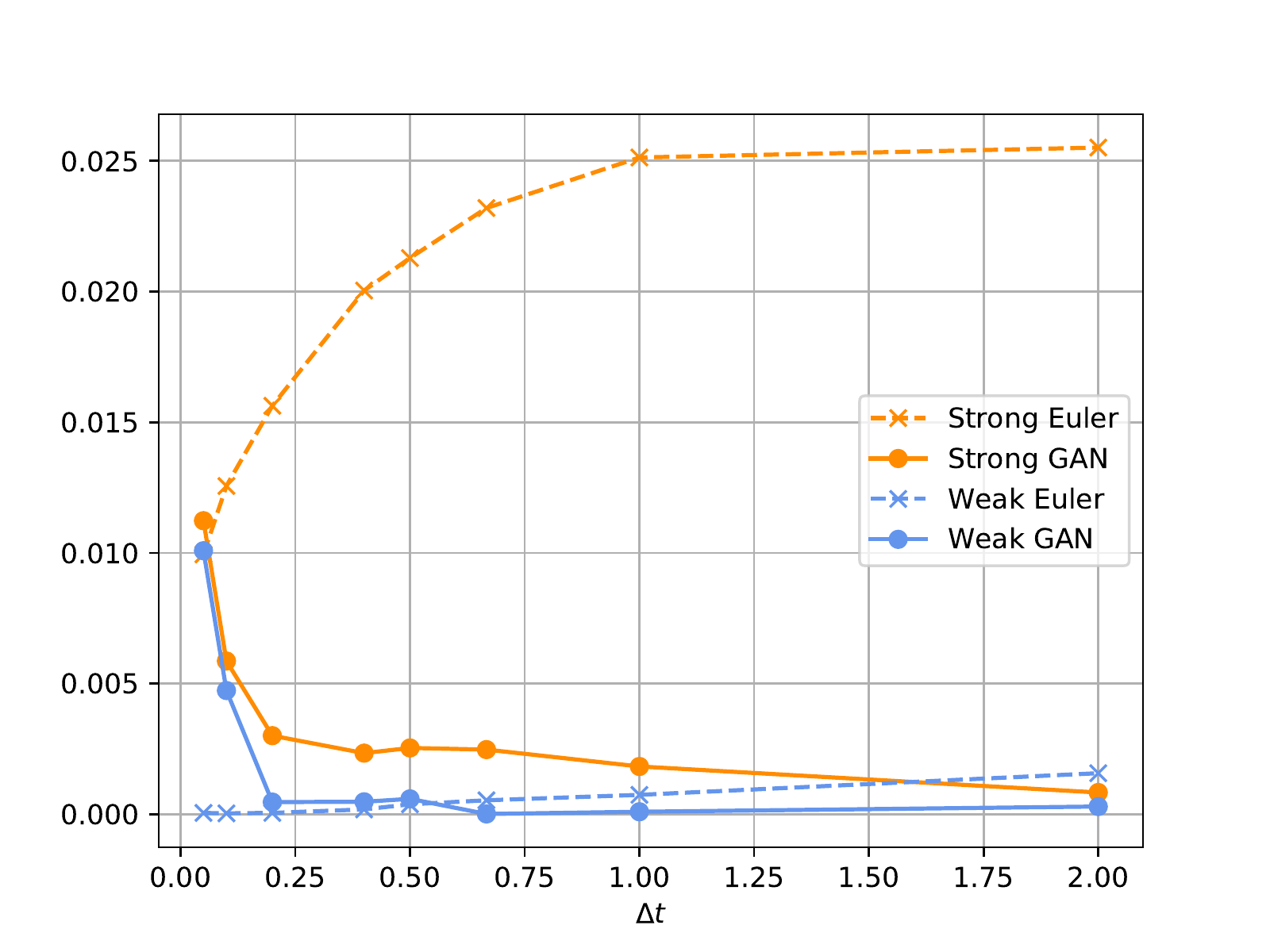}
        \caption{$S_0=0.01$, supervised}
        \label{fig:Feller_not_weak_strong_S0_0_01_constrained}
    \end{subfigure}    
    \caption{Weak and strong errors of artificial paths obtained with the vanilla and supervised GAN. In both cases, the strong error outperforms the discrete-time schemes even at low values of $\Delta t$, suggesting that both GANs have learned a strong approximation. However, the vanilla GAN did not manage to find a strong approximation on all test problems (see e.g.\ Figure \ref{fig:grand_figure_RvsZ_1}).}
    \label{fig:Feller_not_weak_strong}
\end{figure}

We iteratively sample from the process ${\hat{S}_{t+\Delta t}\mid S_t}$ with both the vanilla GAN and supervised GAN, on a discretisation $\{0,\Delta t,\hdots,N\Delta t\}$ with $\Delta t=\frac{T}{N}$ and $T=2$. The input to both GANs, $Z\sim N(0,1)$, is stored at each time point and re-used for the Euler and Milstein approximation. Note that if we chose a different $Z$ for the discrete-time schemes, we could not compare the results path-wise. The experiment is repeated for $N \in \{40,20,10,5,4,3,2,1\}$ steps, yielding different choices of $\Delta t$. Using this setup, 100,000 paths were generated for each choice of $N$ and the weak and strong error have been plotted versus $\Delta t$ in Figure \ref{fig:Feller_not_weak_strong}. 

\begin{figure}[h!]
\centering
    \begin{subfigure}{0.49\linewidth}
        \centering
        \includegraphics[width=\linewidth]{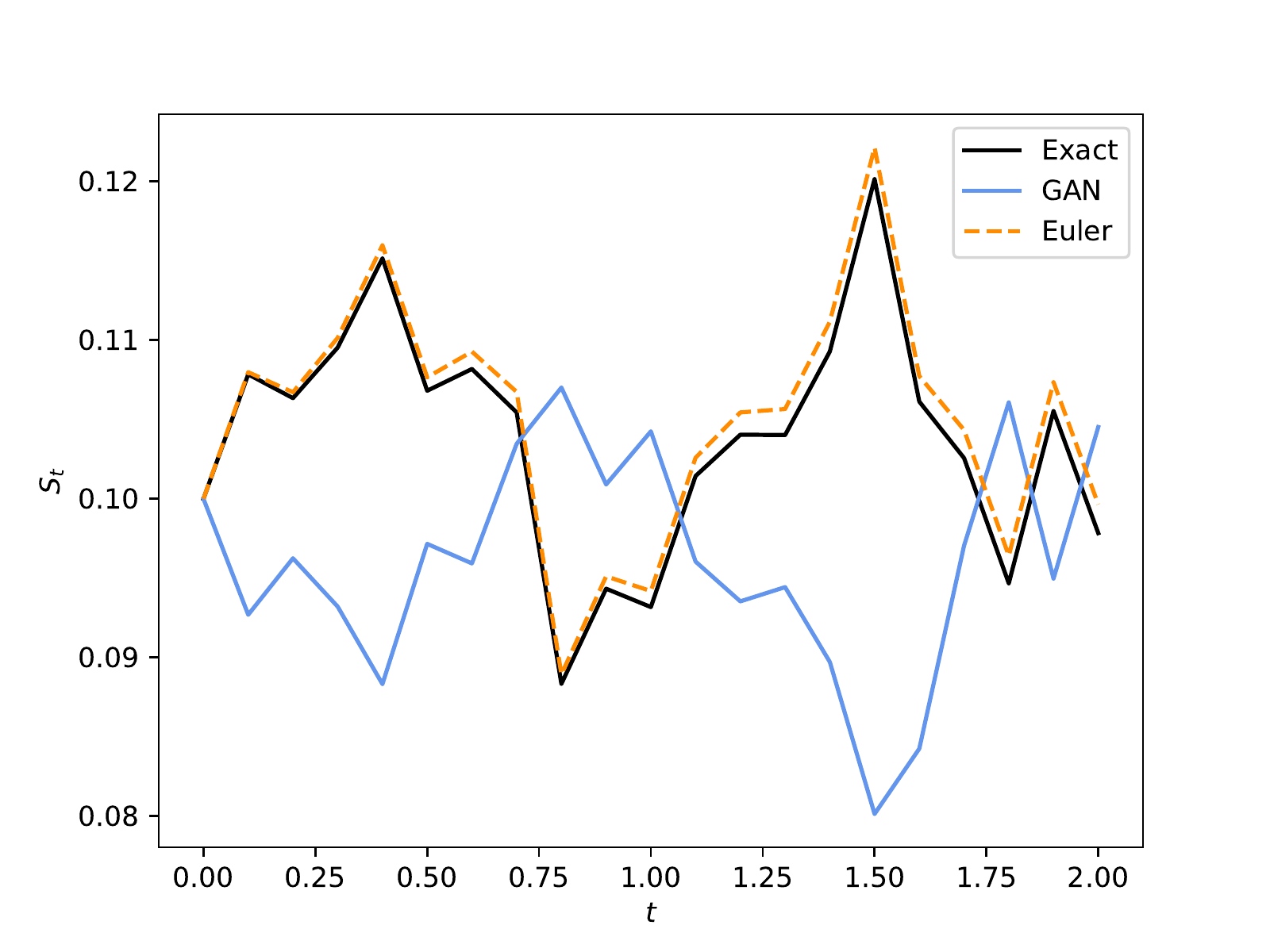}
        \caption{Vanilla GAN path}
        \label{fig:CIR_example_path_0}
    \end{subfigure}
    \begin{subfigure}{0.49\linewidth}
        \centering
        \includegraphics[width=\linewidth]{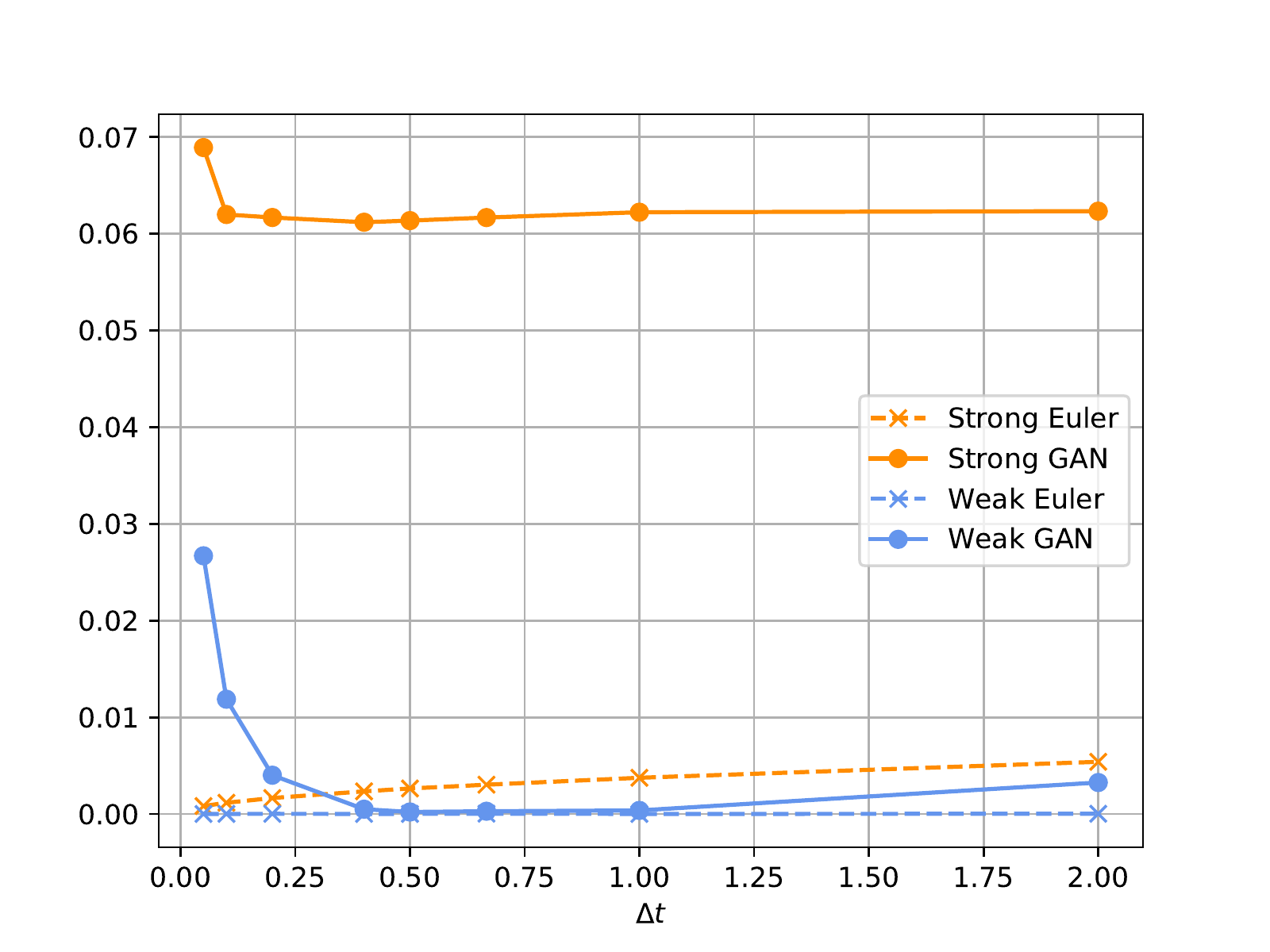}
        \caption{Vanilla GAN: $e_w$ and $e_s$}
        \label{fig:CIR_weak_strong_unconstrained}
    \end{subfigure}
    \begin{subfigure}{0.49\linewidth}
        \centering
        \includegraphics[width=\linewidth]{img/Paths/CIR_Feller_not_Constrained_S0_0_01_weak_strong_plot.pdf}
        \caption{Supervised GAN: $e_w$ and $e_s$}
        \label{fig:CIR_weak_strong_constrained}
    \end{subfigure}
    \caption{Example of failure of the vanilla GAN to provide a strong approximation on the CIR process if the Feller condition was satisfied. It fails to converge path-wise (Figure \ref{fig:CIR_example_path_0}), which is reflected in the strong error (Figure \ref{fig:CIR_weak_strong_unconstrained}). $S_0$ was set to $0.1$.}
    \label{fig:example_failure_vanilla_weak_strong}
\end{figure}

When the Feller condition was not satisfied, the modified Milstein scheme did not perform better than the modified Euler scheme, which is why it was left out of this experiment. Both GANs yield a lower strong error even at small values of $\Delta t$ across all three figures, which suggests that both GANs provide a strong approximation. For the vanilla GAN, this is a special case, as we see in Figure \ref{fig:example_failure_vanilla_weak_strong}, in which we provide an example if the Feller condition is satisfied. We study this phenomenon in more detail in the succeeding paragraph. The weak error in figures \ref{fig:Feller_not_weak_strong_S0_0_1_vanilla} and \ref{fig:Feller_not_weak_strong_S0_0_1_constrained} on this problem is relatively high compared to the Euler scheme, which can be explained by the choice of parameters in this experiment. The mean reversion parameter $\bar{S}$ was set to $0.1$, which is equal to $S_0$. This means that the Euler scheme starts at exactly the correct mean from time $t_0$. If we change $S_0$ to $0.01$, the supervised GAN also outperforms the Euler scheme in weak error at approximately $\Delta t$ greater than $0.5$. The performance of both GANs is not uniform in $\Delta t$, which is particularly pronounced at low values. The opposite is true for the Euler scheme, which becomes increasingly accurate for decreasing $\Delta t$. Note that for an ideal GAN, the weak and strong error would not depend on $\Delta t$. 

\subsection{Map learned by vanilla and supervised GAN}

We now test the reasoning in Section \ref{subsec:GAN_as_parametric_map} empirically and study the map learned by both GAN architectures, i.e.\ the output $\big(S_{t+\Delta t}\mid S_t\big) (\omega)$ given an input $Z(\omega) \sim N(0,1)$ for an event $\omega \in \mathcal{F}_t$. Instead of $S_{t+\Delta t}\mid S_t$, we plot the output of the pre-processed data $R_t$ on which the GANs were trained, i.e.\ logreturns for GBM and CIR with Feller condition violated, scaling with $\bar{S}$ if the Feller condition is violated. In Figure \ref{fig:grand_figure_RvsZ_1}, we show three different examples of the vanilla GAN failing to provide a strong approximation, although the approximation of the distribution is close.  

\def\triplewidth{0.32}

\begin{figure}[h!]
\centering
    \begin{subfigure}{\triplewidth\linewidth}
        \centering
        \includegraphics[width=\linewidth]{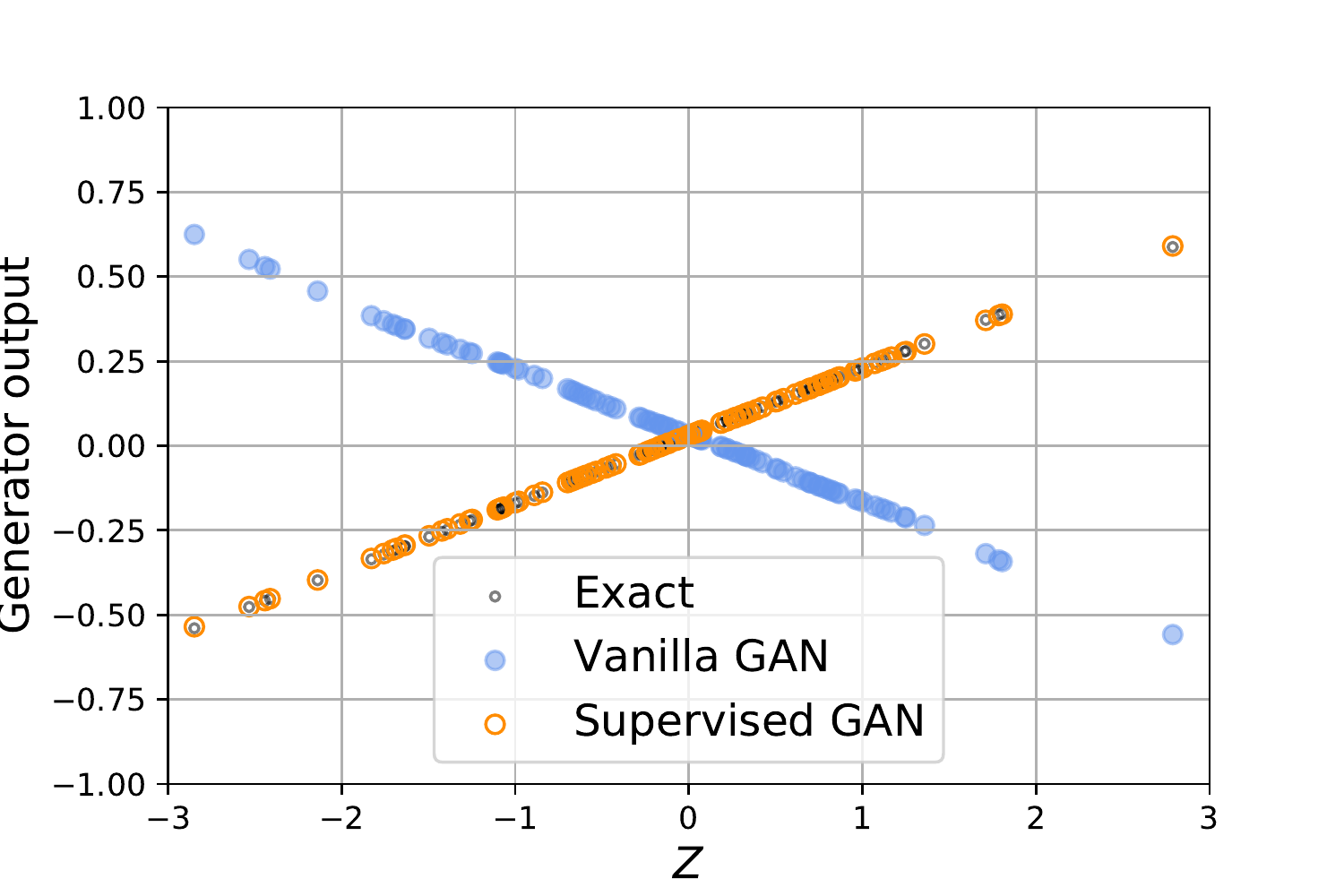}
    \end{subfigure}
    \begin{subfigure}{\triplewidth\linewidth}
        \centering
        \includegraphics[width=\linewidth]{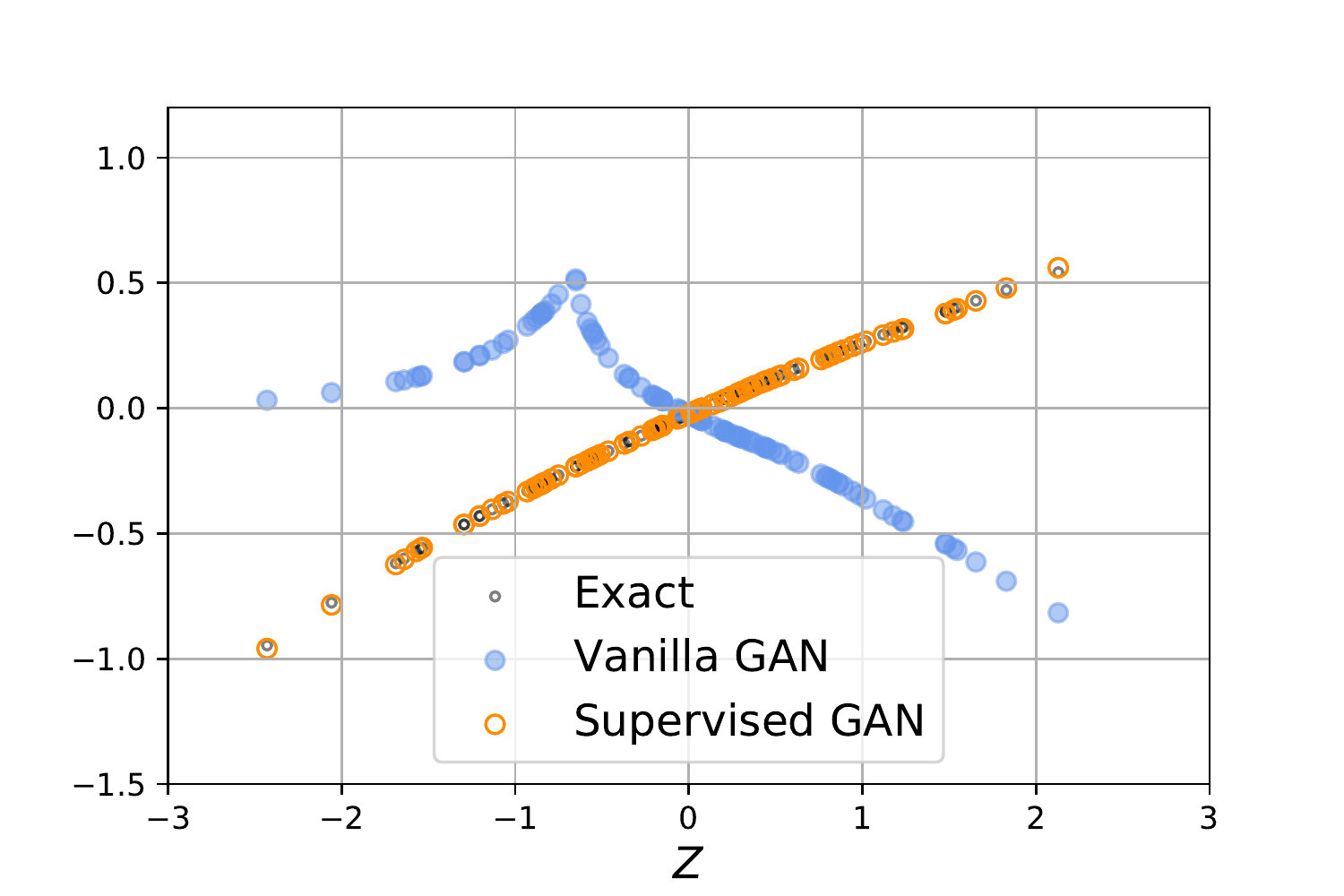}
    \end{subfigure}
    \begin{subfigure}{\triplewidth\linewidth}
        \centering
        \includegraphics[width=\linewidth]{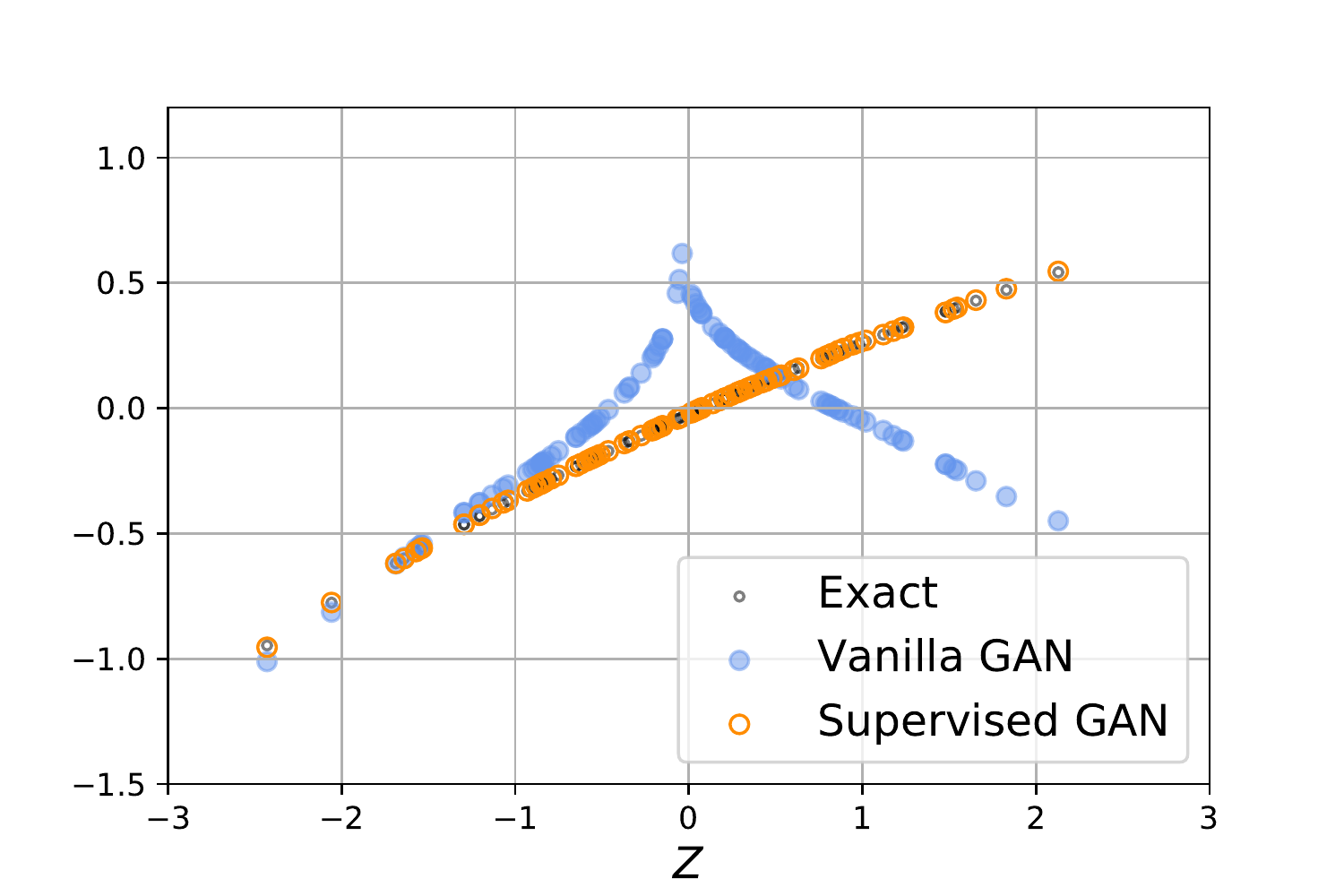}
    \end{subfigure}
    \begin{subfigure}{\triplewidth\linewidth}
        \centering
        \includegraphics[width=\linewidth]{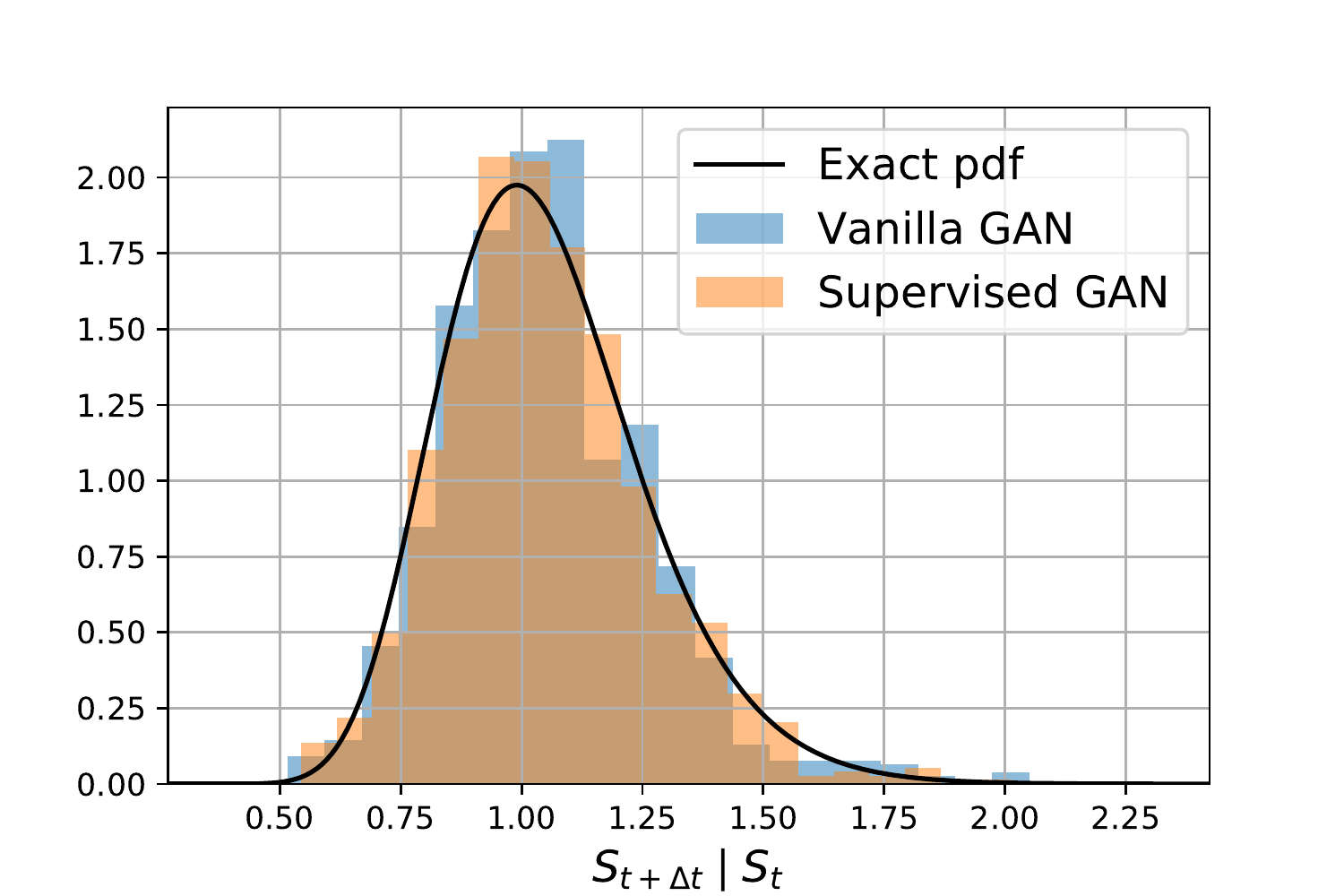}
        \caption{GBM}
        \label{fig:grand_figure_RvsZ_1_left}
    \end{subfigure}
    \begin{subfigure}{\triplewidth\linewidth}
        \centering
        \includegraphics[width=\linewidth]{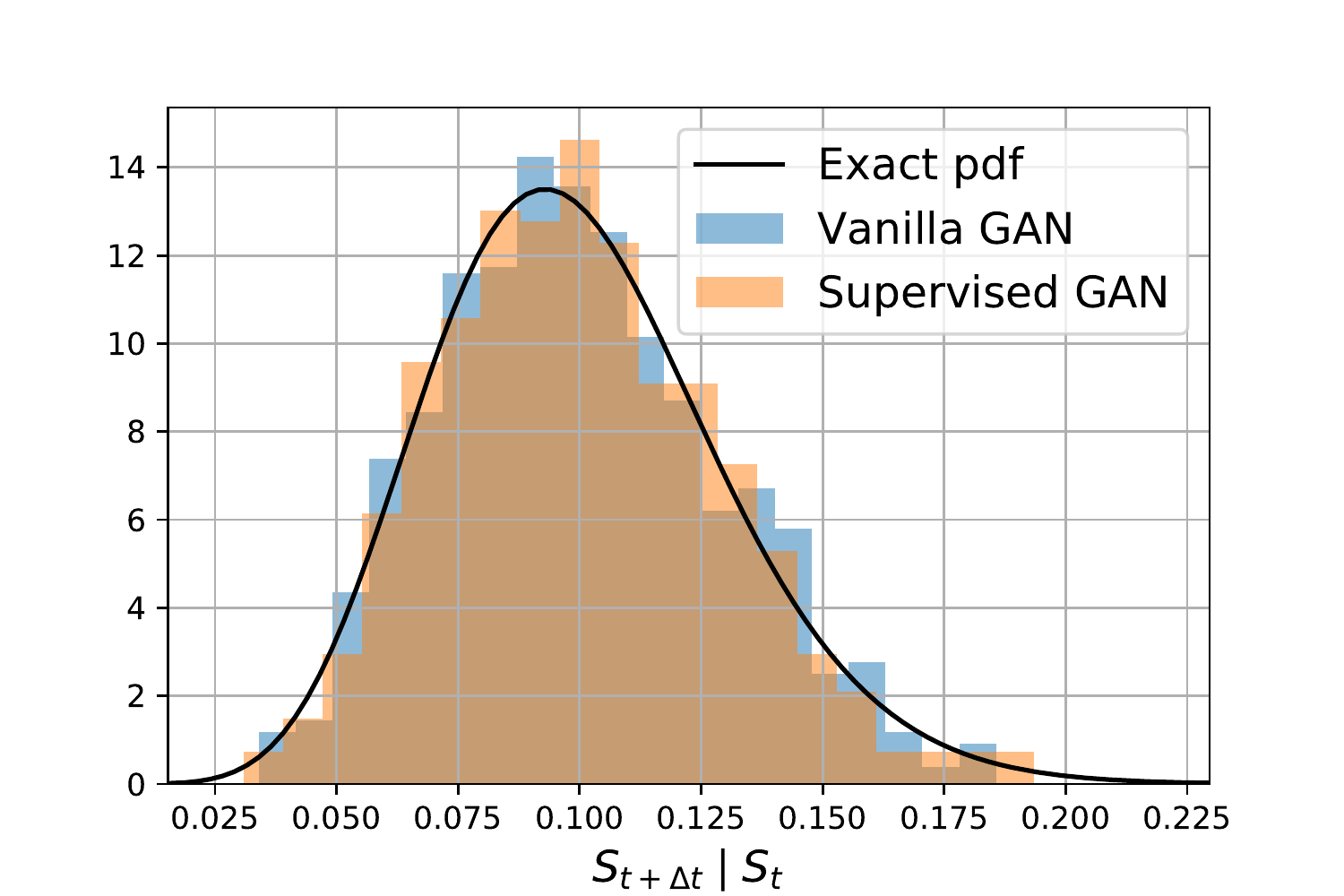}
        \caption{GANs from Figure \ref{fig:CIR_example_path_0}}
        \label{fig:grand_figure_RvsZ_1_middle}
    \end{subfigure}
    \begin{subfigure}{\triplewidth\linewidth}
        \centering
        \includegraphics[width=\linewidth]{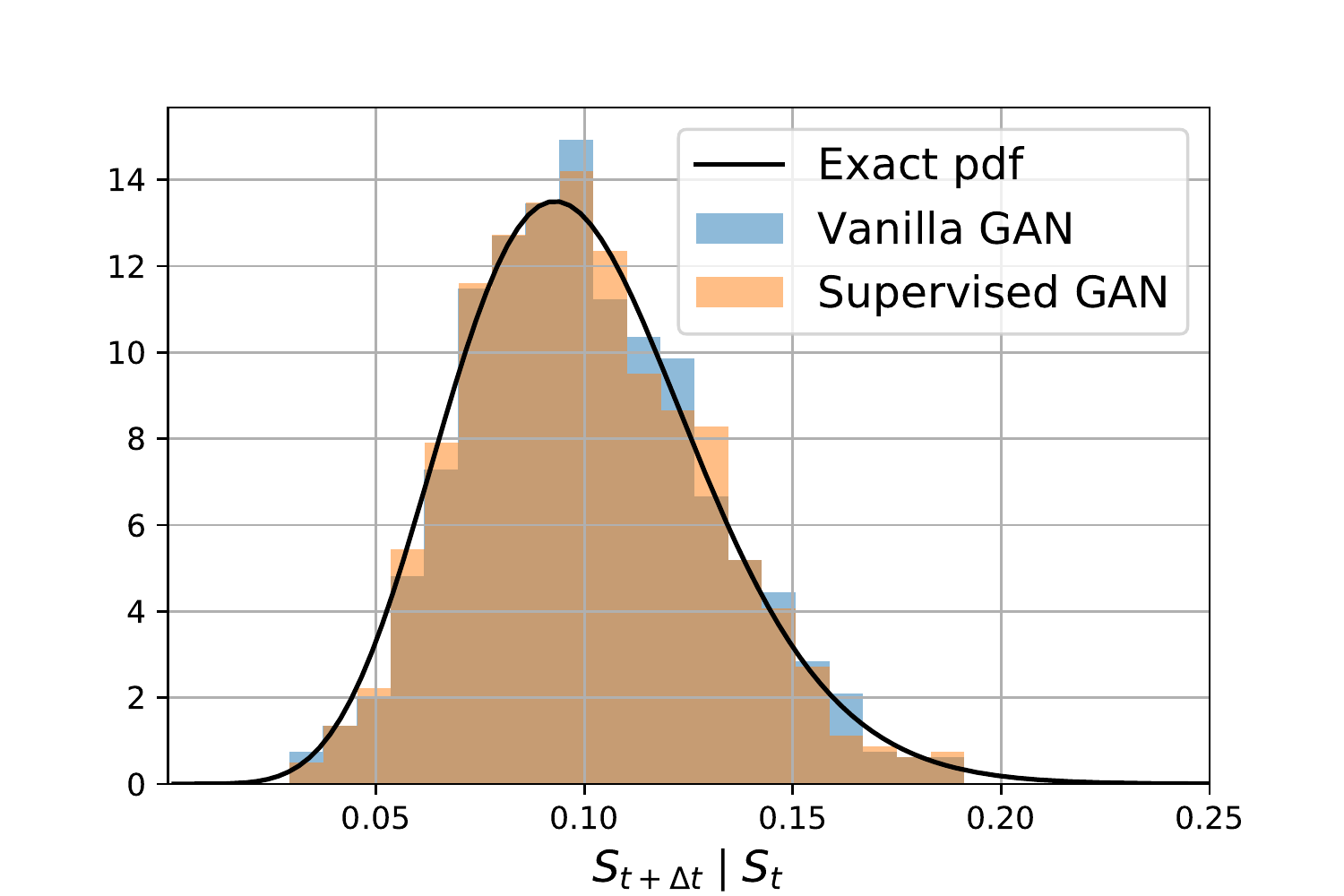}
        \caption{Feller satisfied, second example}
        \label{fig:grand_figure_RvsZ_1_right}
    \end{subfigure}    
    \caption{Top row: the map $Z\mapsto G_{\theta}(Z,S_t,\Delta t)$ with $S_t=0.1$ and $\Delta t=1$ for three different examples. Each figure shows a scatter plot of the generator output of both GANs on the same 100 input samples $Z$. Bottom row: corresponding histograms of based on 1000 input samples $Z$.}
    \label{fig:grand_figure_RvsZ_1}
\end{figure}

Each of the examples in Figure \ref{fig:grand_figure_RvsZ_1} gives rise to different pathological behaviour on the side of the vanilla GAN. The map on the left corresponds to `mirrored' paths compared to the strong solution, corresponding to the weakly unique `twin' solution to the strong solution with $Z \leftarrow -Z$, which is equal in distribution, but not path-wise. This is exactly the $\varphi^{-}$ from Example \ref{ex:lognormal}. Note how the logreturns transformation makes the GBM problem trivial: the conditional GAN learns the slope and intercept of a straight line. In the centre and right figure, the vanilla GAN has not simply learnt a weakly unique solution with opposite sign, it has learned a map that gives rise to a similar distribution as the reference, but corresponds to an entirely different map from $Z$ to the GAN output. This would correspond to a map that generates samples within some $\varepsilon$ from the target distribution, but where $\varphi_{\theta}$ itself is very different from $\varphi^+$ and $\varphi^-$, as we discussed in Equation \eqref{eq:JS_eps}. This again led to the paths not being equal path-wise to the strong solution. The maps in the centre and rightmost figures are not bijective, since there are two returns for some inputs $Z$. Furthermore, the rightmost example shows that the vanilla GAN may be highly sensitive to small changes in the input. E.g. for $Z$ around $0$, the output can change very rapidly for a small perturbation in $Z$. \par 

In all experiments performed in preparation for this work, the supervised GAN was able to provide a strong approximation, which is visible in Figure \ref{fig:grand_figure_RvsZ_1} by the orange data points completely overlapping with the exact samples. The supervised GAN thus learns the map corresponding to the inverse function $F^{-1}_{S_{t+\Delta t}\mid S_t}(F_Z(Z))$. 

\subsection{Supervised GAN discriminator output}
\begin{figure}[h]
    \centering
    \includegraphics[width=1\linewidth]{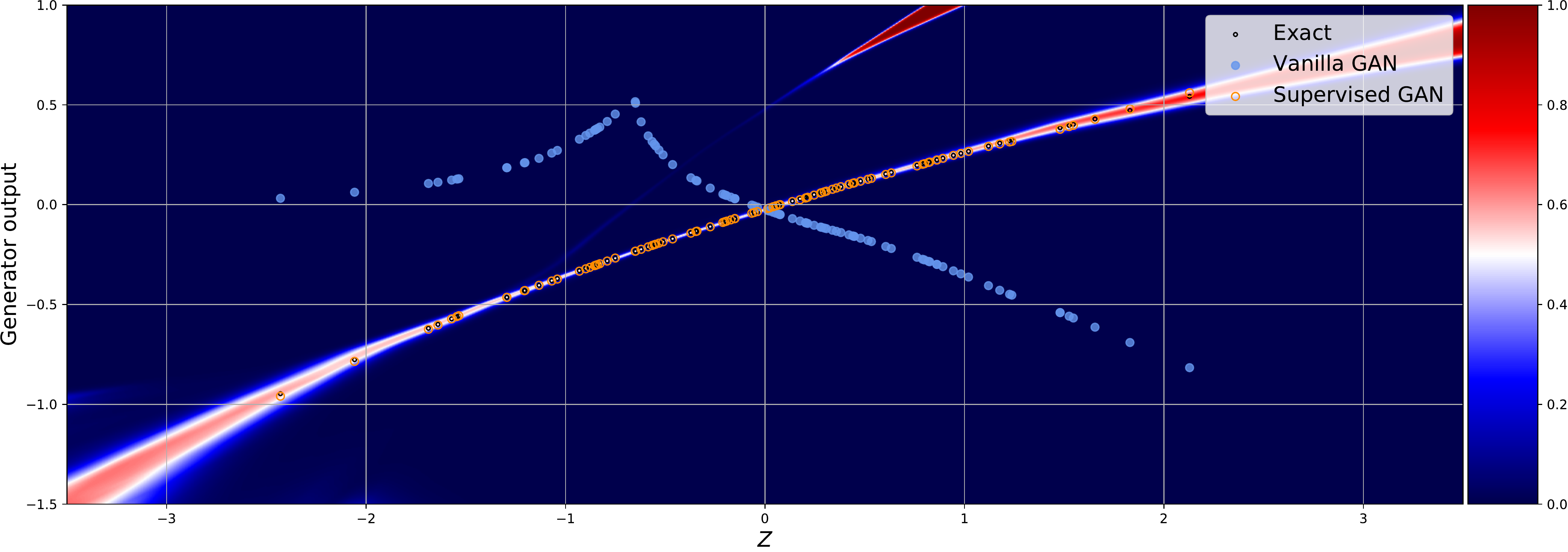}
    \caption{Discriminator output corresponding to figures \ref{fig:CIR_example_path_0} and \ref{fig:grand_figure_RvsZ_1_middle} output after 40,000 training steps for varying $Z$ and  $G_{\theta}(Z,S_t,\Delta t)$, with fixed $S_t=0.1$ and $\Delta t=1.0$. The discriminator identifies the region in which the exact samples lie for each combination of $(Z,\Delta t,S_t)$.}
    \label{fig:D_conf_example_1}
\end{figure}

We can visualise how the supervised GAN learns by visualising the discriminator output and overlay the generator output. This way, we show explicitly how the discriminator scores each input sample. The generator is given by the function $G_{\theta}:\mathbb{R}^{3}\rightarrow \mathbb{R}$ with input $(Z,S_t,\Delta t)$, while the discriminator is given by $D_{\alpha}:\mathbb{R}^4\rightarrow [0,1]$ with inputs $((S_{t+\Delta t}\mid S_t),Z,S_t,\Delta t)$. If we fix $S_t$ and $\Delta t$, say at $0.1$ and $1.0$, respectively, we can visualise the discriminator output on $[0,1]$ with a colormap on the space $(Z, G_{\theta}(Z,S_t,\Delta t))$, which is shown in Figure \ref{fig:D_conf_example_1}. 

Upon convergence of the GAN, the discriminator output will be around 0.5 in a neighbourhood of the generated samples, as it is no longer able to distinguish between the reference and generated samples. This corresponds to the `white band' in Figure \ref{fig:D_conf_example_1}, which is exactly where the exact samples and supervised GAN samples can be found. If the vanilla GAN samples would have been provided to the supervised GAN discriminator, it would have classified all samples as fake, as is visible in the figure by the fact that all the vanilla GAN data points lie in the dark blue region. This shows how the supervised GAN discriminator rules out any map other than the strong solution.

\section{Discussion}
\label{sec:Discussion}

\bfquad{Supervised learning} One could argue that GANs are not needed to solve our problem, since the map $Z\mapsto \varphi_{\theta}(Z)$ could have equally been trained using only a `generator' combined with an $L_2$ loss. This is possible since we had available the underlying map $F_Z^{-1} \circ F_{S_{t+\Delta t}\mid S_t} $ and were able to build a training set with examples of $((S_{t+\Delta t}\mid S_t),Z)$. However, using a supervised variant of the GAN as a reference model allowed us to compare both GAN-architectures directly, using the same learning algorithm. \par 

\bfquad{Beyond GBM and the CIR process} For general 1D \Ito\ SDEs, where $F_{S_{t+\Delta t}\mid S_t}$ is not available analytically, one could use an empirical analogue instead, without any changes to the supervised GAN architecture. The only requirement is that the empirical approximation should be strictly increasing in order to find a unique $Z$ for each data sample, which could be achieved e.g.\ with a non-decreasing interpolation scheme between the data points defining the ECDF. For higher dimensional SDEs, the prior input to the generator should be increased for each degree of freedom. If the Brownian motions are correlated, they can be written as a product covariance matrix and a vector independent Brownian motions, using Cholesky decomposition \cite{oosterlee2019_book}. The covariance matrix would be a function of the correlation coefficients of each of the correlated Brownian motions. A conditional GAN could then be trained, with correlation parameters $\rho_1,\rho_2,\hdots$ as an additional conditional input. \par 

\bfquad{Large time steps} We showed that the supervised GAN is able to approximate the conditional distribution accurately for large time steps. `Large' here meant large compared with a discrete-time approximation, which we used to benchmark our results. However, this may be considered unfair, since time steps of e.g.\ 1,2 are unrealistic for discrete-time schemes. On the other hand, the supervised GAN outperformed the discrete-time schemes on time steps below 1 as well, only struggling with the smallest of time steps. The comparison was sufficient to show that the supervised GAN is able to approximate the target SDE path-wise. \par 

\bfquad{Data pre-processing} On all benchmarks, performance of both GANs decreased the lower we chose $\Delta t$. This may seem counter-intuitive, as discrete-time schemes improve with decreasing $\Delta t$. However, since our model approximates an exact simulation scheme, the accuracy should theoretically not depend on $\Delta t$ at all. The dependence of performance on $\Delta t$ reflects the ability of the GAN to approximate the target distribution conditional on $\Delta t$. Neural networks tend to learn slower on input samples with lower variance \cite{lecun1998neural}. This is because the gradient update for each weight scales with the variance of the input samples. Although the data were pre-processed by taking logreturns or scaling with $\bar{S}$, the \textit{in-class} variance is still non-constant. The more conditional classes are added that affect the variance, the more pronounced this result would be. An example would be if the parameter $\gamma$ from the CIR process were added as an additional conditional input. \par 

One way to counter the in-class variance would be to standardise each class individually. However, the post-processing step would then require knowledge of the mean and variance of the training set batches. A different route may be through scaling each training point with its corresponding $\Delta t$ and $S_0$. However, in order to achieve unit variance, one would need very specific knowledge of the output distribution, which may be restrictive. Additionally, the heavy tails in the distributions make traditional standardisation techniques ineffective. \par 

\bfquad{Full parameter range} The conditional GAN architecture for modelling the conditional distribution could be further generalised to include the full parameter set of the SDE, allowing the GAN to learn an entire family of SDEs at once, as is done in \cite{liu2020sevenleague} for the SCMC implementation. In this work, we developed a conditional GAN that was sufficient to demonstrate path-wise convergence. In future work, it would be interesting to test the GAN on the full parameter range of SDEs as well, if the challenge of pre-processing the data without incorporating knowledge of the target distribution could be resolved. \par 

\section{Conclusion and Outlook}
\label{sec:Conclusion}
We proposed a GAN-based architecture for exact simulation of \Ito\ SDEs. Specifically, we approximated the conditional probability distribution of 1D geometric Brownian motion (GBM) and the Cox-Ingersoll-Ross (CIR) process with a conditional GAN. The GAN was conditioned on the time interval length and the preceding value along the path and was used to construct artificial asset paths by iterative sampling from the conditional distribution. We argued that for unsupervised generative models based on divergence measures, there are no guarantees about the input-output map learned by the neural network. This is because the network parameters are varied only to minimise a quantity such as the Jensen-Shannon divergence, but no restriction is applied on the underlying map. We demonstrated experimentally how this could lead to non-unique and non-parsimonious input-output maps by the generator. In the context of SDEs, we showed how this implies that the vanilla GAN is unable to reliably provide a strong approximation. We replaced the vanilla GAN by a supervised GAN, which learns how a random input maps to the target variable explicitly. This supervised GAN was able to provide a strong approximation in all cases. Additionally, the approximation in distribution by the supervised GAN was more accurate under identical learning parameters and network capacity. We see two main directions for future work. Firstly, our findings motivate users of generative models to study the input-output map learned by the model explicitly and verify qualitative properties such as smoothness. This aligns well with efforts to constrain the generator, such as the `potential flow generator' introduced in \cite{yang2020potential_flow_GAN}, that uses optimal transport to constrain the generator map. Secondly, our conditional GAN architecture could be further extended to include the SDE parameters as well, as is done for the `Seven-League' collocation sampler in \cite{liu2020sevenleague}. This would allow exact simulation of entire classes of SDEs instead of a specific choice of parameters. Since we showed how supervised learning can be used for \Ito\ SDEs, the GAN architecture itself can be replaced by a single generator, trained on e.g. the mean-squared error. Extensions of our architecture, along with the methods we used for studying the output may be applied on more general problems, such as higher dimensional SDEs or non-\Ito\ SDEs. 


\footnotesize

\section*{Declarations}
\textbf{Competing interests}: the authors declare that they have no conflict of interest. 

\bibliographystyle{unsrt}
\bibliography{References}

\vfill

\newpage 

\appendix 

\normalsize


\section*{Appendix}

\section{Network architectures}
\label{appendix:NN_architecture_training}

The architectures of the feed-forward neural networks of the generator and discriminator are shown in table \ref{tab:architectures}. $c$ equals the amount of conditional parameters of the conditional GAN. $c=1$ for GBM and $c=2$ for the CIR process. If the discriminator is informed with $Z$, i.e. the supervised GAN, the discriminator input is further increased by 1 for the input $Z$. The batch size was set to $1,000$. Batches were sampled uniformly with replacement from a training set of $10^5$ training samples. The GAN was trained for a fixed amount of 200 epochs. An additional learning rate schedule was created to stabilise GAN training. This schedule was used for the generator, where the learning rate was decreased by a factor $1.05$ every $n_{\mathrm{LR}}=500$ iterations. 

\begin{table}[h!]
    \centering
    \caption{Network architectures}
    \label{tab:architectures}
    \begin{tabular}{l l l l l l}
    \toprule
	&	\multicolumn{2}{l}{\textbf{Generator}}				&	\quad	&	\multicolumn{2}{l}{\textbf{Discriminator}}	\\ \midrule				
Optimiser	&	Adam		&		&	\quad	&	Adam		&		\\	
	&	\multicolumn{2}{l}{$lr=10^{-4},\beta_1=0.5,\beta_2=0.999$}				&	\quad	&	\multicolumn{2}{l}{$lr=5\times 10^{-4},\beta_1=0.5,\beta_2=0.999$}				\\	\midrule
	&			&		&	\quad	&			&		\\	
Layer	&	Size		&	Activation	&	\quad	&	Size		&	Activation	\\	\midrule
Input layer	&	1+$c$		&	LeakyReLU,  negative slope=0.1	&	\quad	&	1+$c$		&	LeakyReLU,  negative slope=0.1	\\	
Hidden layers 1-4	&	200		&	LeakyReLU,  negative slope=0.1	&	\quad	&	200		&	LeakyReLU,  negative slope=0.1	\\	
Output layer	&	1		&	None	&	\quad	&	1		&	Sigmoid	\\		   

\bottomrule
\end{tabular}
    
\end{table}

The learning rate for the discriminator was set to 5$\times$ that of the generator learning rate, which was found to lead to faster convergence in the first epochs and a better approximation of the optimal discriminator, cf. \cite[section 2]{biau2020some}. \par 
Saturating activation functions, such as a tanh or sigmoid, have also been considered. However, since the heavy left-tails of the target distributions persist after a pre-processing step (for the CIR process), the distribution of the hidden state will have a heavy tail as well. A saturating activation would then be undesirable, as it makes the tail less important in the saturating region. ReLU-type activations are not affected, as they are non-zero on $[0,\infty)$ and do not saturate. \par 

\subsection{KS statistic and Wasserstein distance during training}
\label{appendix:KS_Wass_training}

\begin{figure}[h!]
    \centering
    \begin{subfigure}{0.49\linewidth}
        \includegraphics[width=\linewidth]{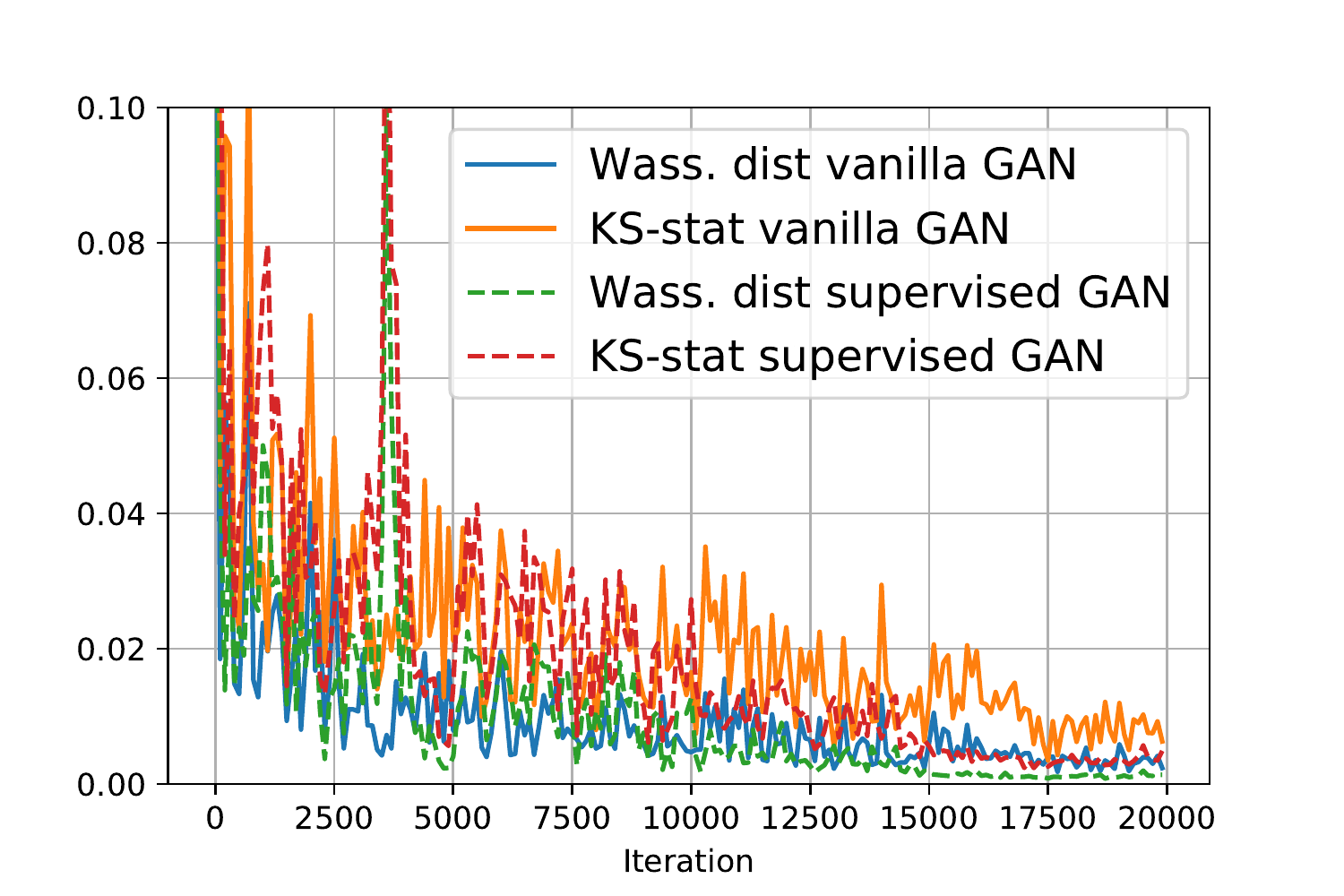}
        \caption{GBM}
        \label{fig:KS_Wass_training_GBM}
    \end{subfigure}
    \begin{subfigure}{0.49\linewidth}
        \includegraphics[width=\linewidth]{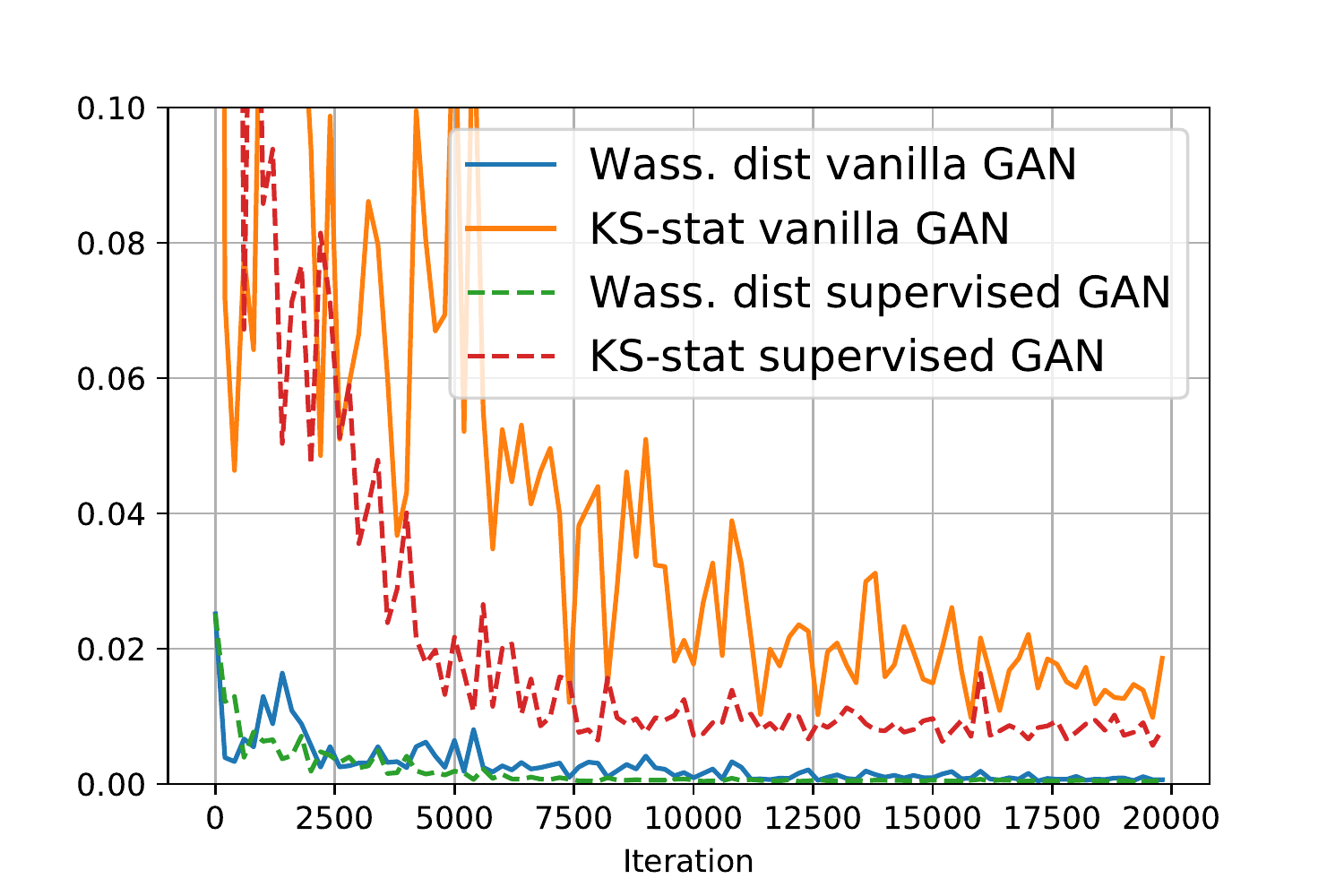}
        \caption{CIR, Feller condition satisfied}
        \label{fig:KS_Wass_training_CIR}
    \end{subfigure}
    \begin{subfigure}{0.49\linewidth}
        \includegraphics[width=\linewidth]{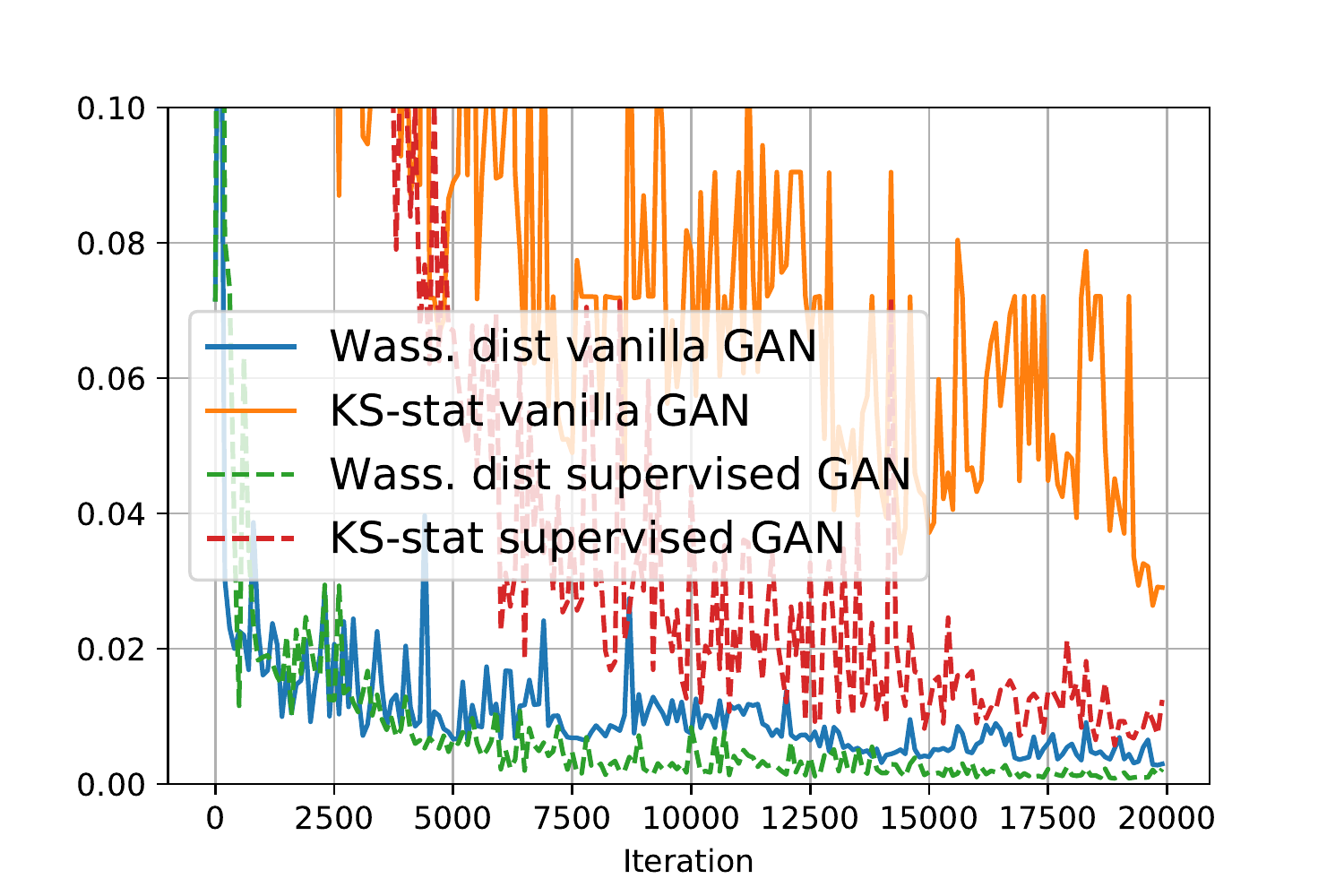}
        \caption{CIR, Feller condition violated}
        \label{fig:KS_Wass_training_CIR_Feller_not}
    \end{subfigure}    
    \caption{KS statistic and Wasserstein distance during training on a test set of 100,000 i.i.d. samples from the distribution of $S_{t+\Delta t}\mid S_t$. The default parameters listed in Section \ref{appendix:model_parameters} were used to generate test sets. The supervised GAN converges faster than the vanilla GAN on both metrics.}
    \label{fig:KS_Wass_training}
\end{figure}

\begin{figure}[h!]
    \centering
    \begin{subfigure}{0.49\linewidth}
        \includegraphics[width=\linewidth]{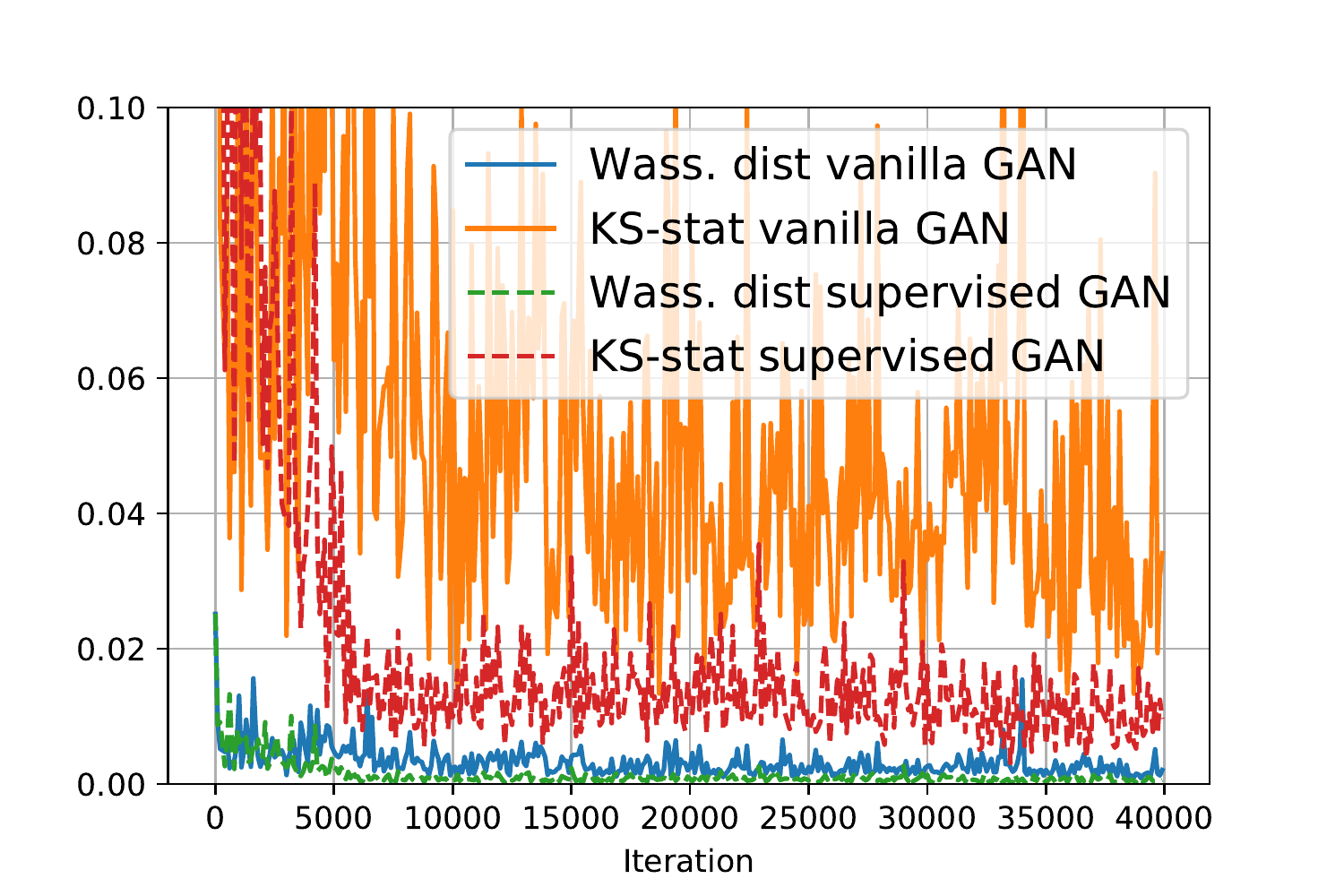}
        \caption{No schedule}
        \label{fig:LR_schedule_None}
    \end{subfigure}
    \begin{subfigure}{0.49\linewidth}
        \includegraphics[width=\linewidth]{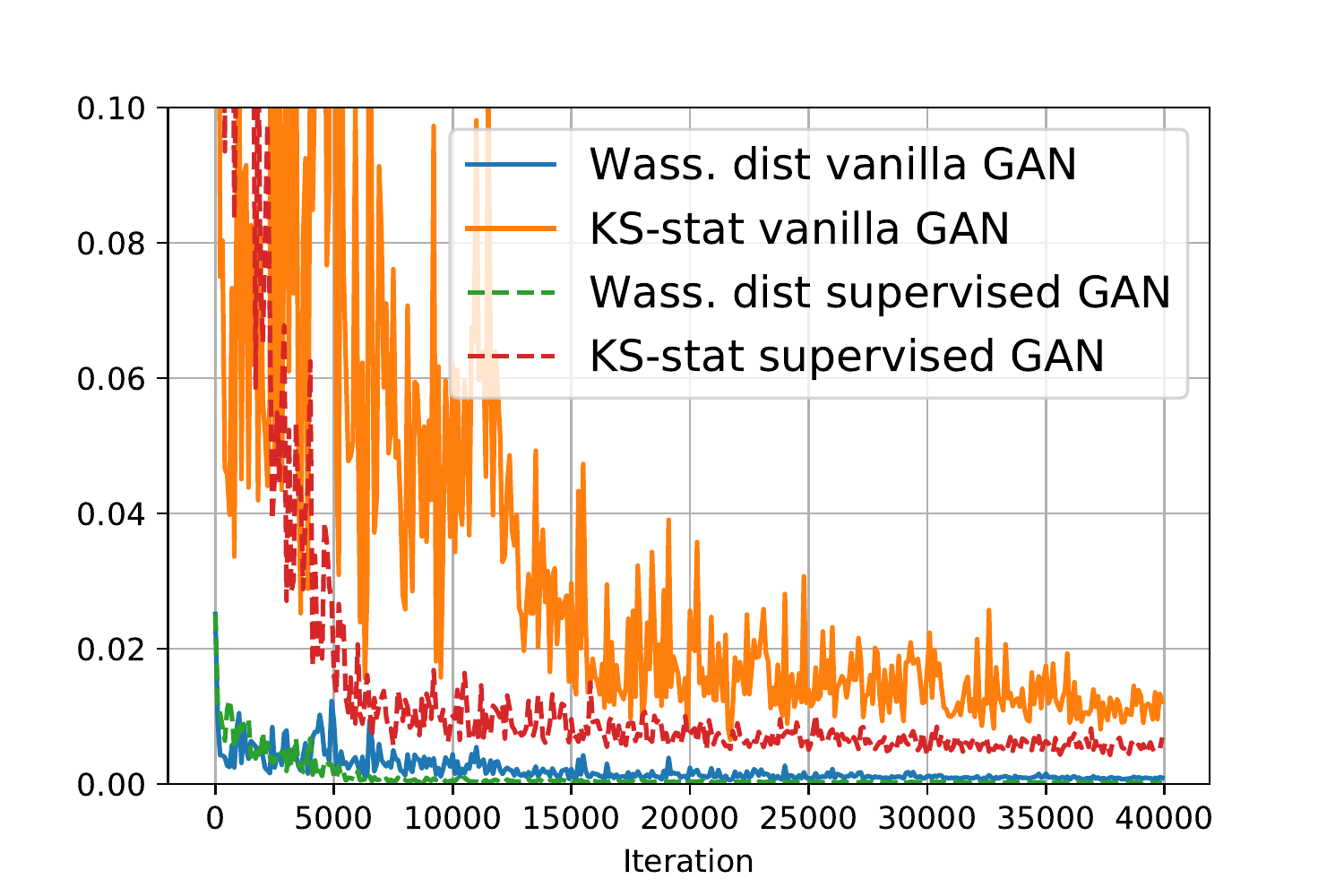}
        \caption{Divide by 1.05 every 500 iterations.}
        \label{fig:LR_schedule_1_05}
    \end{subfigure}
    \caption{Example of the effect of the learning rate schedule during training on the CIR process if the Feller condition is satisfied. The gradually decreasing learning rate allows the generator to more accurately converge to the target distribution.}
    \label{fig:LR_schedule}
\end{figure}

The KS statistic and Wasserstein distance were computed during training on a test set of $10^5$ samples for GBM and the two instances of the CIR process. Figure \ref{fig:KS_Wass_training} compares the training process of the vanilla and supervised GAN. Under equal training conditions, the supervised GAN converges faster and achieves a better approximation in distribution in all three cases. Figure \ref{fig:LR_schedule} shows the effect of using a learning rate schedule for the generator. 

\section{Results on GBM and the case Feller condition satisfied}
\label{appendix:additional_results}

\begin{figure}[h!]
    \centering
    \begin{subfigure}{0.49\linewidth}
        \includegraphics[width=\linewidth]{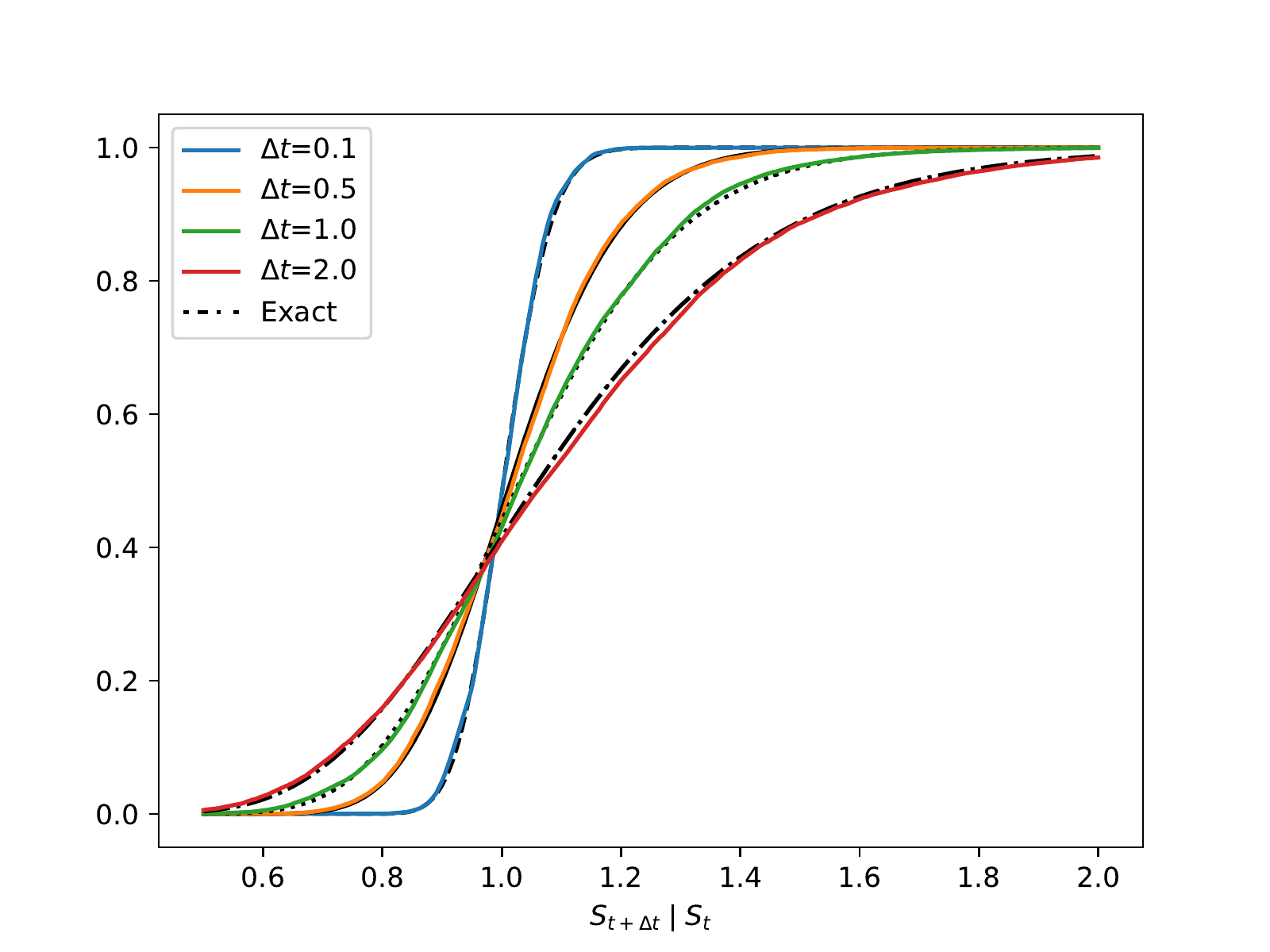}
        \caption{Vanilla GAN}
        \label{fig:CGAN_GBM_t_vanilla}
    \end{subfigure}
    \begin{subfigure}{0.49\linewidth}
        \includegraphics[width=\linewidth]{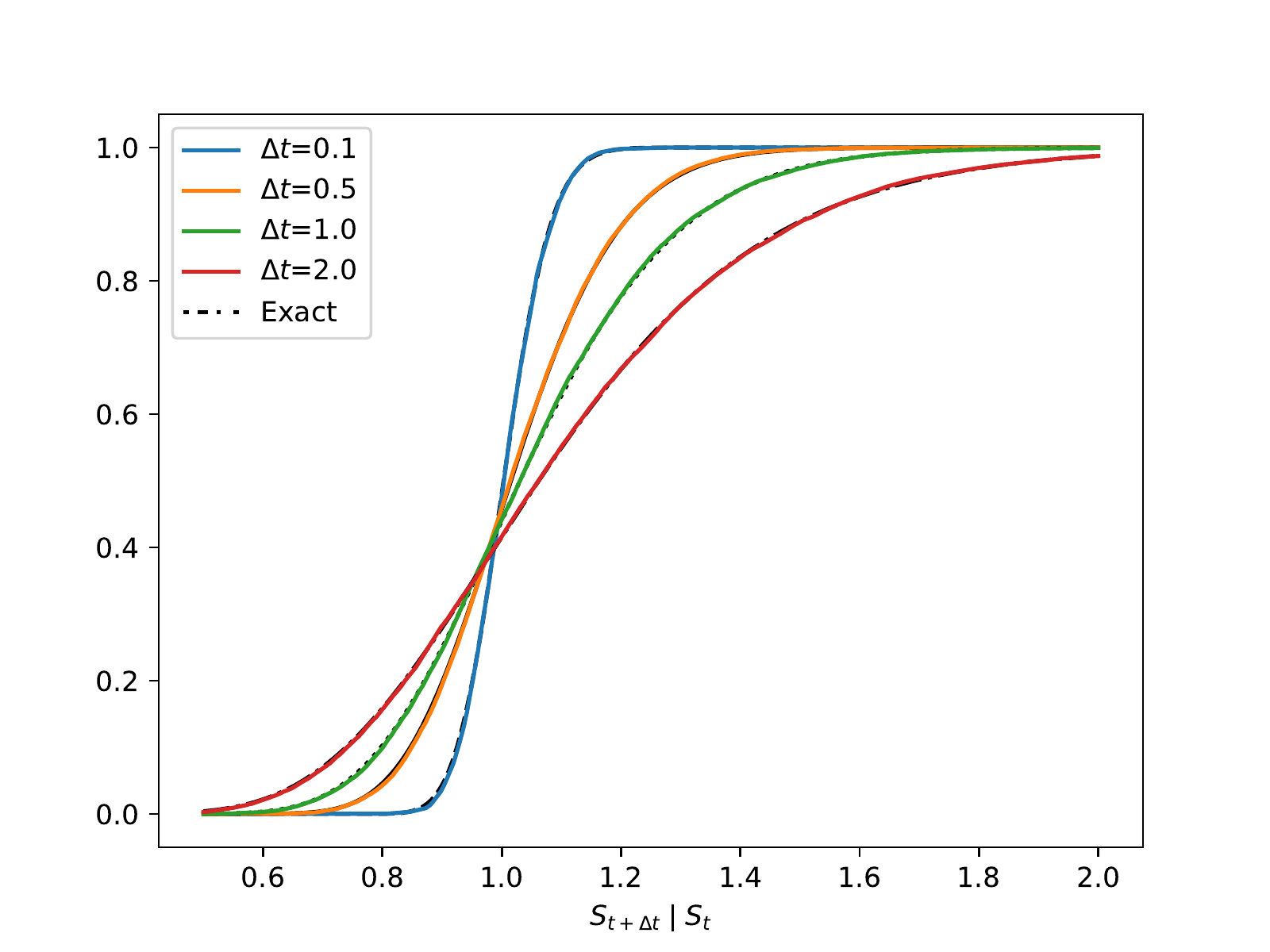}
        \caption{Supervised GAN}
        \label{fig:CGAN_GBM_t_constrained}
    \end{subfigure}
    \caption{ECDF plots of the conditional GAN output on the GBM problem for various choices of $\Delta t$. Note that pre-processing the data with logreturns removes dependence on $S_t$. In this example, $S_t=1$.}
    \label{fig:CGAN_GBM_vanilla_vs_constrained}
\end{figure}

\begin{figure}[h!]
    \centering
    \begin{subfigure}{0.49\linewidth}
        \includegraphics[width=\linewidth]{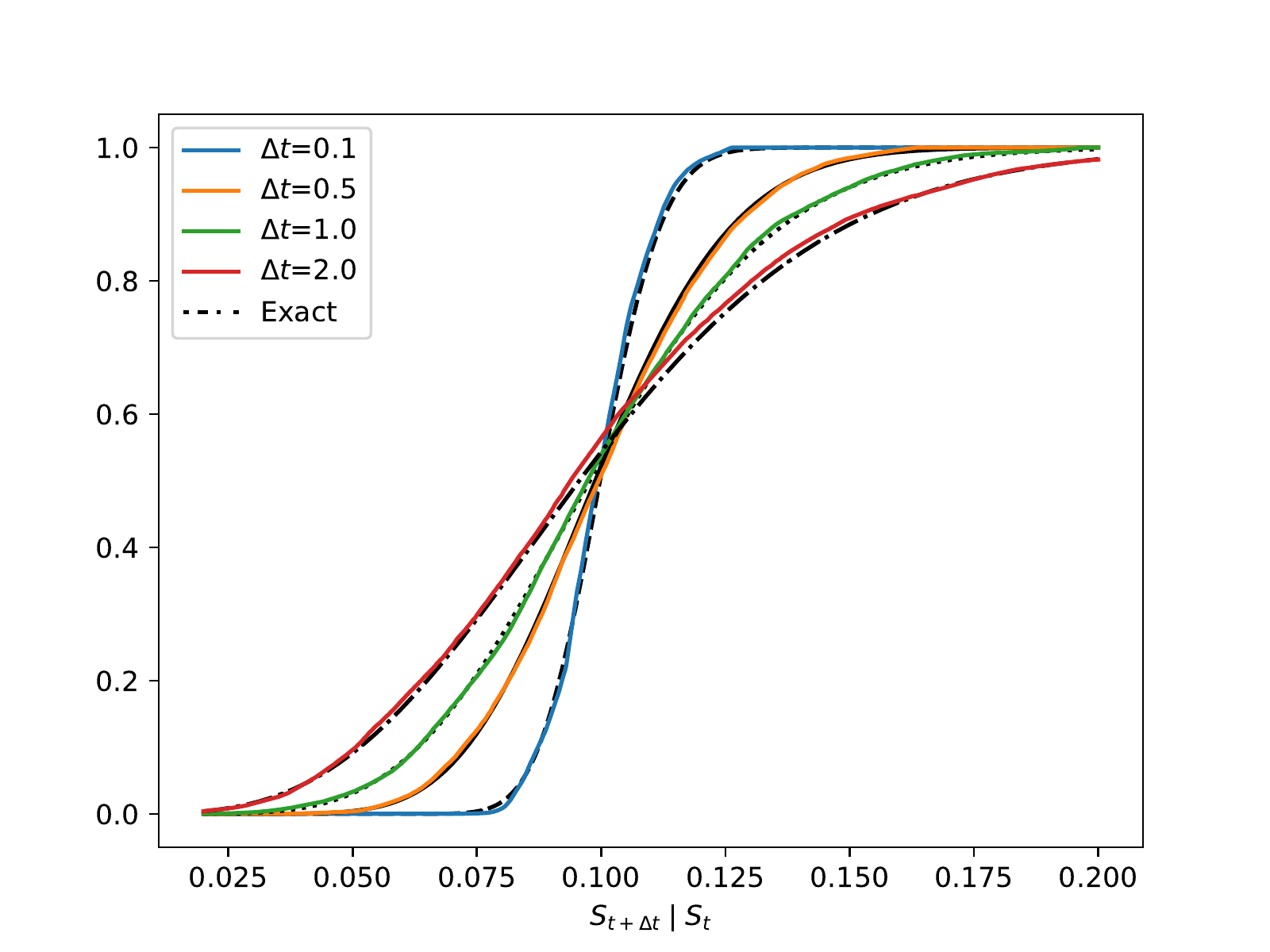}
        \caption{Vanilla GAN}
        \label{fig:CGAN_CIR_t_vanilla}
    \end{subfigure}
    \begin{subfigure}{0.49\linewidth}
        \includegraphics[width=\linewidth]{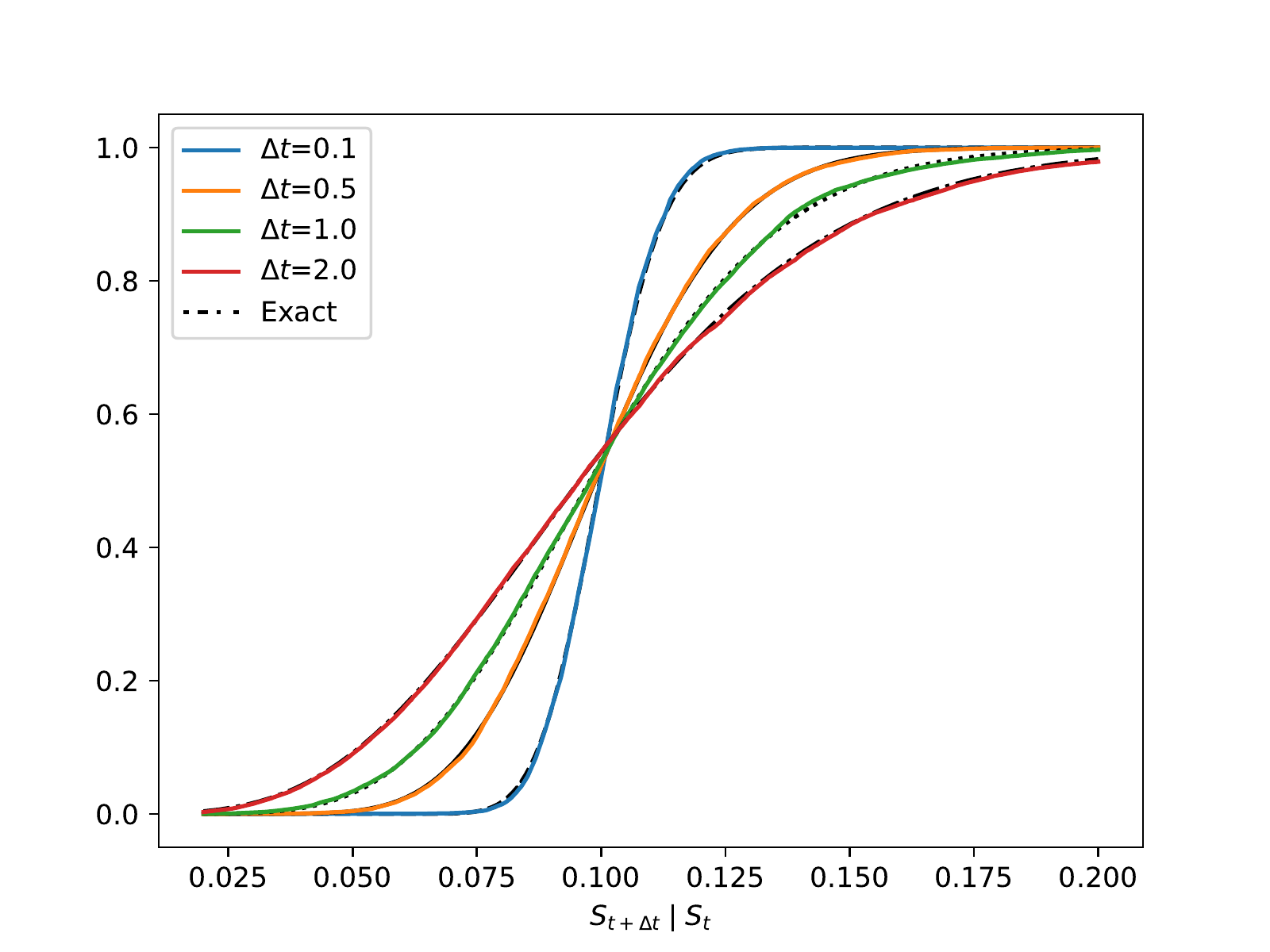}
        \caption{Supervised GAN}
        \label{fig:CGAN_CIR_t_constrained}
    \end{subfigure}
    \caption{ECDF plots of the conditional GAN output on the CIR process with the Feller condition satisfied, for various choices of $\Delta t$. $S_t$ was held fixed at $0.1$.}
    \label{fig:CGAN_CIR_vanilla_vs_constrained}
\end{figure}

\begin{figure}[h!]
\centering
    \begin{subfigure}{0.49\linewidth}
        \centering
        \includegraphics[width=\linewidth]{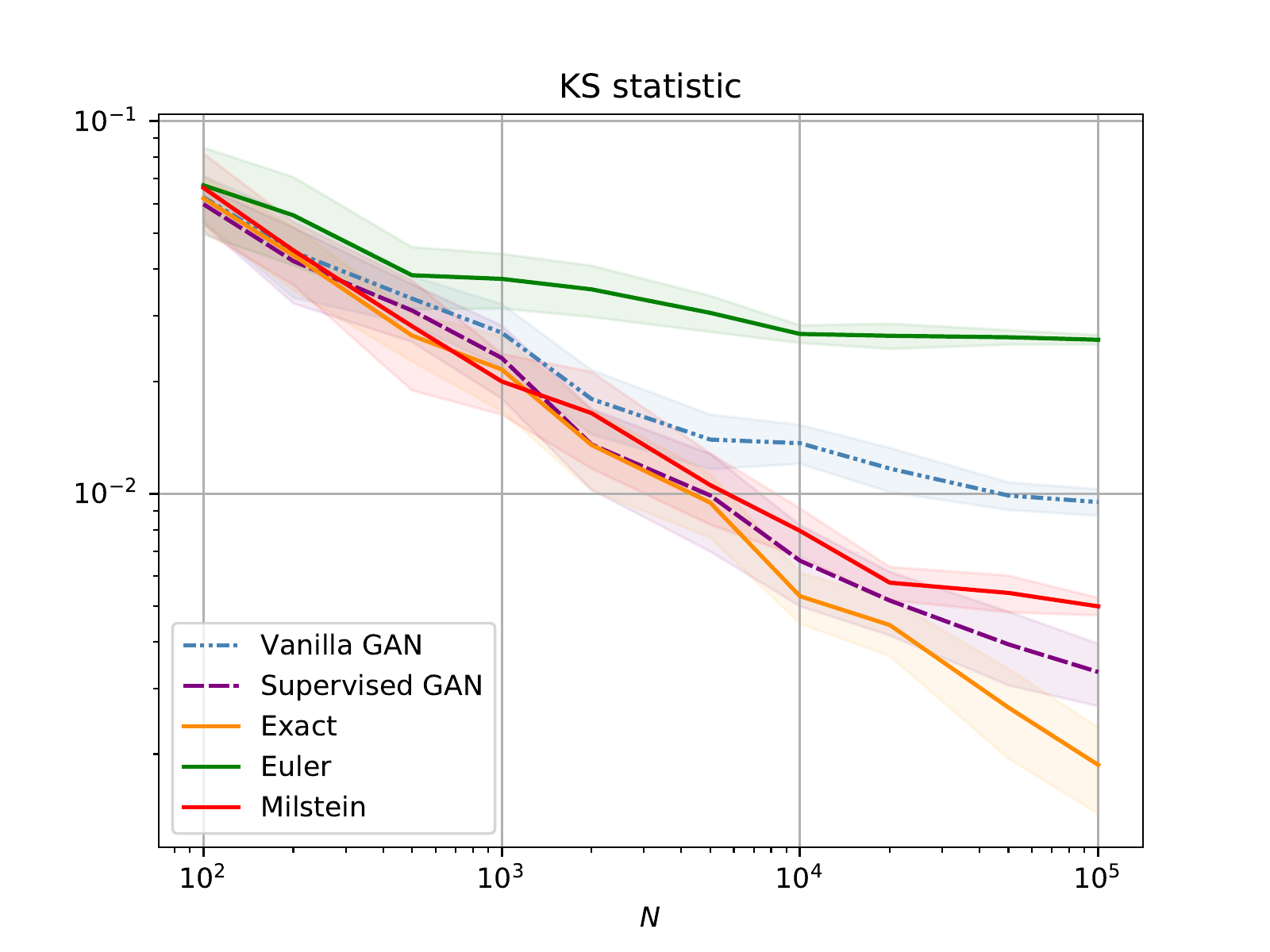}
        \caption{GBM, KS statistic}
        \label{fig:benchmark_N_GBM_t_0_4_KS}
    \end{subfigure}
    \begin{subfigure}{0.49\linewidth}
        \centering
        \includegraphics[width=\linewidth]{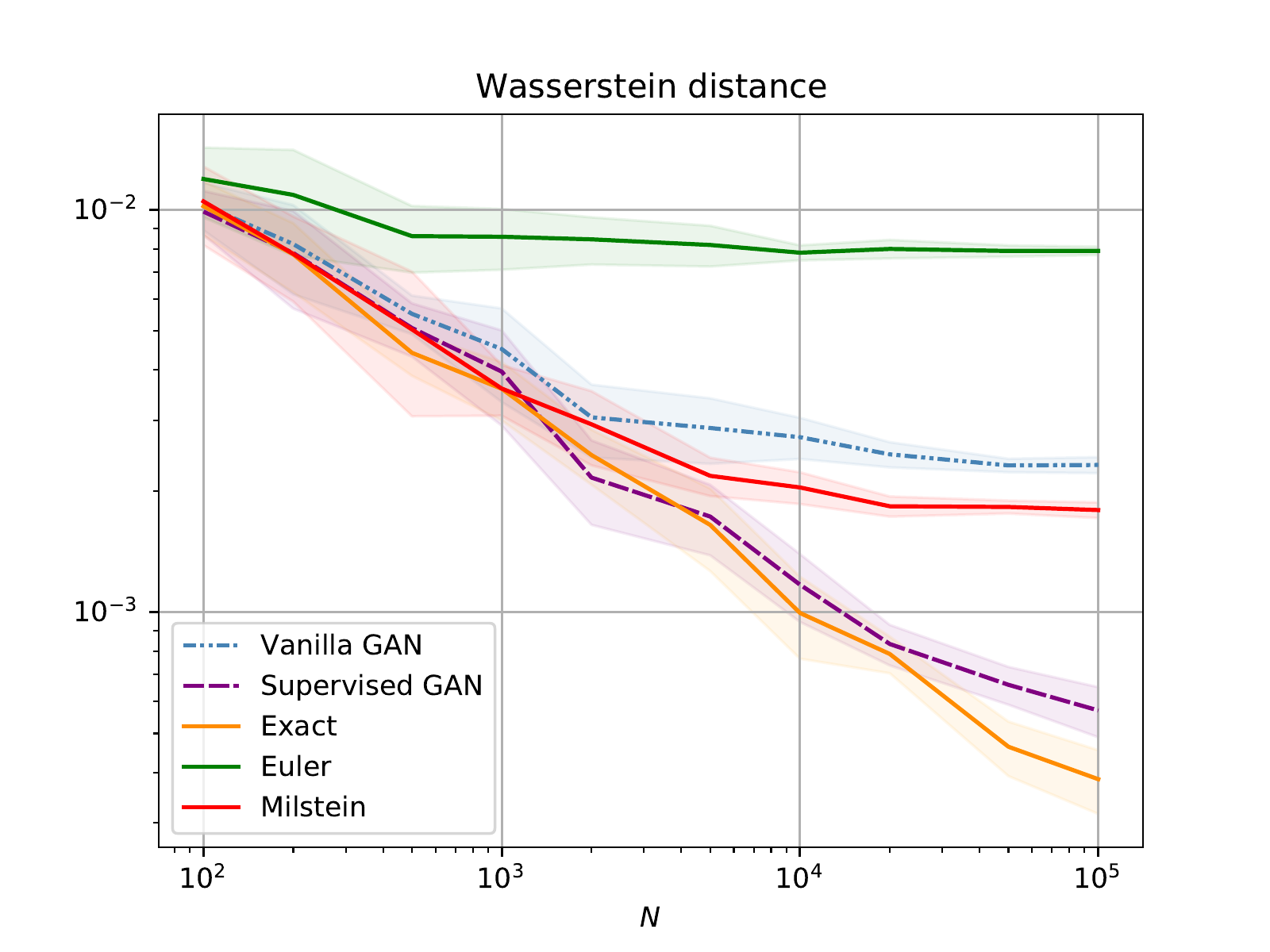}
        \caption{GBM, Wasserstein distance}
        \label{fig:benchmark_N_GBM_t_0_4_Wass}
    \end{subfigure}
    \begin{subfigure}{0.49\linewidth}
        \centering
        \includegraphics[width=\linewidth]{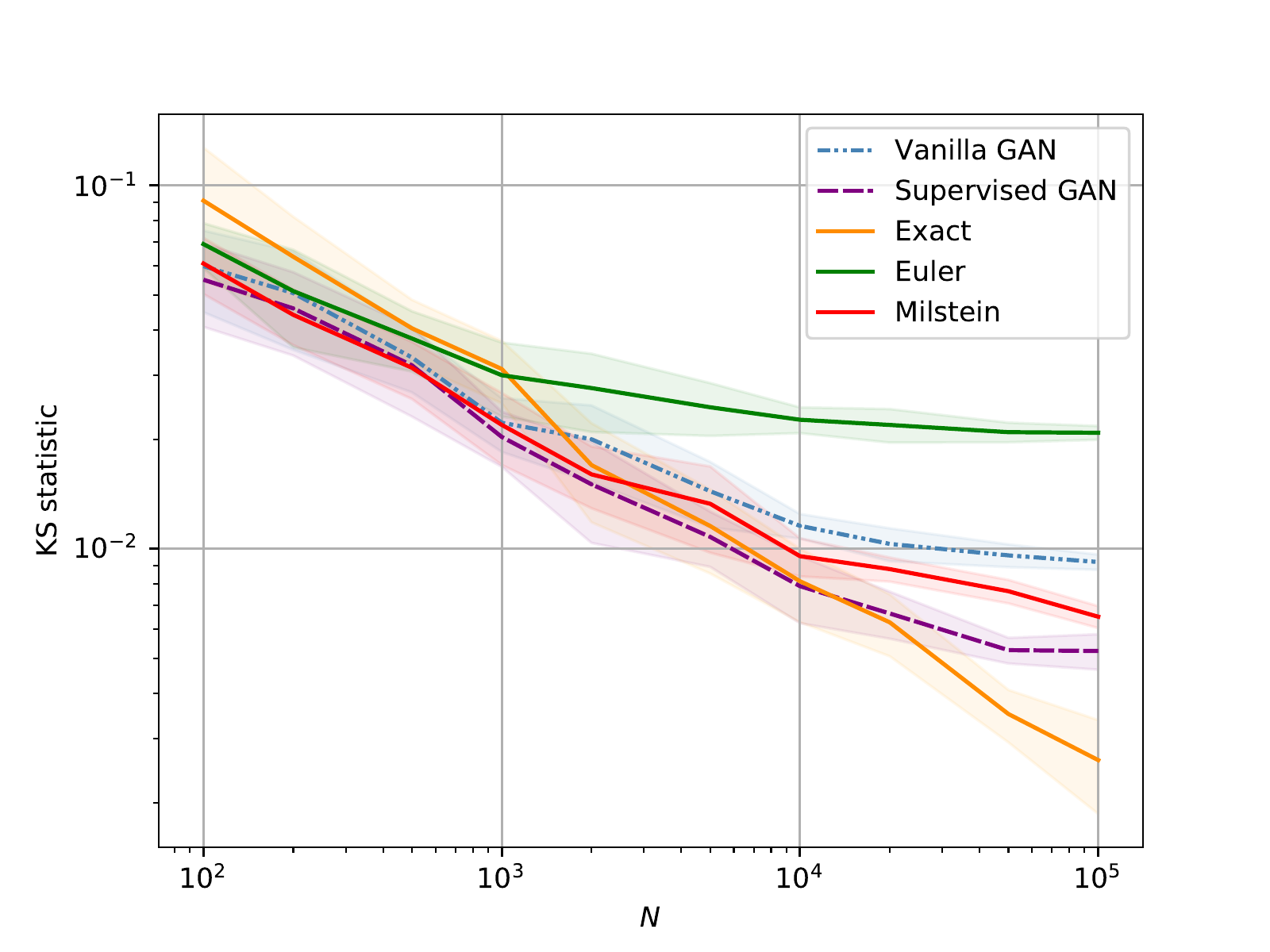}
        \caption{CIR, KS statistic}
        \label{fig:benchmark_N_CIR_t_0_4_KS}
    \end{subfigure}
    \begin{subfigure}{0.49\linewidth}
        \centering
        \includegraphics[width=\linewidth]{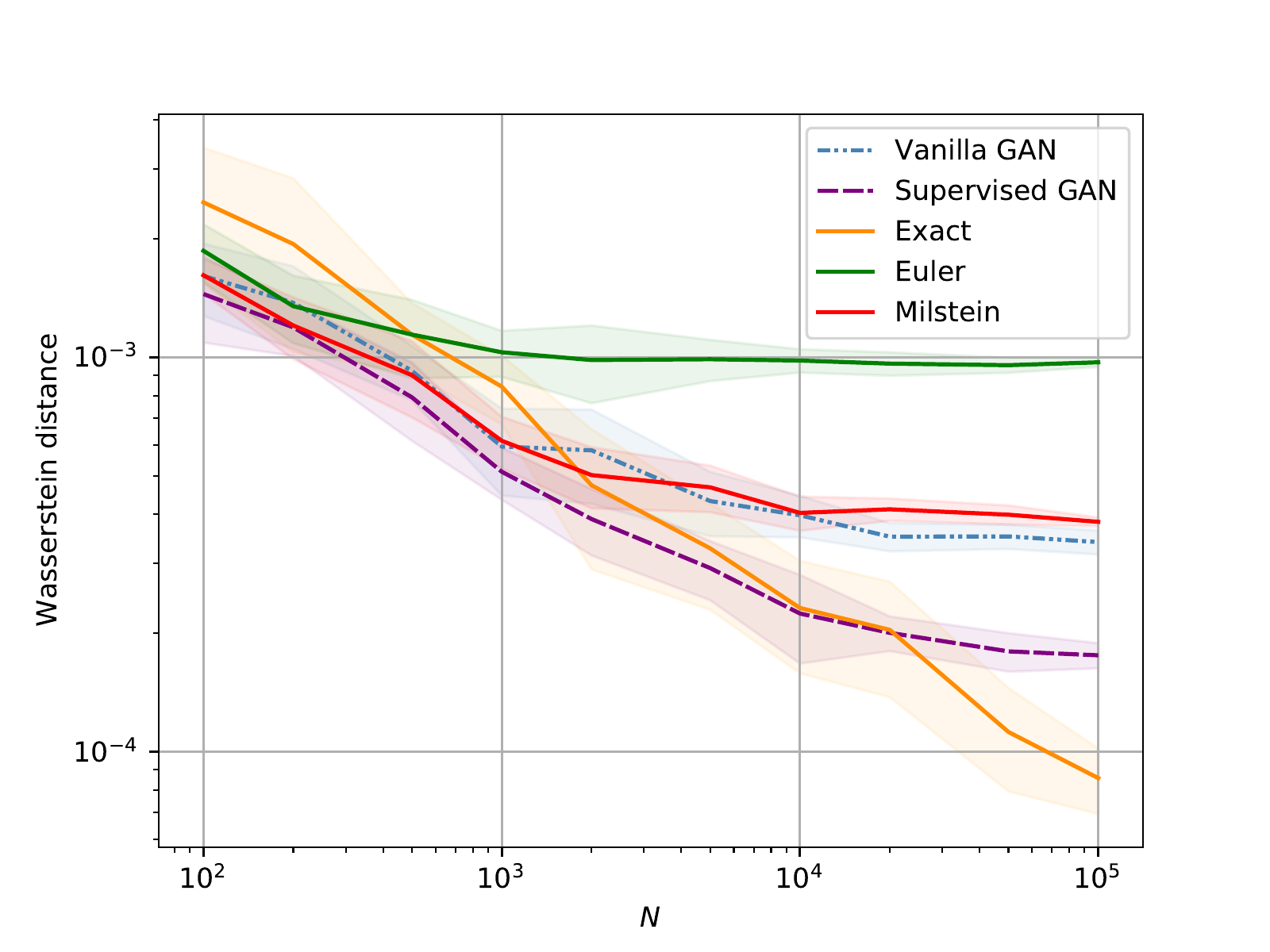}
        \caption{CIR, Wasserstein distance}
        \label{fig:benchmark_N_CIR_t_0_4_Wass}
    \end{subfigure}    
    \caption{KS statistic and Wasserstein distance at $\Delta t=0.4$, versus the size of the test set. The confidence bands show the standard deviation based on 10 repetitions of the experiment, i.e.\ 10 iid samples of $N$ random inputs to both GANs. The mean of both statistics is reported in the solid and dashed lines.}
    \label{fig:benchmark_N_Feller_GBM_CIR}
\end{figure}

\subsection{Weak and strong error}
\label{appendix:weak_strong}
Here, we show the weak and strong error obtained with artificial paths constructed with the vanilla and supervised GAN, this time for GBM and the CIR process if the Feller condition is satisfied. In the GBM example, the vanilla GAN happened to find a strong approximation, i.e. the same map as the conditional inverse distribution on the prior (Equation \eqref{eq:sample_S_from_Z}). On the example we show for the CIR process with the Feller condition satisfied, it did not manage to provide a strong approximation. 

\begin{figure}[h!]
    \centering
    \begin{subfigure}{0.49\linewidth}
        \includegraphics[width=\linewidth]{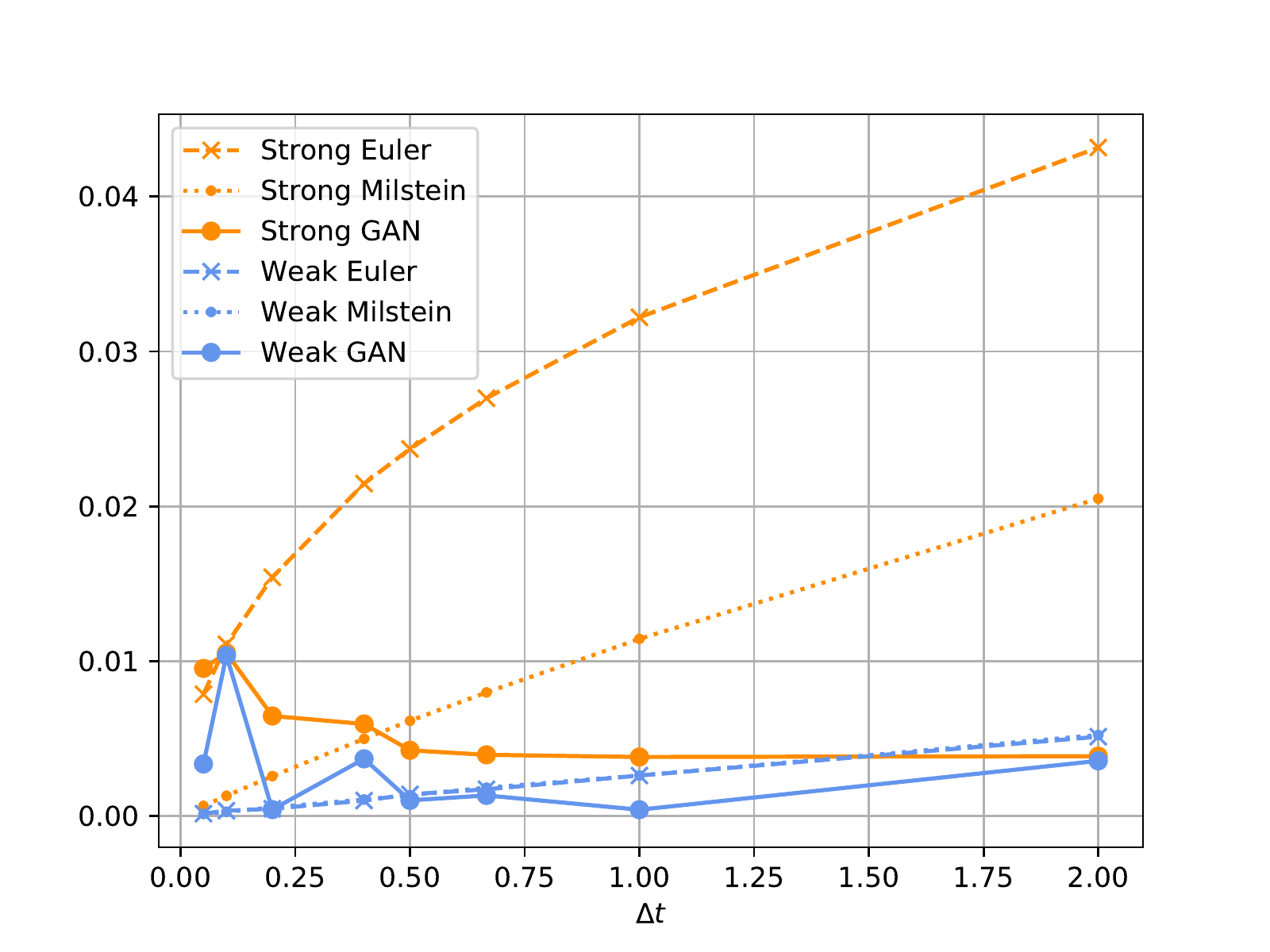}
        \caption{Vanilla GAN}
        \label{fig:GBM_weak_strong_vanilla}
    \end{subfigure}
    \begin{subfigure}{0.49\linewidth}
        \includegraphics[width=\linewidth]{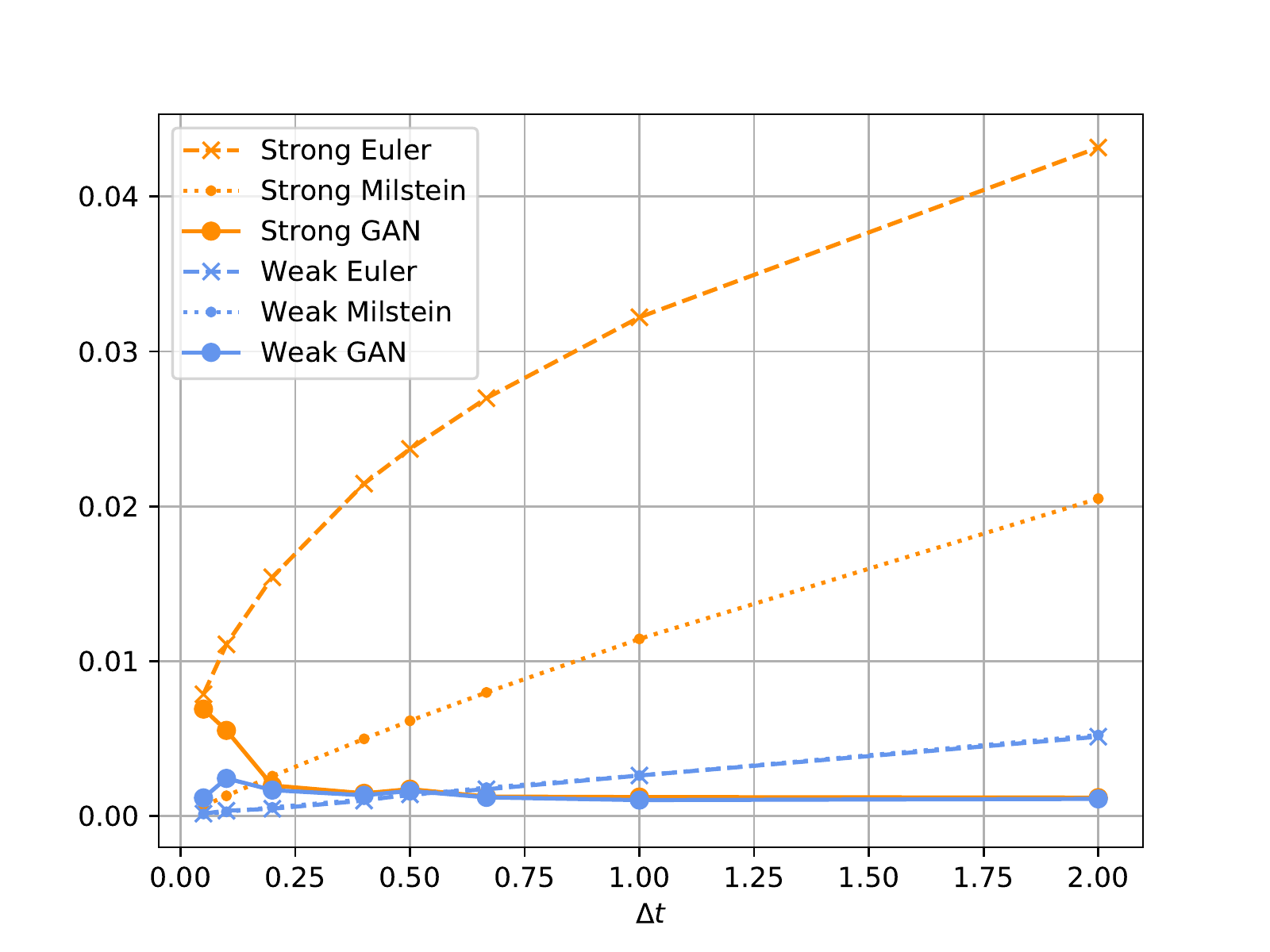}
        \caption{Supervised GAN}
        \label{fig:GBM_weak_strong_supervised}
    \end{subfigure}
    \caption{GBM: Weak and strong errors at time $T=2$ of paths constructed using the vanilla GAN and supervised GAN, compared with the Euler scheme, Milstein scheme and exact scheme. $S_0=1$, $\Delta t=0.05$, i.e. 40 steps between 0 and 2.}
    \label{fig:GBM_weak_strong}
\end{figure}

\begin{figure}[h!]
    \centering
    \begin{subfigure}{0.49\linewidth}
        \includegraphics[width=\linewidth]{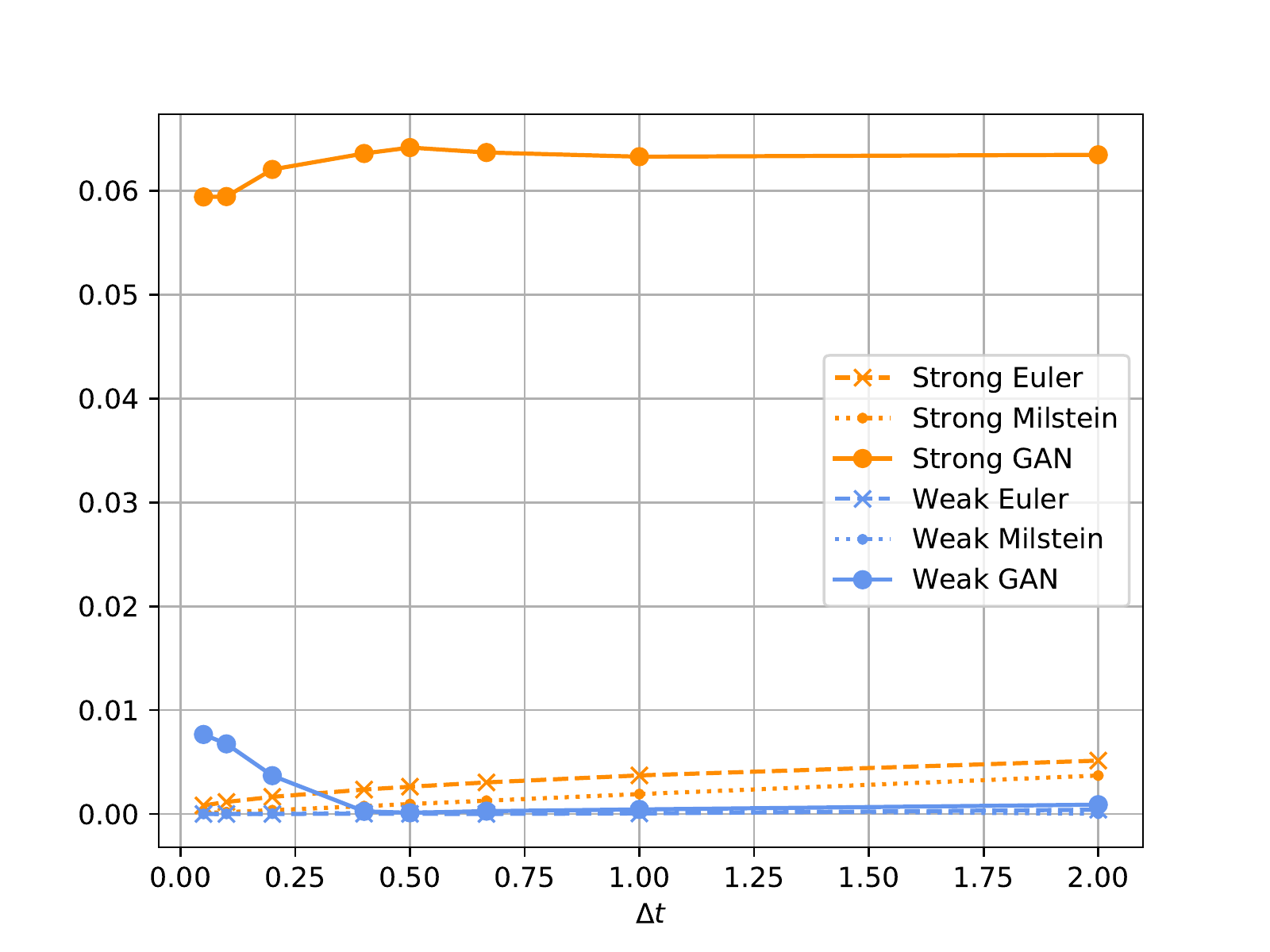}
        \caption{Vanilla GAN}
        \label{fig:CIR_weak_strong_vanilla_0_1}
    \end{subfigure}
    \begin{subfigure}{0.49\linewidth}
        \includegraphics[width=\linewidth]{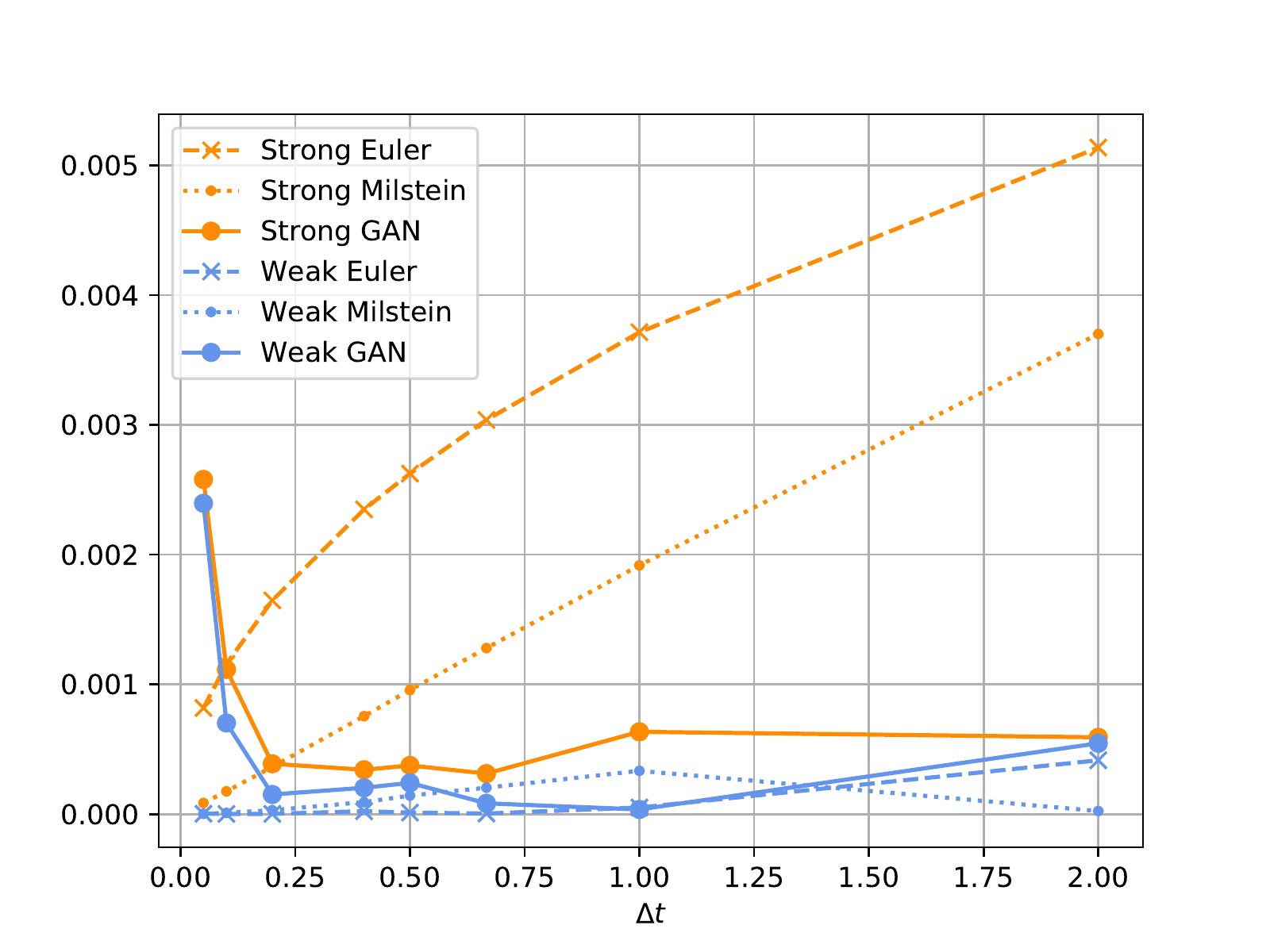}
        \caption{Supervised GAN}
        \label{fig:CIR_weak_strong_supervised_0_1}
    \end{subfigure}
    \begin{subfigure}{0.49\linewidth}
        \includegraphics[width=\linewidth]{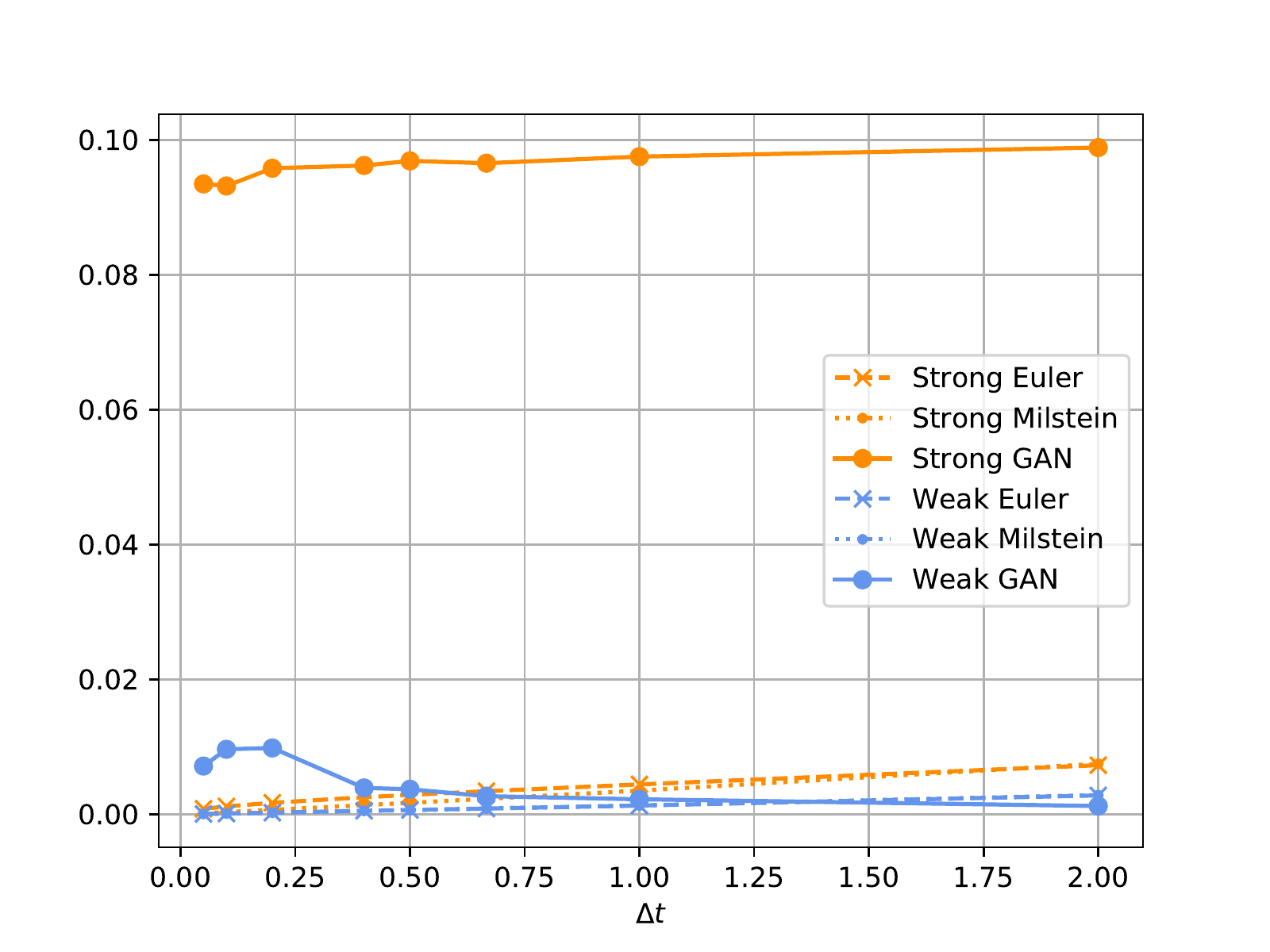}
        \caption{Vanilla GAN}
        \label{fig:CIR_weak_strong_vanilla_0_25}
    \end{subfigure}
    \begin{subfigure}{0.49\linewidth}
        \includegraphics[width=\linewidth]{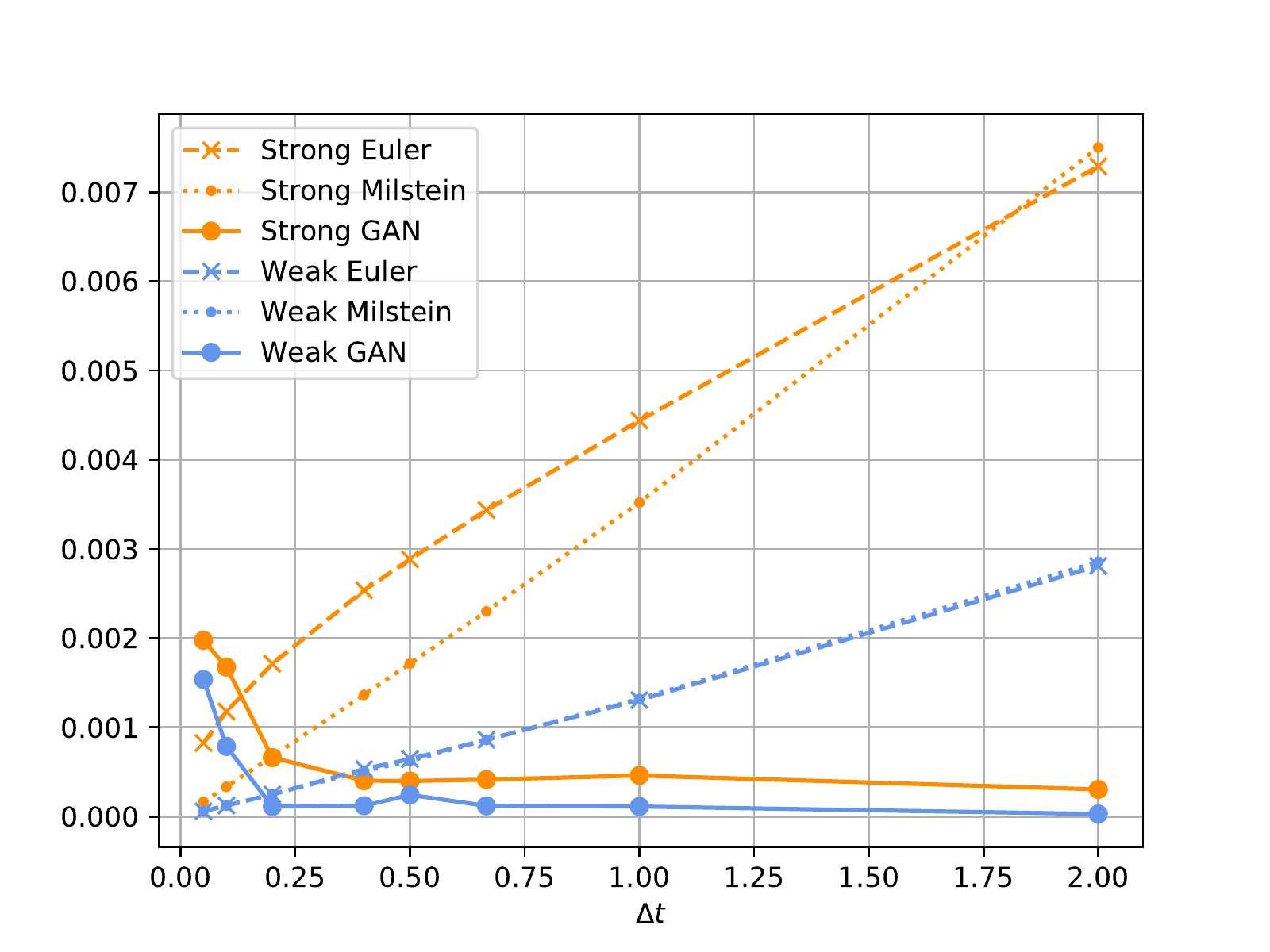}
        \caption{Supervised GAN}
        \label{fig:CIR_weak_strong_supervised_0_25}
    \end{subfigure}    
    \caption{CIR process with Feller condition satisfied: weak and strong errors at time $T=2$ of paths constructed using the vanilla GAN and supervised GAN, compared with the Euler scheme, Milstein scheme and exact scheme. $\Delta t$ ranges between $0.05$ and 2. In \ref{fig:CIR_weak_strong_vanilla_0_1} and \ref{fig:CIR_weak_strong_supervised_0_1}, $S_0=\bar{S}$, which gives the discrete-time schemes an advantage, as they have the mean reversion `built-in'. Therefore, we also show the results if $S_0\neq \bar{S}$, e.g. if $S_0=0.25$ in \ref{fig:CIR_weak_strong_vanilla_0_25} and \ref{fig:CIR_weak_strong_supervised_0_25}.}
    \label{fig:CIR_weak_strong}
\end{figure}

\section{Details on the model parameters}
\label{appendix:model_parameters}
\bfquad{CIR process SDE parameters} In the case of the CIR process, the conditional random process $S_{t+\Delta t} \mid S_t$ follows a scaled non-central $\chi^2$-distribution with some non-centrality parameter $\xi$, degrees of freedom $\delta$ and scaling factor $\bar{c}$ \cite{cox2005theory,oosterlee2019_book}:

\begin{equation}
    S_{t+\Delta t} \mid S_t\  \sim\  \bar{c}\ \chi^2(\xi,\delta),
    \label{eq:sample_CIR_r}
\end{equation}

where $\bar{c}, \xi$ and $\delta$ are related to the SDE parameters as shown in equations \eqref{eq:ksi}-\eqref{eq:c_delta_t}, cf. \cite[p.392]{cox2005theory}. 
\begin{align}
    \xi(S_t,\Delta t) &= \frac{4\kappa S_t e^{-\kappa \Delta t}}{\gamma^2\left( 1 - e^{-\kappa \Delta t} \right)}, \label{eq:ksi} \\
    \delta &= \frac{4\kappa \bar{S}}{\gamma^2}, \\
    \bar{c}(\Delta t) &= \frac{\gamma^2}{4\kappa} \left( 1 - e^{-\kappa \Delta t} \right). \label{eq:c_delta_t}
\end{align}

\bfquad{Modified Euler and Milstein scheme for the CIR process} A practical consideration for the CIR process is that discrete-time schemes could give rise to negative values, which are problematic when computing the square root in Equation \eqref{eq:SDE_CIR}. Therefore, the Euler scheme will be replaced by what we will refer to as the (partially) \textit{`truncated' Euler scheme}, as mentioned e.g. in \cite{labbe2010discr_CIR}. In the case of the CIR process it is given by: 
\begin{equation}
\label{eq:trunc_Euler}
    \hat{S}_{t+\Delta t} = \hat{S}_{t} + \kappa(\bar{S}-\hat{S}_{t} )\Delta t + \gamma \sqrt{\hat{S}_{t}^+}\sqrt{\Delta t}Z, 
\end{equation}
where $S_{0}\in \mathbb{R}$ and $\hat{S}_{t}^+:=\max(\hat{S}_{t},0)$. $Z\sim N(0,1)$. Note that the truncated Euler scheme may still produce negative paths, in which case the term with the Brownian motion is equal to zero at step $t+\Delta t$. \par 

A modified version of the Milstein scheme can be defined as well. In \cite{hefter2018Milstein_CIR}, such a Milstein-type scheme has been proposed specifically for the CIR process, which we implemented as a reference to the CIR process. This \textit{truncated Milstein scheme} is given by in Equation \eqref{eq:trunc_Milstein_scheme}. 
\begin{equation}
    \label{eq:trunc_Milstein_scheme}
    \hat{S}_{t_{k+1}} = \left( \left( \max\left(\frac{1}{2}\gamma \sqrt{\Delta t}, \sqrt{\max\left(\frac{1}{2}\gamma \sqrt{\Delta t},\hat{S}_{t_k}\right)} + \frac{1}{2}\gamma \sqrt{\Delta t}Z_k \right) \right)^2 + \left( \kappa \bar{S} - \frac{1}{4} \gamma^2 - \kappa \hat{S}_{t_k} \right)\Delta t \right)^+,
\end{equation}
where $\hat{S}_{t_0} = S_{t_0}$ and $(\cdot)^+:=\max(\cdot,0)$. The one-step order of convergence of this scheme depends on the previous value $S_{t_k}$, time step $\Delta t$ and degrees of freedom parameter $\delta$ \cite{hefter2018Milstein_CIR}. However, the authors of \cite{hefter2018Milstein_CIR} show that the scheme converges in $L^p$ with order $\frac{1}{2p}\min(1,\delta)$. Note that we could have used a Milstein scheme analogous to the truncated Euler scheme, but this led to inferior weak and strong errors compared to the scheme in \cite{hefter2018Milstein_CIR}. 

\bfquad{Default training/testing parameters} For GBM, the default parameters were $\mu=0.05$ and $\sigma=0.2$. The CIR parameters were $\bar{S}=0.1,S_t=0.1,\kappa=0.1, \gamma\in \{0.1,0.3\}$, where $\gamma=0.1$ corresponds to the case where the Feller condition is satisfied ($\delta = 4$) and $\gamma=0.3$ to the case where the Feller condition is violated $(\delta = 0.44)$. \par 

\bfquad{Conditional GAN training} Both GANs were trained on a range of parameters of $\Delta t$ and (for the CIR process) the `previous value' $S_t$. It was found that choosing a discrete range parameters that recur many times outperformed a continuous range of unique parameters. For example, for $\Delta t$, a fixed list was created of times of interest $\Delta t \in \{0.05,0.1,0.2,0.4,0.5,0.67,1,2\}$, each of which occured with equal frequency in a dataset of $N_{\text{train}}$ training samples. This outperformed a continuous range of $N_{\text{train}}$ time steps on $[0,2]$. If more than one conditional parameter is chosen, a training set must be defined as the Cartesian product of the two discrete sets of training parameters. This was achieved by randomly permuting each vector of training samples and concatenating the results. This created ordered vectors of the pairs $(S_t,\Delta t)$, which were provided as input to a function that draws samples from the exact distribution of $S_{t+\Delta t}\mid S_t$ and provides corresponding standard normal variates $Z$ to train the supervised GAN. \par

\section{Autocorrelation structure for results on CIR process}
\label{appendix:autocorrelation}
To test whether the GAN has correctly captured the dependence of the conditional distribution on the previous $S_t$, we show a plot of $S_{t+\Delta t}\mid S_t$ versus $S_t$. These plots were made by first drawing $1,000$ samples using the exact distribution of $S_{t}\mid S_0$ at $t=1$ and $S_0 = 0.1$. Then, an additional $1,000$ samples $S_{t+\Delta t}\mid S_t$ were drawn from the exact distribution conditional on the samples $S_t$, with $\Delta t=1$. $Z$ was computed using Equation \eqref{eq:sample_Z_from_S} and provided as input to the generator. 

\begin{figure}[h]
    \centering
    \begin{subfigure}{0.49\linewidth}
        \includegraphics[width=\linewidth]{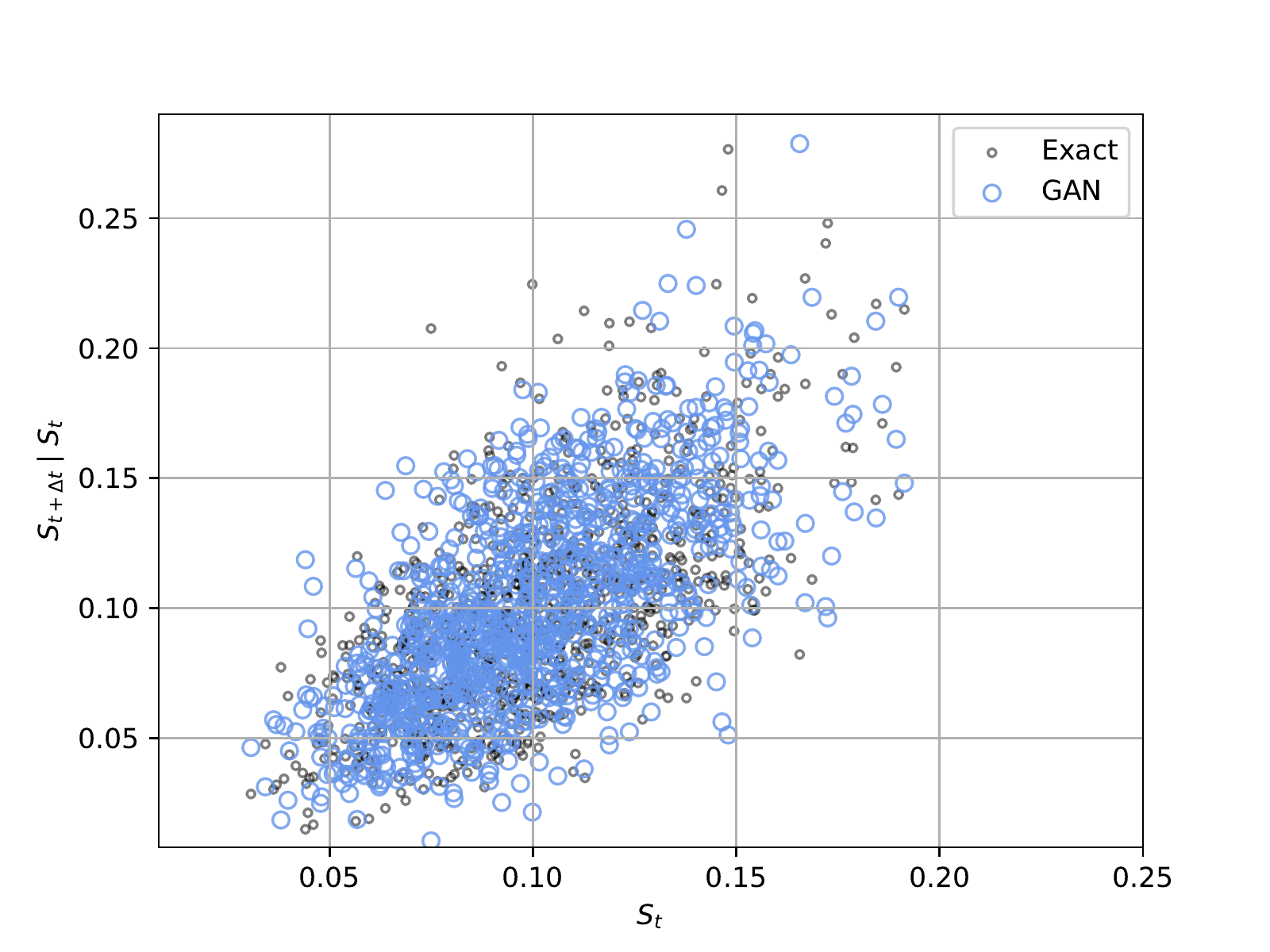}
        \caption{Vanilla GAN}
        \label{fig:CIR_vanilla_autocorr}
    \end{subfigure}
    \begin{subfigure}{0.49\linewidth}
        \includegraphics[width=\linewidth]{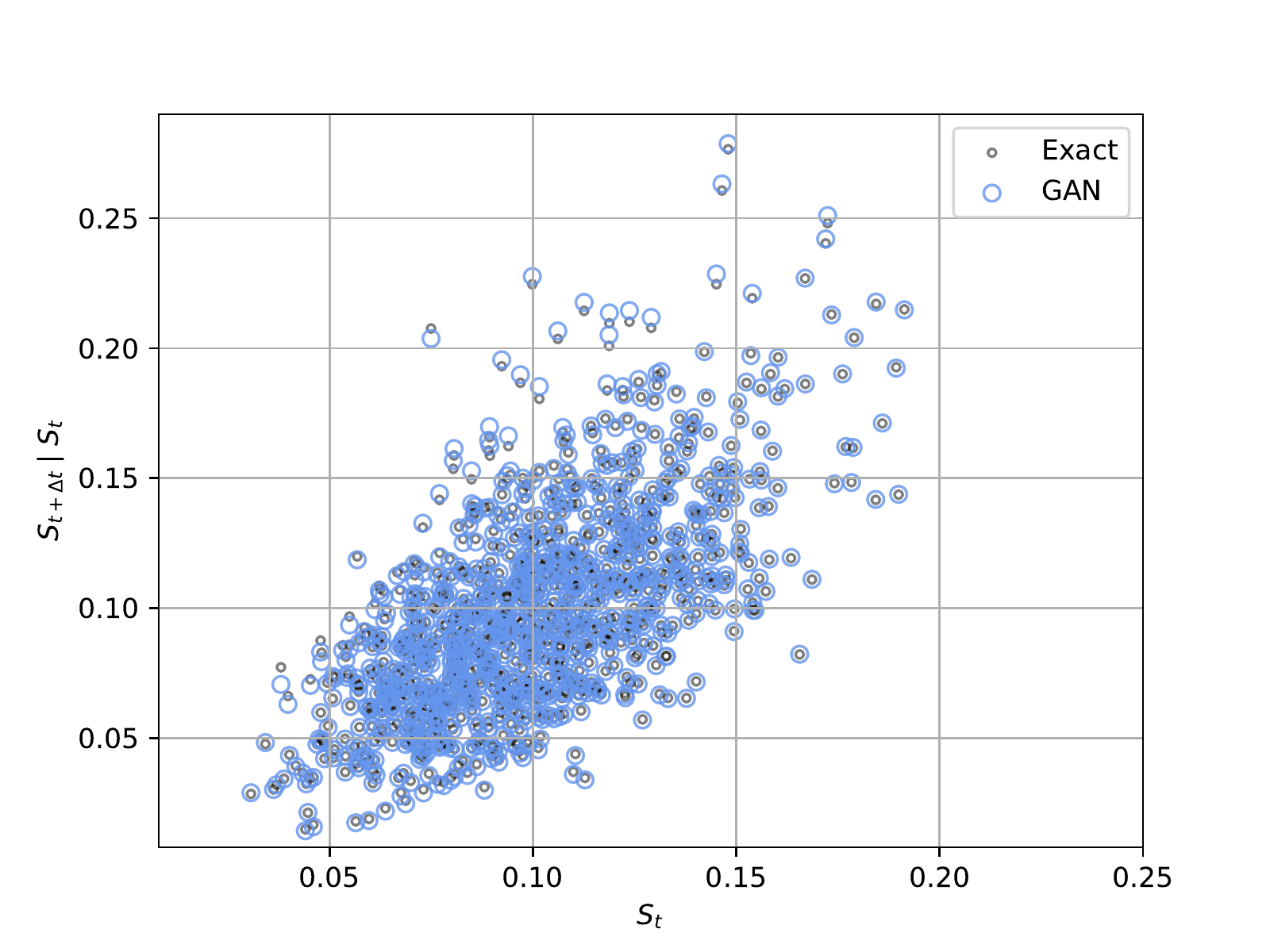}
        \caption{Supervised GAN}
        \label{fig:CIR_supervised_autocorr}
    \end{subfigure}
    \begin{subfigure}{0.49\linewidth}
        \includegraphics[width=\linewidth]{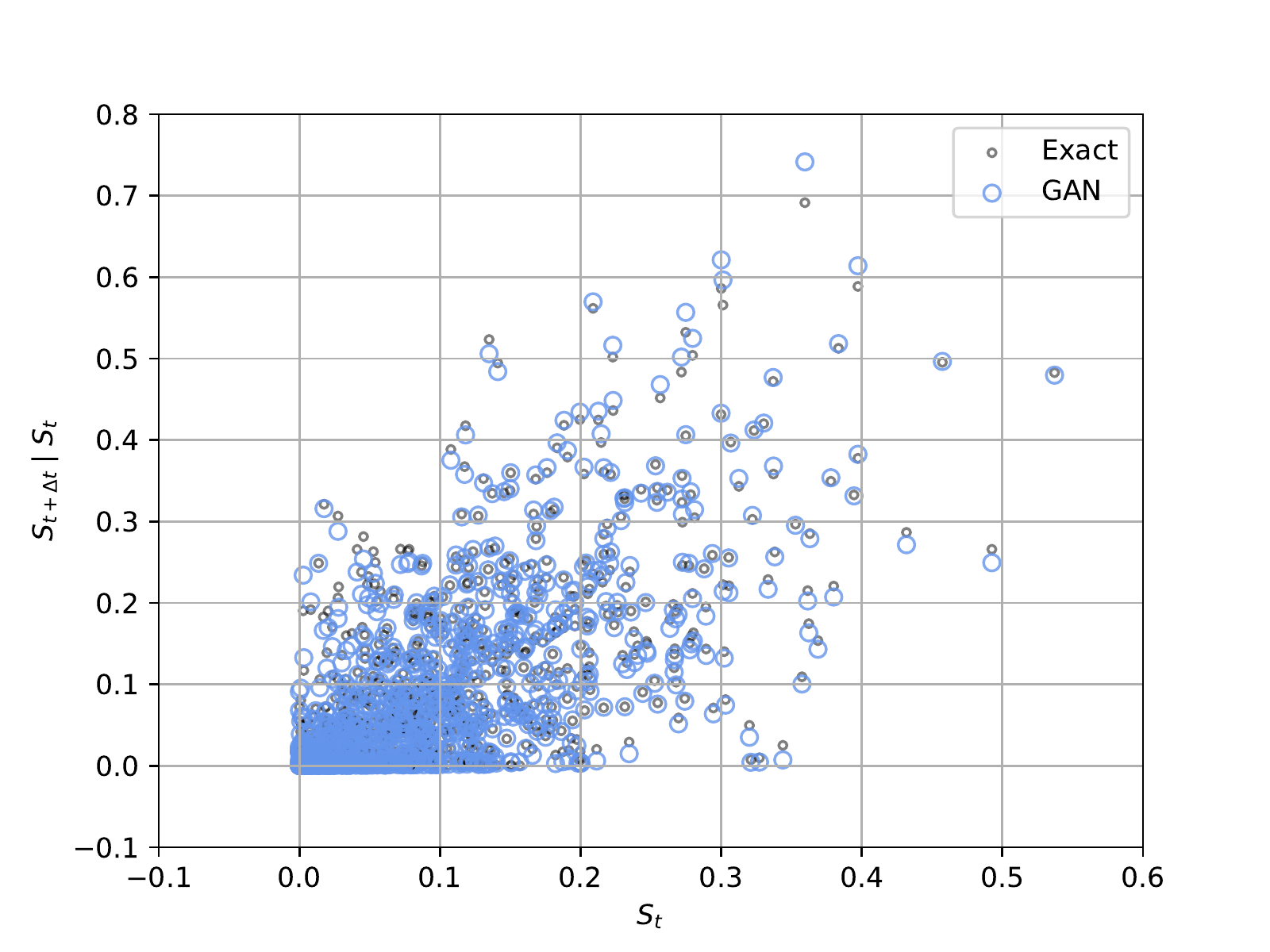}
        \caption{Vanilla GAN}
        \label{fig:CIR_Feller_not_vanilla_autocorr}
    \end{subfigure}
    \begin{subfigure}{0.49\linewidth}
        \includegraphics[width=\linewidth]{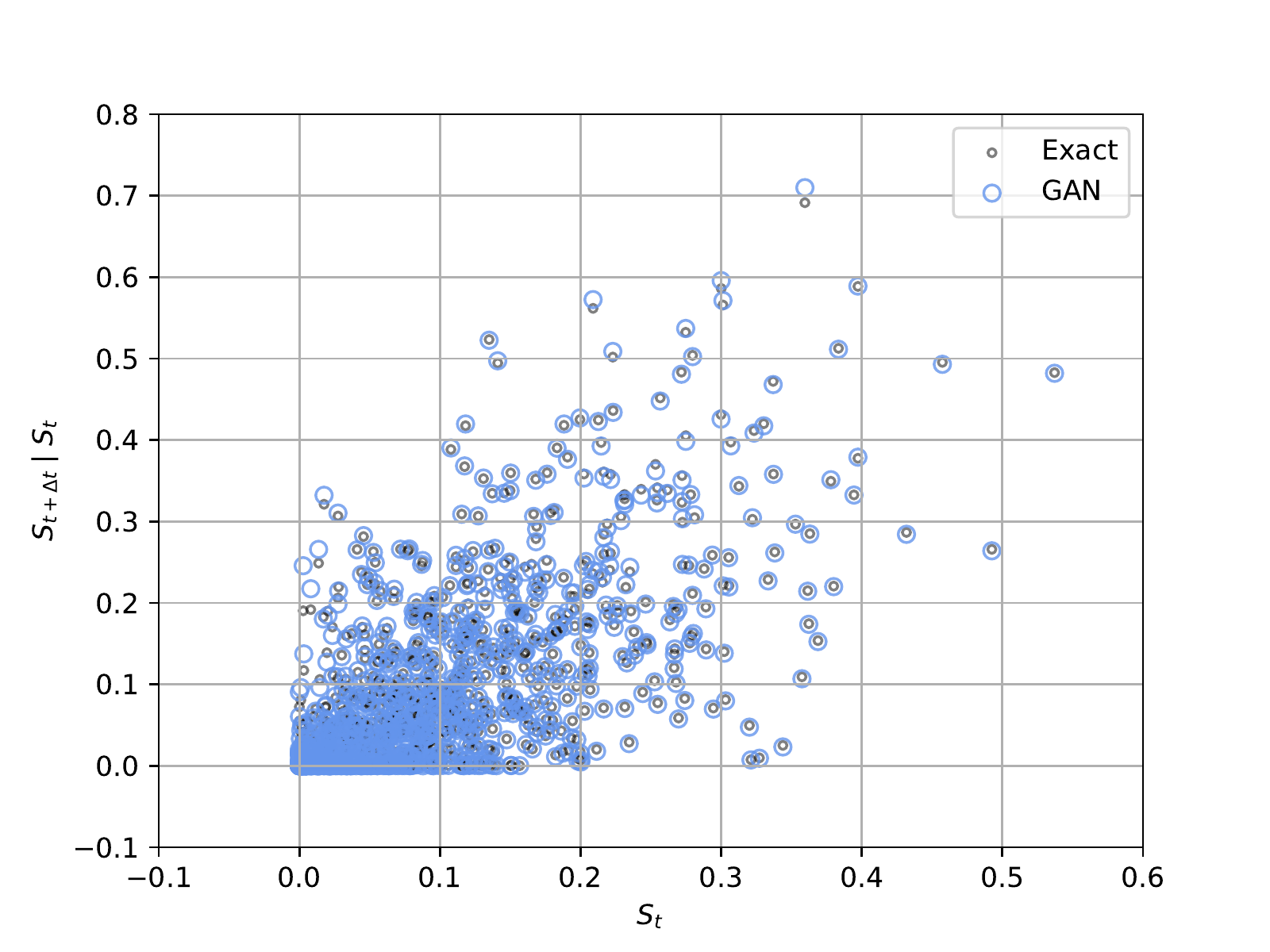}
        \caption{Supervised GAN}
        \label{fig:CIR_Feller_not_supervised_autocorr}
    \end{subfigure}    
    \caption{Scatter plot of $S_{t+\Delta t}\mid S_t$ versus $S_t$. In \ref{fig:CIR_vanilla_autocorr} and \ref{fig:CIR_supervised_autocorr}, the Feller condition is satisfied, while in \ref{fig:CIR_Feller_not_vanilla_autocorr} and \ref{fig:CIR_Feller_not_supervised_autocorr} it is violated. In both cases, the similar shapes of the point clouds reveal that the autocorrelation structures of the GAN output and reference samples are similar, demonstrating good weak approximation capabilities for $\Delta t=1$. Strong approximation is reflected in the data points overlapping with the exact samples.}
    \label{fig:CIR_autocorr}
\end{figure}

\end{document}